\documentclass[pdflatex,sn-mathphys-num,iicol]{sn-jnl}

\usepackage[T1]{fontenc}
\input{glyphtounicode}\pdfgentounicode=1
\usepackage[utf8]{inputenc}
\usepackage{lmodern}
\usepackage{microtype}
\usepackage{amsmath,amssymb,amsfonts}
\usepackage{bm}
\usepackage{graphicx}
\usepackage{booktabs,tabularx,array}
\usepackage{xcolor}
\usepackage{tikz}
\usetikzlibrary{arrows.meta,calc}
\usepackage{xurl}
\def\burl#1{\url{#1}}
\usepackage[capitalise,noabbrev]{cleveref}

\newcommand{\vd}{\mathbf{d}}
\newcommand{\vr}{\mathbf{r}}
\newcommand{\vt}{\mathbf{t}}
\newcommand{\vn}{\mathbf{n}}
\newcommand{\vl}{\mathbf{l}}
\newcommand{\vc}{\mathbf{c}}

\newcommand{\vv}{\mathbf{v}}
\newcommand{\fmin}{f_{\min}}

\newcommand{\vm}{\mathbf{m}}
\newcommand{\OR}{Online Resource~1}

\theoremstyle{thmstyleone}
\newtheorem{theorem}{Theorem}
\newtheorem{proposition}[theorem]{Proposition}
\newtheorem{lemma}[theorem]{Lemma}
\newtheorem{corollary}[theorem]{Corollary}
\theoremstyle{thmstylethree}
\newtheorem{definition}[theorem]{Definition}
\theoremstyle{thmstyletwo}
\newtheorem{remark}[theorem]{Remark}

\begin{document}

\title[What Right-Angle Corners Reveal About an Uncalibrated Camera]{What Right-Angle Corners Reveal About an Uncalibrated Camera: Convex Cones, Certified Bounds and Identifiability}

\author*{\fnm{Michael} \sur{Werman}}\email{michael.werman@mail.huji.ac.il}

\affil{\orgdiv{School of Computer Science and Engineering}, \orgname{The Hebrew University of Jerusalem}, \orgaddress{\city{Jerusalem}, \country{Israel}}}

\abstract{The corner of a box or a room, where three perpendicular edges meet, appears in an image as three segments meeting at a point. We ask what such junctions reveal about an uncalibrated camera's intrinsic parameters. Each junction is consistent with an explicit convex set of cameras; when only the focal length is unknown, this set is an interval in closed form. These sets are computed from the image alone, so calibrating from many junctions is a convex feasibility problem. Almost every junction bounds the focal length from below; we characterize when one also bounds it from above, and when many junctions determine it exactly. For a known vertex, the bounds can be certified against any bounded error in the measured edge directions.}

\keywords{camera calibration, absolute conic, right-angle corners, Manhattan world, identifiability, convex feasibility\\[4pt]\textbf{MSC Classification:} 68T45, 68U10, 65D19, 51N15, 52A20, 90C25}

\maketitle

\section{Introduction}\label{sec:intro}

A right-angle corner, three mutually perpendicular edges meeting at a point, appears in an image as a junction of three line segments. Such corners are ubiquitous: on boxes, in rooms, at doorframes and on furniture. Unlike a generic feature point, a corner carries metric information, namely three directions known to be mutually orthogonal. This paper asks what such junctions reveal about the intrinsic parameters of the camera that images them. The contribution is theoretical: without locating any vanishing point, we determine exactly which cameras junctions can rule out, each by a single junction, and which no junction can exclude. The answer gives bounds on the camera that the data guarantee, even when there are too few lines to find vanishing points.

Earlier work on what a camera learns from right-angle corners covers two extremes: a single corner seen by a calibrated camera, and a whole box seen by an uncalibrated one. For a single junction and a calibrated camera, \citet{perkins1968} and \citet{kanade1980} analyzed orthographic images of trihedral corners, and \citet{barnard1983} and \citet{horaud1987} back-projected the image lines to interpretation planes. \citet{kanatani1986} derived the closed form under orthography, with its two junction types, fork and arrow. \citet{kanatani1988} and \citet{shakunaga1986,shakunaga1989} extended the analysis to perspective: the orientation is unique up to a mirror, and a condition decides whether a junction can be the image of a right-angle corner at all. The closest precedent for reading that condition as a feasibility problem is the labeling of line drawings \citep{huffman1971,clowes1971}, in particular the algebraic treatment of \citet{sugihara1986}, which decides whether a drawing is realizable by solving a linear feasibility problem. At the other extreme, for a whole parallelepiped and an uncalibrated camera, \citet{wilczkowiak2001,wilczkowiak2002,wilczkowiak2005} fit a canonic projection matrix to six or more vertices; each right angle then contributes one linear equation on the calibration.

The common case lies between these extremes: an uncalibrated camera that sees one corner or a few, as in an image of a room or of a single object. Existing methods treat such scenes by exploiting long line families: Manhattan-world methods \citep{coughlan1999,schindler2004}, calibration from vanishing points \citep{caprile1990,cipolla1999,kosecka2002,wildenauer2012}, single-view metrology \citep{criminisi2000} and vanishing-point detection \citep{tardif2009}. Other approaches require more structure: \citet{jelinek2001} need a full polyhedral model with model-to-image correspondences, and \citet{ramalingam2013} combine Manhattan junctions with vanishing points to estimate indoor layout, as do the layout methods of \citet{hedau2009} and \citet{lee2009}. Learned calibrators \citep{workman2015,holdgeoffroy2018,jin2023,veicht2024} regress the camera directly from pixels. We ask instead what the individual junctions determine.

The starting point is a simple observation. Each image line of a junction back-projects to a plane through the camera center, and because the three lines meet at the image of the vertex, all three planes contain the same viewing ray. For every calibration matrix $K$, this turns the condition that the junction be the image of a right-angle corner into the negativity of a product of three forms, each linear in the dual image of the absolute conic $\omega^{*}=KK^{\top}$ (\cref{prop:general}). In plain terms, the cameras for which a junction can be the image of a right-angle corner form a region bounded by three planes in the space of cameras. Among positive-definite forms this region is a convex cone, the positive combinations of three rank-one forms on the plane of lines through the junction (\cref{thm:convex}); when only the focal length is unknown, the cone reduces to an interval with closed-form endpoints (\cref{thm:interval}). Since each junction's constraint is computed from the image alone, calibration from junctions is a convex feasibility problem (\cref{sec:program}).

Junctions become far more informative when several corners belong to one rectilinear structure, such as a building or a room, whose edges all run along the same three perpendicular directions, a Manhattan frame. The edges of all its corners then converge on the same three vanishing points. Call the line through two of these vanishing points the \emph{edge-on line} of the third direction; classically, it is the vanishing line of the planes perpendicular to that direction. The name records what happens to a corner whose vertex lies on this line: the face spanned by its other two edges is seen edge-on, and those two edges appear as a single image line. These lines organize the main result.
Each junction still imposes only an inequality, but together the junctions of one frame decide exactly which cameras can be rejected (\cref{thm:converse}). When the edge-on line of a direction \emph{passes through} the image, meeting it in a segment (\cref{fig:vlines}), every camera under which that direction's vanishing point is not orthogonal to the other two is rejected by some junction near that line. Conversely, cameras that satisfy these orthogonalities and lie near the true one pass every junction, so no set of junctions can reject them. Junctions therefore never carry more information than the vanishing points, and strictly less when fewer than two edge-on lines pass through the image. We say that junctions \emph{determine} a parameter when every wrong value is rejected by some junction.

Three consequences follow. For the focal length alone, one frame determines it when the edge-on line of a direction tilted out of the image plane passes through the image (\cref{thm:identframe}), and a junction bounds it from above exactly when its vertex lies in a region bounded by circles whose diameters join pairs of vanishing points (\cref{thm:boundmap}). \Cref{thm:fpp} treats focal length and principal point together. Several frames, such as buildings at different angles or different images, determine a general camera exactly when their visible directions do not all lie in the union of a plane and its normal (\cref{thm:frames}); this is the junction form of a familiar critical configuration of self-calibration \citep{hartley1997,sturm1997,deagapito2001,kahl2000}.

Expressed through vanishing points, the statements about frames are classical: orthogonal vanishing points impose linear conditions on the image of the absolute conic \citep{hartley2004,liebowitz1999}. The novelty is that junctions reach part of this information without locating any vanishing point and without knowing any two lines to be parallel. This is also the practical motivation: a cluttered room or a close view supplies junctions with short edges but no line family long enough to locate a vanishing point. On the one real camera tested, which is nearly level and has an ordinary field of view, junctions yield a near-tight lower bound on the focal length but not an estimate, and long line families, where they exist, remain the better route (\OR, Sect.~S12). The real junctions are built from segments labeled by Manhattan direction, whereas the per-junction program requires no such labels.

\paragraph{Contributions.} \Cref{tab:roadmap} summarizes the results by camera model. The new ones are the cone and interval of one junction (\cref{thm:convex,thm:interval}); certified bounds, and their robustness to noise, non-right angles and lens distortion (\cref{prop:certified}; \OR, Sect.~S6); the cameras that the junctions of one structure reject (\cref{thm:converse}); the region in which a junction bounds the focal length, and when one structure determines it (\cref{thm:boundmap,thm:identframe}); the results for several structures and for an unknown principal point (\cref{thm:frames,thm:fpp}); and calibration as a convex program (\cref{prop:program}). Of these, the central ones are the cone of \cref{thm:convex}, the certified bounds of \cref{prop:certified}, the region of \cref{thm:boundmap} in which a junction bounds the focal length from above, and the identifiability condition of \cref{thm:identframe}. Two classical results are restated in junction form because the analysis uses them: the right-angle condition, which rests on the angle between two planes \citep{hartley2004} (\cref{prop:general}), and the cameras that one structure cannot tell apart (\cref{prop:oneframe}).

\begin{table*}[t]
\centering\small
\caption{What junctions reveal about each camera model, with the results that prove it. A frame is the set of corners of one rectilinear structure, whose edges share three perpendicular directions. ``Determined'' means that every wrong value is rejected by some junction}
\label{tab:roadmap}
\begin{tabularx}{\linewidth}{@{}l >{\raggedright\arraybackslash}X >{\raggedright\arraybackslash}X >{\raggedright\arraybackslash}X@{}}
\toprule
 & general camera & focal length only & focal length and principal point\\
\midrule
one junction & a convex cone, the same at every vertex up to coordinates (\cref{thm:convex,cor:alike}) & an interval in closed form (\cref{thm:interval}) & the cone restricted to these cameras (\cref{prop:aspect,prop:pp})\\
one structure (frame) & never determined (\cref{prop:oneframe}); rejects exactly the cameras violating the orthogonalities whose edge-on lines pass through the image (\cref{thm:converse,prop:oneline}) & determined iff the edge-on line of a tilted axis reaches the image away from $\vc^{\star}$ (\cref{thm:identframe,cor:wide}); upper bounds inside circles (\cref{thm:boundmap}) & determined unless cases (1) or (2) of \cref{thm:fpp} hold\\
several structures (frames) & determined iff the visible axes do not lie in a plane and its normal (\cref{thm:frames,cor:horizons}) & as for one structure, each on its own & determined unless cases (1) or (2) of \cref{thm:fpp}\\
\bottomrule
\end{tabularx}
\end{table*}

\paragraph{Organization.} \Cref{sec:single} restates the single-corner result of \citet{kanatani1988}, for a calibrated camera, in the vector form used throughout. \Cref{sec:condition} derives the right-angle condition for an arbitrary camera, \cref{sec:bound} describes the cameras that one junction admits, and \cref{sec:many} establishes what the junctions of rectilinear structures determine. \Cref{sec:experiments} tests these predictions, and \cref{sec:conclusion} summarizes the results and the open questions. \OR{} collects the derivations, experiments and numerical checks that the article cites by section; \cref{tab:notation} lists the notation.

\begin{table*}[t]
\centering\footnotesize
\caption{Principal notation. Image coordinates are centered on the principal point wherever it is taken as known}
\label{tab:notation}
\begin{tabularx}{\linewidth}{@{}l >{\raggedright\arraybackslash}X@{}}
\toprule
Symbol & Meaning \\
\midrule
$K$;\ \ $\omega^{*}=KK^{\top}$;\ \ $\omega=(\omega^{*})^{-1}$ & calibration matrix; dual image of the absolute conic; image of the absolute conic \\
$\vc=(u,v,1)$;\ \ $\rho=\sqrt{u^2+v^2}$ & the junction, which is the image of the corner's vertex; its distance from the principal point \\
$\vl_i=(a_i,b_i,e_i)$ & image line of edge $i$, with $a_i^2+b_i^2=1$, so that $e_i$ is its signed distance from the principal point \\
$\vd_i$ & unit direction of edge $i$ in the scene \\
$\vr=K^{-1}\vc$;\ \ $\vn_i=K^{\top}\vl_i$;\ \ $\vt_i=\vn_i\times\vr$ & viewing ray; normal of the interpretation plane of edge $i$; direction in that plane perpendicular to the ray \\
$g_{ij}=\vt_i\cdot\vt_j$;\ \ $k_i$ & the quantities of the closed form, in which $\vd_i\propto k_i\vr+\vt_i$ \\
$A_{ij}=a_ia_j+b_ib_j$ & cosine of the raw image angle between half-edges $i$ and $j$ \\
$\psi_i$ & signed angle from the direction pointing away from the principal point to half-edge $i$ \\
$x=f^2$;\ \ $x_{ij}=-e_ie_j/A_{ij}$;\ \ $f_{ij}=\sqrt{x_{ij}}$ & squared focal length; the threshold contributed by the pair $(i,j)$ \\
$S$;\ \ $F$;\ \ $\fmin$ & the pairs with a positive threshold; the admissible set of focal lengths of one junction; its infimum \\
$\vv_i$;\ \ $\ell_i$ & vanishing point of axis $i$; edge-on line of axis $i$, which is the vanishing line through the other two vanishing points \\
$\vm_k=\vc\times\vl_k$;\ \ $\nu_k$ & generators of the cone of cameras a junction admits; the coordinates of $\omega^{*}$ on them, all positive exactly when the junction is admissible \\
$\Xi=K^{-1}\omega^{*}K^{-\top}$;\ \ $\bar{\vd}$ & a candidate camera relative to the true one; the first two coordinates of an axis $\vd$, its direction in the image plane \\
$N=V^{-1}\omega^{*}V^{-\top}$;\ \ $\boldsymbol\zeta=V^{-1}\vc$ & a candidate camera and a junction in the basis $V=[\vv_1\,\vv_2\,\vv_3]$ of the vanishing points; $N$ is diagonal exactly when they are mutually orthogonal \\
$z_a=\vd_a\cdot\hat{\mathbf z}$;\ \ $w_a=z_a/\zeta_a$;\ \ $\gamma_a$ & the sine of the tilt of axis $a$ out of the image plane; the quantities of \eqref{eq:gform}, whose extremes over the junctions give the joint admissible set of focal lengths \\
$\lambda_{ij}$;\ \ $\mu=f^2/\rho^2$ & the quantities of the angle parameterization of \cref{thm:interval} \\
$f'$;\ \ $s=f'/f$ & a candidate focal length; its ratio to the true one \\
$\Gamma^{\pm}$ & the suprema of $\pm\gamma_a$ in \eqref{eq:capF} \\
$\hat{\mathbf z}$;\ \ $\vc^{\star}_k$ & the optical axis; the point of $\ell_k$ nearest the principal point \\
$\ell_{\mathrm{seg}}$ & the length of an observed edge segment, in pixels \\
$\varphi_i$;\ \ $\mathcal U$ & a bound on the error of the measured angle of edge $i$; the uncertain set of \cref{prop:certified} \\
$P$;\ \ $\mathcal I$ & the corner's vertex in the scene; the image, a convex polygon, in \cref{sec:many} \\
$M$;\ \ $\sigma$ & number of junctions; standard deviation of the endpoint noise, in pixels \\
\bottomrule
\end{tabularx}
\end{table*}

\section{The single corner, restated}\label{sec:single}

This section restates \citet[Theorem~4]{kanatani1988} to fix the notation and the sign condition on which the rest of the paper builds.

\subsection{Setup}\label{sec:setup}

\begin{figure}[t]
\centering
\begin{tikzpicture}[scale=0.94,>=Stealth,font=\footnotesize]
  \coordinate (O) at (0,0);
  \coordinate (P) at (5.6,3.1);
  \coordinate (D1) at ($(P)+(1.35,-0.45)$);
  \coordinate (D2) at ($(P)+(0.20,1.30)$);
  \coordinate (D3) at ($(P)+(-1.50,0.75)$);
  \coordinate (c)  at ($(O)!0.45!(P)$);
  \coordinate (pp) at (2.700,0.673);
  \fill[black!11] (O) -- (P) -- (D1) -- cycle;
  \fill[black!11] (O) -- (P) -- (D2) -- cycle;
  \fill[black!22] (O) -- (P) -- (D3) -- cycle;
  \draw[black!45] (O) -- (D3);
  \draw[black!45,dash dot] (O) -- (pp);
  \fill[white,opacity=0.6] (1.363,2.625) -- (3.263,1.827) -- (4.037,-1.279) -- (2.137,-0.481) -- cycle;
  \draw[black!60] (1.363,2.625) -- (3.263,1.827) -- (4.037,-1.279) -- (2.137,-0.481) -- cycle;
  \fill[black!55] (pp) circle (1.0pt);
  \draw[black!45,dash dot] (pp) -- ($(O)!1.55!(pp)$);
  \draw[dashed] (O) -- (P);
  \draw[->,thick] (O) -- ($(O)!0.20!(P)$);
  \coordinate (e1) at ($(c)!0.60cm!($(O)!0.45!(D1)$)$);
  \coordinate (e2) at ($(c)!0.55cm!($(O)!0.45!(D2)$)$);
  \coordinate (e3) at ($(c)!0.60cm!($(O)!0.45!(D3)$)$);
  \draw[thick] (c) -- (e1) (c) -- (e2) (c) -- (e3);
  \fill (c) circle (1.3pt);
  \draw[->,thick] (P) -- (D1);
  \draw[->,thick] (P) -- (D2);
  \draw[->,thick] (P) -- (D3);
  \fill (P) circle (1.5pt);
  \fill (O) circle (1.5pt);
  \draw[->] ($(O)!0.72!(D3)$) -- ++(-0.26,0.58);
  \coordinate (Q) at ($(O)!0.78!(P)$);
  \draw[->,thick,black!70] (Q) -- ($(Q)!0.75cm!90:(P)$);
  \node[left=0pt] at ($(Q)!0.75cm!90:(P)$) {$\vt_3$};
  \node[below left=-1pt] at (O) {$O$};
  \node[above left=-2pt] at ($(O)!0.20!(P)$) {$\vr$};
  \node[below=1pt] at (c) {$\vc$};
  \node[right=1pt] at (P) {$P$};
  \node[right] at (D1) {$\vd_1$};
  \node[above] at (D2) {$\vd_2$};
  \node[left] at (D3) {$\vd_3$};
  \node[right=-1pt] at (e1) {$\vl_1$};
  \node[right=-1pt] at (e2) {$\vl_2$};
  \node[left=-1pt] at (e3) {$\vl_3$};
  \node at (3.55,2.62) {$\Pi_3$};
  \node[left=-1pt] at ($(O)!0.72!(D3)+(-0.26,0.58)$) {$\vn_3$};
\end{tikzpicture}
\caption{Corner geometry. The corner $P$ has unit edge directions $\vd_i$, and each image line $\vl_i$ through the junction $\vc$ pins its edge to the interpretation plane $\Pi_i$ through the camera center $O$, with normal $\vn_i$. The dashed viewing ray $\vr$ lies in all three planes ($\Pi_1$ and $\Pi_2$ are drawn lighter), and $\vt_3$ is the direction in $\Pi_3$ perpendicular to the ray. The image plane is perpendicular to the optical axis (dash-dotted) and centered on the principal point, so the junction lies off center, as in a real image. The constraints are $\vd_i\perp\vn_i$ for each edge and $\vd_1\cdot\vd_2=\vd_2\cdot\vd_3=\vd_3\cdot\vd_1=0$}
\label{fig:geometry}
\end{figure}
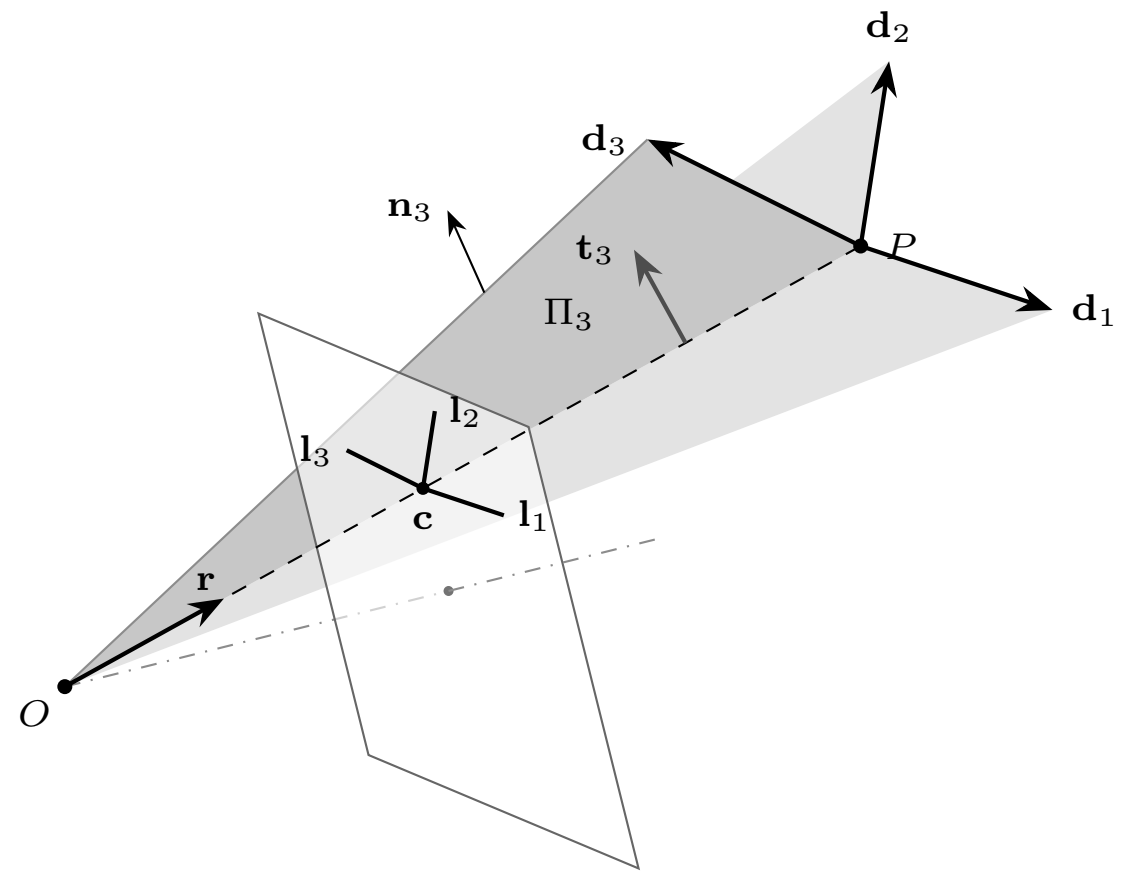

A trihedral right-angle corner is a point $P$ with three unit edge directions $\vd_1,\vd_2,\vd_3$ satisfying $\vd_i\cdot\vd_j=0$ for $i\neq j$. A pinhole camera with intrinsic matrix $K$ and center $O$ at the origin images $P$ at $\vc$ and maps each edge onto an image line $\vl_i$ through $\vc$. Together with $O$, each $\vl_i$ spans an interpretation plane $\Pi_i$ \citep{barnard1983}, whose normal in camera coordinates is $\vn_i=K^{\top}\vl_i$, and the edge $\vd_i$ lies in $\Pi_i$. Because the three lines pass through the same image point, the ray $\vr=K^{-1}\vc$ lies in all three planes (\cref{fig:geometry}); this shared ray is what the rest of the paper exploits.

Let $\vt_i=\vn_i\times\vr$, the direction in $\Pi_i$ perpendicular to the ray. Then $\{\vr,\vt_i\}$ spans $\Pi_i$, and each edge decomposes into a component along the ray and one across it:
\begin{equation}
\vd_i=\alpha_i\,\vr+\beta_i\,\vt_i .
\end{equation}
The coefficient $\alpha_i$ measures how the edge moves along the viewing ray, toward or away from the camera, and $\beta_i$ how it moves across the ray. The three $\vt_i$ lie in the plane perpendicular to $\vr$, and the angles between them are those a camera would see if it were rotated to look straight at the corner, the \emph{canonical view}. This is Kanatani's standard transformation, written as a change of basis.

\paragraph{Orientation convention.} We write each line as $\vl_i=(a_i,b_i,e_i)$ in centered pixel coordinates, with $a_i^2+b_i^2=1$ and the sign chosen so that the observed half-edge leaves $\vc$ in the image direction $(b_i,-a_i)$. This choice makes $\vt_i$ point along the observed half-edge for every $f$: for $K=\mathrm{diag}(f,f,1)$ and $\vc=(u,v,1)$, the derivative of the projection along $\vt_i$ at $\vc$ is $(f^2+u^2+v^2)\,(b_i,-a_i)/P_z$, a positive multiple of the observed direction, where $P_z$ is the depth of $P$. The convention only fixes names. Admissibility, defined in \cref{sec:sign}, does not depend on it, since reversing $\vt_i$ changes the sign of two of the products $g_{ij}$ introduced below; what the convention fixes are the fork and arrow labels, and which of the two mirror-image solutions of \cref{sec:mirror1} is called which.

\subsection{Closed form}\label{sec:closed}

Recovering the corner amounts to recovering, for each edge, the ratio $k_i=\alpha_i/\beta_i$ of its motion along the ray to its motion across it. With $g_{ij}=\vt_i\cdot\vt_j$, the orthogonality of two edges gives, since $\vr\perp\vt_i$,
\begin{equation}
\begin{aligned}
&\vd_i\cdot\vd_j=\alpha_i\alpha_j\|\vr\|^2+\beta_i\beta_j g_{ij}=0\\
&\qquad\Longrightarrow\quad k_ik_j=-\frac{g_{ij}}{\|\vr\|^2}.
\end{aligned}
\label{eq:kk}
\end{equation}
Each pair of edges thus fixes the product of two ratios. The three such equations in three unknowns imply
\begin{equation}
\begin{aligned}
k_1^2&=-\frac{g_{12}\,g_{31}}{g_{23}\,\|\vr\|^2},\\
k_2^2&=-\frac{g_{12}\,g_{23}}{g_{31}\,\|\vr\|^2},\\
k_3^2&=-\frac{g_{23}\,g_{31}}{g_{12}\,\|\vr\|^2},
\end{aligned}
\label{eq:k2}
\end{equation}
which is equation (7.3) of \citet{kanatani1988} in vector form. The products fix the relative signs of the $k_i$, so the only remaining freedom is a global sign, and the edges are  $\vd_i=(k_i\vr+\vt_i)/\|k_i\vr+\vt_i\|$.

\subsection{The sign condition}\label{sec:sign}

The closed form exists only for some junctions, because the right sides of \eqref{eq:k2} must be positive. That condition is the subject of the rest of the paper, so we state it formally.

\begin{definition}[Junction; admissibility]\label{def:admissible}
A \emph{junction} is a point $\vc$ of the image together with three distinct image lines $\vl_1,\vl_2,\vl_3$ through it. A camera $K$ \emph{admits} the junction if the junction is the image under $K$ of a right-angle corner, that is, if there are mutually orthogonal directions $\vd_1,\vd_2,\vd_3$ with $\vd_i$ in the interpretation plane $\Pi_i$ of $\vl_i$ and not along the viewing ray. We then also say that the junction is \emph{admissible} under $K$, or has a \emph{right-angle interpretation} under $K$. Throughout, $K$ is upper triangular with positive diagonal and $K_{33}=1$; points and lines of the image are vectors of $\mathbb R^3$ determined up to a nonzero factor, and admissibility does not depend on the factors chosen.
\end{definition}

\begin{lemma}[Sign condition]\label{lem:sign}
A camera admits a junction if and only if the quantities $g_{ij}=\vt_i\cdot\vt_j$ of \cref{sec:closed} satisfy
\begin{equation}
g_{12}\,g_{23}\,g_{31}<0 ,
\label{eq:sign}
\end{equation}
and the corner is then determined up to the global sign of the ratios $(k_1,k_2,k_3)$ of \cref{sec:closed}.
\end{lemma}
\begin{proof}
Since $\vn_i\neq0$ and $\vn_i\cdot\vr=\vl_i\cdot\vc=0$, the vector $\vt_i=\vn_i\times\vr$ is nonzero, so $\vr$ and $\vt_i$ span $\Pi_i$ and every direction of $\Pi_i$ off the ray is a multiple of $k_i\vr+\vt_i$; admissibility is therefore \eqref{eq:kk} for the three pairs. Their product gives $(k_1k_2k_3)^2\|\vr\|^6=-g_{12}g_{23}g_{31}$, and no $g_{ij}$ vanishes: otherwise $k_ik_j=0$, say $k_i=0$, so $\vt_i$ is orthogonal to the other two $\vt$'s in the plane perpendicular to the ray, and their image lines coincide. Conversely, under \eqref{eq:sign} the right sides of \eqref{eq:k2} are positive, and each choice of sign for $k_1$ fixes $k_2$ and $k_3$ through \eqref{eq:kk}, which determines the products $k_ik_j$.
\end{proof}

\begin{figure}[t]
\centering
\begin{tikzpicture}[scale=0.92,font=\footnotesize]
  \tikzset{edge/.style={thick},ang/.style={black!70}}
  \begin{scope}[shift={(0,0)}]
    \foreach \a in {90,210,330} \draw[edge] (0,0) -- (\a:1.1);
    \fill (0,0) circle (1.4pt);
    \node[ang] at (150:0.62) {$120^\circ$};
    \node[ang] at (30:0.62)  {$120^\circ$};
    \node[ang] at (270:0.55) {$120^\circ$};
    \node at (0,-1.45) {(a)};
  \end{scope}
  \begin{scope}[shift={(3.05,0)}]
    \foreach \a in {10,90,170} \draw[edge] (0,0) -- (\a:1.1);
    \fill (0,0) circle (1.4pt);
    \node[ang] at (50:0.66)  {$80^\circ$};
    \node[ang] at (130:0.66) {$80^\circ$};
    \draw[black!55] (10:1.25) arc (10:170:1.25);
    \node[ang] at (90:1.47) {$160^\circ$};
    \node at (0,-1.45) {(b)};
  \end{scope}
  \begin{scope}[shift={(6.1,0)}]
    \foreach \a in {0,70,190} \draw[edge] (0,0) -- (\a:1.1);
    \fill (0,0) circle (1.4pt);
    \node[ang] at (35:0.68)  {$70^\circ$};
    \node[ang] at (130:0.66) {$120^\circ$};
    \node[ang] at (275:0.55) {$170^\circ$};
    \node at (0,-1.45) {(c)};
  \end{scope}
\end{tikzpicture}
\caption{Junction types, drawn as a camera would see them if rotated to look straight at the vertex (the canonical view). Each label is the angle between two edges at the vertex, at most $180^\circ$. A junction is the image of a right-angle corner exactly when an odd number of these three angles are obtuse: (a) the fork, with three; (b) the arrow, with one; (c) two, which no right-angle corner produces}
\label{fig:junctions}
\end{figure}
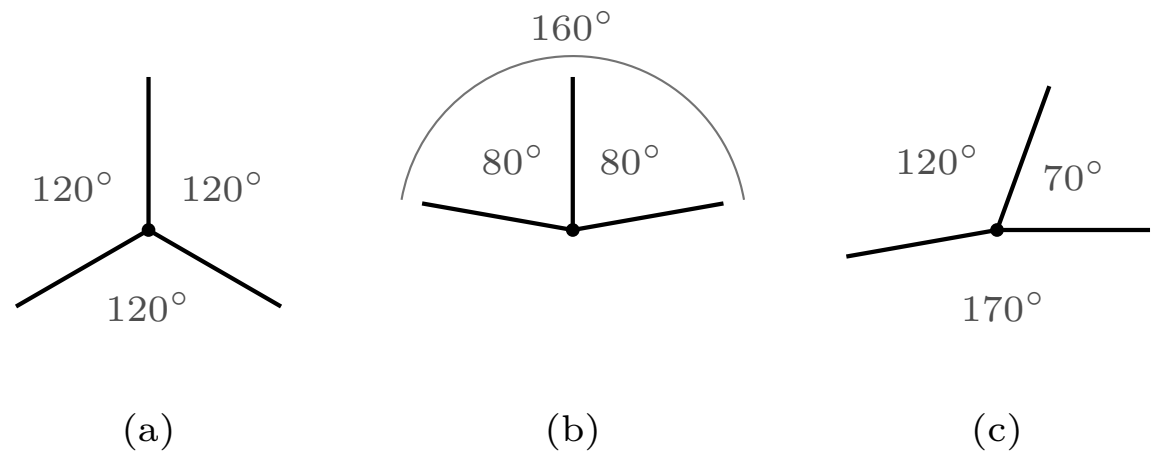

Since $g_{ij}$ is a positive multiple of the cosine of the angle between edges $i$ and $j$ in the canonical view, the condition requires an odd number of those angles to be obtuse: all three, as in the fork, or exactly one, as in the arrow (\cref{fig:junctions}). This is the fork-or-arrow rule \citep{kanatani1986,kanatani1988,shakunaga1989}, applied to the canonical angles rather than to the raw image angles, from which they differ away from the principal point. For a junction away from the principal point, the canonical view is not the actual image: it is the view after turning the camera about its center until the vertex lies on the principal ray, and the canonical angles differ from the raw ones by an amount that depends on the focal length. At the principal point the two coincide for every $f$, so such a junction says nothing about the focal length (\cref{rem:exceptional}); as $f$ grows, the required rotation shrinks and the canonical angles approach the raw ones. Because the canonical view depends on $K$, so does the test; \cref{sec:bound} turns this dependence into a constraint on the camera.

Two configurations are degenerate. If $g_{ij}=0$, the third edge, the one other than $i$ and $j$, lies along the ray and projects to a point; if two edges lie on one image line, their interpretation planes coincide and the system loses rank. We call a corner \emph{nondegenerate} if neither occurs.

\subsection{The mirror}\label{sec:mirror1}

The global flip $(k_1,k_2,k_3)\mapsto-(k_1,k_2,k_3)$ reflects each $\vd_i$ in the plane through $O$ perpendicular to $\vr$, so the two solutions are a corner and its mirror image. This is the Necker ambiguity \citep[Theorem~4]{kanatani1988}: for a fork, one solution has all edges receding from the camera ($\alpha_i>0$, a box seen from outside) and the other has all edges approaching (a room corner seen from inside). \OR{} (Sects.~S2 and S4) shows when a second corner resolves the ambiguity, and what opacity adds.

\section{The right-angle condition for an uncalibrated camera}\label{sec:condition}

Every result in \cref{sec:single}, like the earlier work on single corners cited in \cref{sec:intro}, assumes $K$ known. The condition \eqref{eq:sign} for a junction to be the image of a right-angle corner involves $K$ through the ray, the normals and the directions $\vt_i$, and at first sight it depends on the camera in a complicated way. This section shows that the dependence is in fact simple: for any calibration matrix, it is linear, pair by pair, in $\omega^{*}=KK^{\top}$.

\subsection{Three planes through one ray}\label{sec:identity}

Write $\vn_i=K^{\top}\vl_i$ for the normal of the interpretation plane of the image line $\vl_i$, and $\vr=K^{-1}\vc$ for the viewing ray of the junction $\vc$. Then $\vn_i\cdot\vr=\vl_i^{\top}KK^{-1}\vc=\vl_i\cdot\vc=0$ because the line passes through the junction: whatever $K$ is, each interpretation plane contains the ray. Everything that follows rests on this observation, and it simplifies the condition at once. By the identity $(\mathbf{x}\times\mathbf{z})\cdot(\mathbf{y}\times\mathbf{z})=(\mathbf{x}\cdot\mathbf{y})\|\mathbf{z}\|^2-(\mathbf{x}\cdot\mathbf{z})(\mathbf{y}\cdot\mathbf{z})$, valid for all $\mathbf{x},\mathbf{y},\mathbf{z}\in\mathbb{R}^3$, the cross products reduce to scalar products of the normals:
\begin{equation}
g_{ij}=\vt_i\cdot\vt_j=(\vn_i\times\vr)\cdot(\vn_j\times\vr)=\|\vr\|^2\,(\vn_i\cdot\vn_j),
\label{eq:gfactor}
\end{equation}
and since $\|\vr\|^2>0$, the sign condition \eqref{eq:sign} becomes
\begin{equation}
(\vn_1\cdot\vn_2)(\vn_2\cdot\vn_3)(\vn_3\cdot\vn_1)<0 ,
\label{eq:planes}
\end{equation}
a statement about the three interpretation planes alone: the junction admits a right-angle interpretation exactly when an odd number of the three products $\vn_i\cdot\vn_j$ are negative, a parity that does not depend on the signs chosen for the normals. To express this through the image lines, note that
\begin{equation}
\vn_i\cdot\vn_j=\vl_i^{\top}\,\omega^{*}\,\vl_j,\qquad \omega^{*}=KK^{\top},
\label{eq:diac}
\end{equation}
the second relation defining the dual image of the absolute conic, whose use for the angle between two planes is classical \citep{triggs1997,hartley2004}. Because $K$ is upper triangular with positive diagonal, $\omega^{*}$ determines it, so working with $\omega^{*}$ loses nothing. Substituting gives the general statement.

\begin{proposition}[General form]\label{prop:general}
For any calibration matrix $K$, a junction of three concurrent image lines admits a right-angle interpretation if and only if
\[
\bigl(\vl_1^{\top}\omega^{*}\vl_2\bigr)\bigl(\vl_2^{\top}\omega^{*}\vl_3\bigr)\bigl(\vl_3^{\top}\omega^{*}\vl_1\bigr)<0 ,
\]
a product of three forms each \emph{linear} in the six entries of $\omega^{*}$.
\end{proposition}
\begin{proof}
By \cref{lem:sign}, \eqref{eq:gfactor} and \eqref{eq:diac}.
\end{proof}

\noindent A junction therefore imposes one inequality on the camera, and its meaning is geometric: each factor changes sign exactly when the two interpretation planes of its pair become \emph{perpendicular}, so these three conditions bound the cameras that admit the junction. The vertex itself need not be visible: the argument uses only the concurrency of the three lines, so a corner whose vertex is occluded, or lies outside the image, can still be used by intersecting its three edges. 

\subsection{Why three edges, and why right angles}\label{sec:why}

\paragraph{Why three edges.} Two edges are not enough, and \cref{prop:general} explains why: a pair contributes a single form, and a single form imposes no constraint. Indeed, every camera admits an L-junction as the image of a right angle. In the notation of \cref{lem:sign}, the directions $k_1\vr+\vt_1$ and $k_2\vr+\vt_2$ are orthogonal exactly when $k_1k_2\|\vr\|^2=-g_{12}$, and this equation has the solutions $k_1\neq0$ arbitrary and $k_2=-g_{12}/(k_1\|\vr\|^2)$ for every camera. A constraint first appears with the product over three pairs in \eqref{eq:planes}. Concurrency is equally essential: it makes $\vn_i\cdot\vr=0$, and hence the forms linear. Three \emph{nonconcurrent} lines pose a different problem, treated in \OR{} (Sect.~S3).

\paragraph{Why right angles.} For a prescribed scene angle other than $90^\circ$, the normalizations $\|\vt_i\|^2$ survive in the pairwise condition. Since they depend nonlinearly on $f$, the linearity in $f^2$ is lost, and the admissible set of a trihedral with known non-right angles is not characterized here (\OR, Sect.~S14). Small departures from $90^\circ$ are analyzed in \OR{} (Sect.~S6).

\section{One junction}\label{sec:bound}

A single junction imposes one inequality, that of \cref{prop:general}. This section describes the set of cameras that satisfy it: a convex cone in general (\cref{sec:convex}); an interval in closed form when the focal length is the only unknown (\cref{sec:admissible}); two further camera models, with an unknown aspect ratio or principal point (\cref{sec:slices}); and certified bounds under measurement error and model error (\cref{sec:noise}).

\subsection{The admissible cameras form a convex cone}\label{sec:convex}

Consider first a general camera. The three forms of \cref{prop:general} are linear in $\omega^{*}$, but the condition that their product be negative does not by itself define a convex set; positive-definiteness is what makes it one. The cameras that admit a junction form a cone whose generators can be read from the junction.

\begin{theorem}[The admissible cameras: a simplicial cone of forms]\label{thm:convex}
Let a junction consist of three distinct image lines $\vl_1,\vl_2,\vl_3$ concurrent at $\vc$, and put $\vm_k=\vc\times\vl_k$. A positive-definite $\omega^{*}$ admits a right-angle interpretation of the junction if and only if its restriction to the plane $\vc^{\perp}=\{\boldsymbol\chi:\boldsymbol\chi\cdot\vc=0\}$ of lines through the junction is a positive combination of the three rank-one forms $(\vm_k\cdot\boldsymbol\chi)^2$, that is,
\begin{equation}
\boldsymbol\chi^{\top}\omega^{*}\boldsymbol\chi=\sum_{k=1}^{3}\nu_k\,(\vm_k\cdot\boldsymbol\chi)^2\quad\text{for all }\boldsymbol\chi\perp\vc
\label{eq:simplex}
\end{equation}
with $\nu_1,\nu_2,\nu_3>0$.
Equivalently, writing $\Delta_{ki}=\det(\vc,\vl_k,\vl_i)$ and $\{i,j,k\}=\{1,2,3\}$,
\begin{equation}
\nu_k=\frac{\vl_i^{\top}\omega^{*}\vl_j}{\Delta_{ki}\Delta_{kj}}>0\qquad\text{for }k=1,2,3 .
\label{eq:nuk}
\end{equation}
The admissible set is therefore a convex cone, and it is never empty. It is one cell of the arrangement of the three forms $\vl_i^{\top}\omega^{*}\vl_j$, intersected with the cone of positive-definite matrices, and on it each form has the sign of $\Delta_{ki}\Delta_{kj}$, which is determined by the junction alone, without reference to $K$.
\end{theorem}

\begin{proof}
Each $\vl_i$ passes through the junction, so $\vl_i\cdot\vc=0$ and all three lie in the plane $\vc^{\perp}$. On that plane the form $q_k(\boldsymbol\chi)=(\vm_k\cdot\boldsymbol\chi)^2$ has rank one, and its polarization satisfies $q_k(\vl_i,\vl_j)=(\vm_k\cdot\vl_i)(\vm_k\cdot\vl_j)=\Delta_{ki}\Delta_{kj}$, because $(\vc\times\vl_k)\cdot\vl_i=\det(\vc,\vl_k,\vl_i)$. This vanishes if $i$ or $j$ equals $k$ and not otherwise, the three lines being distinct. The $q_k$ are therefore linearly independent, since evaluating a relation $\sum_k\alpha_kq_k=0$ at $(\vl_i,\vl_j)$ leaves $\alpha_k\Delta_{ki}\Delta_{kj}=0$. The quadratic forms on a plane make up a space of dimension three, so the restriction of $\omega^{*}$ to $\vc^{\perp}$ is $\sum_k\nu_kq_k$ for unique real $\nu_k$, and evaluating at $(\vl_i,\vl_j)$ gives $\vl_i^{\top}\omega^{*}\vl_j=\nu_k\Delta_{ki}\Delta_{kj}$.

Since $\Delta_{ki}=-\Delta_{ik}$, the product of the three forms is
\[
\prod_{k}\nu_k\Delta_{ki}\Delta_{kj}=-\,\nu_1\nu_2\nu_3\,(\Delta_{12}\Delta_{23}\Delta_{31})^2 ,
\]
so by \cref{prop:general} the junction is admissible exactly when $\nu_1\nu_2\nu_3>0$: all three positive, or exactly two negative. The second cannot happen when $\omega^{*}$ is positive-definite, because $\vl_k^{\top}\omega^{*}\vl_k=\nu_i\Delta_{ik}^2+\nu_j\Delta_{jk}^2$ must be positive for each $k$. The junction is therefore admissible exactly when every $\nu_k$ is positive, which is \eqref{eq:simplex} and \eqref{eq:nuk}.

The conditions \eqref{eq:nuk} are three linear inequalities on $\omega^{*}$ whose signs are fixed by the junction, so the admissible set is one cell of the arrangement, intersected with the positive-definite cone, and it is convex. It is not empty: any positive $\nu_k$ give a positive-definite form on $\vc^{\perp}$, because the functionals $\vm_k\cdot\boldsymbol\chi$ have no common zero there, and every positive-definite form on $\vc^{\perp}$ arises from a valid camera, by extending it to a positive-definite $\omega^{*}$ on $\mathbb R^3$, rescaling so that $\omega^{*}_{33}=1$, and factoring $\omega^{*}=KK^{\top}$ with $K$ upper triangular.
\end{proof}

The vectors $\vm_k$ are image quantities: for a line $\boldsymbol\chi$ through the junction, $\vm_k\cdot\boldsymbol\chi=\det(\vc,\vl_k,\boldsymbol\chi)$ vanishes exactly when $\boldsymbol\chi=\vl_k$, so $(\vm_k\cdot\boldsymbol\chi)^2$ is the form that vanishes on the $k$-th edge line. The theorem constrains only the restriction of $\omega^{*}$ to $\vc^{\perp}$: the simplicial cone is one of restricted forms, and the admitted $\omega^{*}$ are all the positive-definite extensions of its members.

The cone has a simple geometric picture. Taken up to scale, the positive-definite forms on the plane $\vc^{\perp}$ form a disc, the Beltrami--Klein model of the hyperbolic plane, whose boundary circle consists of the rank-one forms. The cameras that a junction admits form the ideal triangle with vertices $q_1$, $q_2$, $q_3$, one for each image line.

\begin{remark}[All junctions at one vertex are alike]\label{cor:alike}
For any two junctions with the same vertex $\vc$, an invertible linear map of $\vc^{\perp}$ carries the forms of the cameras admitted by one onto those admitted by the other. In the Beltrami--Klein model formed by the positive-definite forms on $\vc^{\perp}$ up to scale, the admitted forms are an ideal triangle, and any two such triangles are congruent.
\end{remark}
\begin{proof}
By \cref{thm:convex} the admitted forms are the positive combinations of the $(\vm_k\cdot\boldsymbol\chi)^2$, and the $\vm_k$ restrict to pairwise independent functionals on the plane $\vc^{\perp}$, so there $\vm_3=a\vm_1+b\vm_2$ with $ab\neq0$, and likewise $\vm'_3=a'\vm'_1+b'\vm'_2$. The map $A$ with $A^{\top}\vm'_1=(a/a')\vm_1$ and $A^{\top}\vm'_2=(b/b')\vm_2$ then sends $\vm'_3$ to $\vm_3$, so each $(\vm'_k\cdot A\boldsymbol\chi)^2$ is a positive multiple of $(\vm_k\cdot\boldsymbol\chi)^2$. In the Beltrami--Klein disc of positive-definite forms up to scale, bounded by the rank-one forms, the admitted forms are the ideal triangle spanned by three of them, and linear maps act as hyperbolic isometries.
\end{proof}

Junctions therefore differ not in the shape of their admitted sets but in where those sets lie relative to the cameras of interest, for instance the cameras of \cref{thm:interval}.

Two results below are special cases of this theorem, obtained by restricting the camera to a smaller family. \Cref{thm:interval} restricts it to the cameras $\omega^{*}=\operatorname{diag}(f^2,f^2,1)$, a line in $\omega^{*}$-space, which meets the convex set in an interval; the unknown aspect ratio restricts it to $\operatorname{diag}(X,Y,1)$, a plane, which meets it in a convex region. 

The theorem has two further consequences. First, its nonemptiness clause states that \emph{every} junction of three distinct concurrent lines is the image of a right-angle corner under some camera. The test $F=\emptyset$ of \cref{thm:interval} rejects junctions only because the camera model there is restricted to square pixels, zero skew and a known principal point; for a general camera, no junction is ever excluded. 

Second, because each junction's cell is known from the junction alone, self-calibration from junctions is a fully specified convex feasibility problem; \cref{sec:program} solves it.

\begin{figure}[t]
\centering
\includegraphics[width=0.9\linewidth]{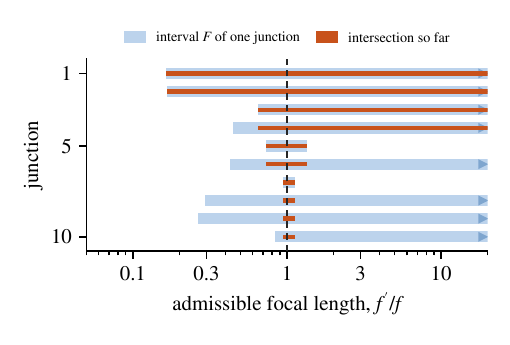}
\caption{The admissible intervals $F$ of \cref{thm:interval} for ten noiseless junctions, each from its own random frame, at a $60^\circ$ field of view (light bars, one row per junction; an arrow marks an interval unbounded above), and the intersection of the intervals so far (thin dark bars), on a logarithmic axis of $f'/f$. Every interval contains the true focal length (dashed line). Only two are bounded above, yet the intersection closes on the true value from both sides, here to $[0.94f,\,1.13f]$}
\label{fig:intervals}
\end{figure}

\subsection{A single unknown focal length}\label{sec:admissible}

Suppose now that pixels are square, skew is zero and the principal point is known, so that $K=\operatorname{diag}(f,f,1)$ has the single unknown $f$. What does one corner say about it? It cannot determine $f$, since four unknowns, $f$ and the three ratios $k_i$, face three orthogonality equations. Not every $f$ is admissible, however, and the set of admissible values is exactly what one corner says about $f$.

With the convention of \cref{sec:setup}, $\vr=(u/f,v/f,1)$ and $\vn_i=(fa_i,fb_i,e_i)$. Writing $x=f^2$ and $\rho=\sqrt{u^2+v^2}$ for the distance from the principal point to the junction, the two factors of \eqref{eq:gfactor} become elementary: $\|\vr\|^2=(x+\rho^2)/x$ and $\vn_i\cdot\vn_j=A_{ij}x+e_ie_j$ with $A_{ij}=a_ia_j+b_ib_j$. Each $\vn_i\cdot\vn_j$ is linear in $f^2$, which is why the condition is so simple; the threshold contributed by a pair is the focal length at which its two interpretation planes become perpendicular. Hence
\begin{equation}
x\,g_{ij}=(x+\rho^2)\,(A_{ij}x+e_ie_j).
\label{eq:lin}
\end{equation}
The factor $x+\rho^2$ is positive for every $x>0$ and common to all three pairs, so it cancels from the sign condition, which reduces to a product of three \emph{linear} functions of $f^2$:
\begin{equation}
L(x)=\prod_{(i,j)}\bigl(A_{ij}x+e_ie_j\bigr)<0,
\label{eq:Lprod}
\end{equation}
the product running over $(1,2),(2,3),(3,1)$. Each factor has at most one positive root, $x_{ij}=-e_ie_j/A_{ij}$, so the admissible set has at most three endpoints and can be read off a sign chart.

\paragraph{Angle parameterization.} A change of variables makes every sign in what follows geometric. Recall $\rho=\sqrt{u^2+v^2}$, and let $\psi_i\in(-\pi,\pi]$ be the signed angle at the corner from the direction $(u,v)/\rho$, pointing away from the principal point, to the observed half-edge direction $(b_i,-a_i)$. Since the line passes through $(u,v)$, $e_i=-(a_iu+b_iv)$, and $(a_i,b_i)$ is that direction under the quarter turn $(x,y)\mapsto(-y,x)$, so
\begin{equation}
e_i=\rho\sin\psi_i,\qquad A_{ij}=\cos(\psi_i-\psi_j),
\label{eq:psi}
\end{equation}
the second identity saying that $A_{ij}$ is the cosine of the raw image angle between half-edges $i$ and $j$. Substituting into \eqref{eq:Lprod} and writing $\mu=f^2/\rho^2$ for the squared focal length in units of $\rho^2$, each factor separates into a prefactor and a linear term:
\begin{equation}
\begin{aligned}
A_{ij}f^2+e_ie_j&=\rho^2\sin\psi_i\sin\psi_j\,\bigl(\lambda_{ij}\,\mu+1\bigr),\\
\lambda_{ij}&=1+\cot\psi_i\cot\psi_j=\frac{\cos(\psi_i-\psi_j)}{\sin\psi_i\sin\psi_j}.
\end{aligned}
\label{eq:lambda}
\end{equation}
Over the three pairs the prefactors multiply to $(\sin\psi_1\sin\psi_2\sin\psi_3)^2$, which is positive when no edge line passes through the principal point, the case treated next. They therefore drop out of the sign condition, which becomes
\begin{equation}
\prod_{(i,j)}\bigl(\lambda_{ij}\,\mu+1\bigr)<0,\qquad \mu=f^2/\rho^2.
\label{eq:mu}
\end{equation}
This form shows two facts. The condition involves the corner only through the three angles $\psi_i$ and the single scale $\rho$, so it constrains $f$ only through $f/\rho$; and each factor is positive at $\mu=0$, so everything depends on which of the $\lambda_{ij}$ are negative.

\begin{theorem}[The admissible set of one junction]\label{thm:interval}\label{prop:bound}
Let $F=\{f>0:\,L(f^2)<0\}$ be the admissible set of one junction.
\begin{enumerate}
\item $L(0)=(e_1e_2e_3)^2\ge 0$, with equality if and only if an edge line passes through the principal point. If $L(0)>0$, then $F$ excludes a neighborhood of $0$, so that $\fmin:=\inf F>0$ whenever $F\neq\emptyset$; otherwise $F$ may reach down to $0$ (\cref{rem:exceptional}).
\item As $x\to\infty$ the sign of $L$ is the sign of $A_{12}A_{23}A_{31}$. If that product is nonzero, $F$ is unbounded above if and only if the raw image angles satisfy the orthographic fork-or-arrow rule.
\item The factor $(i,j)$ has a positive root if and only if $A_{ij}e_ie_j<0$, and on $F$ the sign pattern of $(g_{12},g_{23},g_{31})$ is constant.
\item If no edge line passes through the principal point, let
\[
\begin{aligned}
S&=\{(i,j):\lambda_{ij}<0\}\\&=\{(i,j):\cot\psi_i\cot\psi_j<-1\},\\
f_{ij}&=\rho/\sqrt{-\lambda_{ij}}\qquad\text{for }(i,j)\in S .
\end{aligned}
\]
Then $|S|\le2$, and
\[
F=\begin{cases}
\emptyset, & |S|=0,\\[2pt]
(f_{ij},\,\infty), & S=\{(i,j)\},\\[2pt]
\bigl(\min_S f_{ij},\ \max_S f_{ij}\bigr), & |S|=2,
\end{cases}
\]
a single interval with closed-form endpoints.
\end{enumerate}
\end{theorem}
\begin{proof}
(1) Set $x=0$ in \eqref{eq:Lprod}: $L(0)=\prod e_ie_j=(e_1e_2e_3)^2$, which vanishes only if some $e_i=0$, that is, $\vl_i$ passes through the principal point ($u=v=0$ forces all three). Continuity then gives $L>0$ on some $[0,\delta)$ in $x=f^2$, so $F\cap(0,\sqrt\delta)=\emptyset$ and $\inf F\ge\sqrt\delta$ when $F\neq\emptyset$. (2) Divide by $x^3$ and let $x\to\infty$. With $a_i^2+b_i^2=1$, $A_{ij}$ is the cosine of the raw angle between half-edges $i$ and $j$, so $L<0$ for large $x$, that is, $F$ is unbounded above, exactly when an odd number of the raw angles are obtuse, which is the fork-or-arrow rule. (3) $A_{ij}x+e_ie_j$ has its root at $-e_ie_j/A_{ij}$, positive exactly when $A_{ij}e_ie_j<0$. For the pattern, $g_{12}g_{23}g_{31}$ is a positive multiple of $L(x)$ by \eqref{eq:lin}, so a sign change of any $g_{ij}$ within $F$ would be a zero of $L$ inside $F$.

(4) Write $\tau_i=\cot\psi_i$, so $\lambda_{ij}=1+\tau_i\tau_j$ and $(i,j)\in S$ iff $\tau_i\tau_j<-1$. Since $(\tau_1\tau_2)(\tau_2\tau_3)(\tau_3\tau_1)=(\tau_1\tau_2\tau_3)^2\ge0$, the three products cannot all be negative, so $|S|\le2$. Each factor of \eqref{eq:mu} equals $1$ at $\mu=0$; it is negative for some $\mu>0$ exactly when $\lambda_{ij}<0$, and then for $f>f_{ij}$, while a factor with $\lambda_{ij}\ge0$ satisfies $\lambda_{ij}\mu+1\ge1$ throughout. The product is negative exactly where an odd number of factors are. If $|S|=0$, every factor is positive and $F=\emptyset$. If $S=\{(i,j)\}$, that factor is negative precisely for $f>f_{ij}$ and the others are positive. If $|S|=2$, the two pairs share an index, say $(i,j)$ and $(i,k)$, and their thresholds are distinct, since $\tau_i\neq0$ and $\lambda_{ij}=\lambda_{ik}$ would give $\cot\psi_j=\cot\psi_k$, putting edges $j$ and $k$ on one image line; writing $f_a<f_b$ for them, both factors are positive below $f_a$, exactly one is negative between, and both are negative above $f_b$, so $F=(f_a,f_b)$.
\end{proof}

We adopt the convention $\inf\emptyset=+\infty$. That $F$ can be empty reflects the restricted camera model: by \cref{thm:convex}, every junction is admitted by some general camera. That $F$ is a single interval also follows from \cref{thm:convex}, restricted to these cameras, and parts (2) and (4) agree, since $\prod A_{ij}$ and $\prod\lambda_{ij}$ differ by a square. Part (2) has a simple reason: as $f'$ grows, the rotation to the canonical view shrinks and the canonical angles tend to the raw ones, so large focal lengths are admitted exactly when the raw angles pass the fork-or-arrow rule, the rule of \citet{kanatani1986} for orthographic images. Since only $f/\rho$ enters, the bounds scale with the junction's distance from the principal point. \Cref{fig:intervals} shows the intervals of ten junctions and how they combine.

\begin{corollary}[Similarity invariance]\label{cor:similar}
$F/\rho$ depends only on $(\psi_1,\psi_2,\psi_3)$. Two junctions with the same three angles, at different distances from the principal point, give admissible sets that are scalings of one another by the ratio of their distances. In particular every threshold is proportional to $\rho$, and the bounds degenerate as $\rho\to0$; the relative precision of a junction depends only on its angles $\psi_i$.
\end{corollary}
\begin{proof}
By \eqref{eq:mu}, the condition involves $f$ only through $\mu=f^2/\rho^2$, and the corner only through the angles $\psi_i$.
\end{proof}

The threshold also has a direct expression in the raw image angles. Since $-\lambda_{ij}=-\cos(\psi_i-\psi_j)/(\sin\psi_i\sin\psi_j)$,
\begin{equation}
f_{ij}^2=\frac{-\rho^2\sin\psi_i\sin\psi_j}{\cos(\psi_i-\psi_j)},
\label{eq:thresh}
\end{equation}
so a pair contributes a threshold exactly when $\sin\psi_i\sin\psi_j$ and $\cos(\psi_i-\psi_j)$ have opposite signs. Both signs can be read in the image: the first says whether the two half-edges leave the corner on the same side of the line through the principal point and the vertex, and the second is the cosine of their raw image angle.

\Cref{thm:interval} also supplies a test at no cost: $F=\emptyset$ means that no focal length makes the junction a right-angle corner for a camera of this model. It is not a junction detector, because junctions of a frame pass it wherever they lie (\cref{cor:framepass}); \OR{} (Sect.~S12) measures how often false junctions pass it.

\begin{corollary}[Junctions of a frame pass the test]\label{cor:framepass}
Let three image lines pass through a point $\vc$, one through each vanishing point of three mutually orthogonal scene directions $\vd_1,\vd_2,\vd_3$, with $\vc$ on none of the lines that join two of the vanishing points. Then the true camera admits the junction; in particular, when only the focal length is unknown, the true $f$ lies in $F$ and $F\neq\emptyset$.
\end{corollary}
\begin{proof}
The line through $\vc$ and the vanishing point $K\vd_k$ back-projects to a plane that contains the viewing ray $\vr$ and the direction $\vd_k$, so $\vd_k$ lies in its interpretation plane. The three lines are distinct, because $\vc$ lies on no line joining two vanishing points, and no $\vd_k$ lies along the ray, because $\vc$ is not a vanishing point. The directions are mutually orthogonal, so the true camera admits the junction by \cref{def:admissible}.
\end{proof}

\begin{remark}[The exceptional configurations]\label{rem:exceptional}
The excluded configurations have measure zero, but the bound degrades as they are approached. At the principal point every $e_i=0$, so $L(x)=A_{12}A_{23}A_{31}x^3$ has constant sign and $F$ is empty or all of $(0,\infty)$. With exactly one edge line through the principal point, say $e_1=0$, $L(x)=A_{12}A_{31}x^2(A_{23}x+e_2e_3)$ has at most one positive threshold, and the sign of $A_{12}A_{31}$ decides which side of it is admissible, so the lower bound may be lost. For instance, $F=(0,f_{23})$, an upper bound only, when $A_{12}A_{31}>0$, $A_{23}>0$ and $e_2e_3<0$. Two such edge lines put the corner at the principal point or two edges on one image line. In every case $F$ is still an interval.
\end{remark}

Away from the configurations of \cref{rem:exceptional}, then, one corner yields a lower bound on $f$, and also an upper bound exactly when the junction would be rejected under orthography, as happens near the border of a wide-angle image (\cref{sec:e2}).

\subsection{Other camera models: aspect ratio and principal point}\label{sec:slices}

For the cameras $\omega^{*}=\operatorname{diag}(f^2,f^2,1)$, with a single unknown focal length, the three forms of \cref{prop:general} are linear in the one unknown $f^2$. Two other camera models deserve mention.

\paragraph{Unknown aspect ratio.} With $\omega^{*}=\operatorname{diag}(X,Y,1)$, where $X=f_x^2$ and $Y=f_y^2$, each form $\vl_i^{\top}\omega^{*}\vl_j=a_ia_jX+b_ib_jY+e_ie_j$ is affine in $(X,Y)$.

\begin{proposition}[Unknown aspect ratio]\label{prop:aspect}
For $\omega^{*}=\operatorname{diag}(X,Y,1)$, the cameras that admit a junction are the points $(X,Y)$ of the open positive quadrant at which each form $a_ia_jX+b_ib_jY+e_ie_j$ has the sign of $\Delta_{ki}\Delta_{kj}$, for $\{i,j,k\}=\{1,2,3\}$. They form an open convex polygon, possibly empty or unbounded, cut from the quadrant by three lines. The aspect ratios $f_y/f_x$ of these cameras form an open interval.
\end{proposition}
\begin{proof}
For $X,Y>0$ the form $\omega^{*}$ is positive-definite, and \cref{thm:convex} with the forms above gives the three half-planes. The map $(X,Y)\mapsto\sqrt{Y/X}$ is continuous and open on the open quadrant, so it carries the open convex polygon, which is connected, onto an open interval.
\end{proof}

The products of the three coefficients of $X$, of $Y$ and of the constant terms are the squares $(a_1a_2a_3)^2$, $(b_1b_2b_3)^2$ and $(e_1e_2e_3)^2$. The polygon therefore stays away from the origin and does not extend to infinity along either axis, although it may be unbounded in between: this is the anisotropic form of \cref{prop:bound}(1) and (2). A single junction thus bounds the aspect ratio as well as the focal length, and it detects non-square pixels whenever its interval of aspect ratios excludes $1$.

\paragraph{Unknown principal point.} Writing $\tilde{\mathbf p}=(p_x,p_y,1)$ for the principal point,
\[
\begin{aligned}
\omega^{*}&=f^2\operatorname{diag}(1,1,0)+\tilde{\mathbf p}\tilde{\mathbf p}^{\top},\\
\vl_i^{\top}\omega^{*}\vl_j&=f^2A_{ij}+d_id_j,\qquad d_i=\vl_i\cdot\tilde{\mathbf p},
\end{aligned}
\]
where $d_i$ is the signed distance from the principal point to the edge line $\vl_i$. The form is linear in $f^2$ for a fixed principal point and quadratic in the principal point for a fixed $f$, so these cameras do not form a line or a plane in $\omega^{*}$-space and convexity does not carry over. The admitted set is nonetheless explicit.

\begin{proposition}[Unknown principal point]\label{prop:pp}
Let the focal length $f$ be known. A principal point is admitted by the junction if and only if $\prod_{(i,j)}\bigl(f^2A_{ij}+d_id_j\bigr)<0$. Each factor vanishes on a hyperbola centered at the junction, whose asymptotes are the edge lines $\vl_i$ and $\vl_j$, or on those two lines if $A_{ij}=0$. The principal points admitted for the focal length $f$ are the image of those admitted for $f=1$ under the homothety of ratio $f$ about the junction.
\end{proposition}
\begin{proof}
The first statement is \cref{prop:general} with the form above. Write $\mathbf c$ for the junction and $\mathbf p$ for the principal point, in pixel coordinates. The edge lines pass through $\mathbf c$, so each $d_i$ is a linear function of $\mathbf p-\mathbf c$, and $d_id_j=-f^2A_{ij}$ is a hyperbola with center $\mathbf c$ and asymptotes $d_i=0$ and $d_j=0$ when $A_{ij}\neq0$, and the pair of lines when $A_{ij}=0$. Substituting $\mathbf p=\mathbf c+f\mathbf q$ multiplies every $d_i$ by $f$, so each factor becomes $f^2\bigl(A_{ij}+d_i(\mathbf c+\mathbf q)\,d_j(\mathbf c+\mathbf q)\bigr)$, and the verdict at $\mathbf c+f\mathbf q$ for the focal length $f$ is the verdict at $\mathbf c+\mathbf q$ for $f=1$.
\end{proof}

\begin{figure}[t]
\centering
\includegraphics[width=0.85\linewidth]{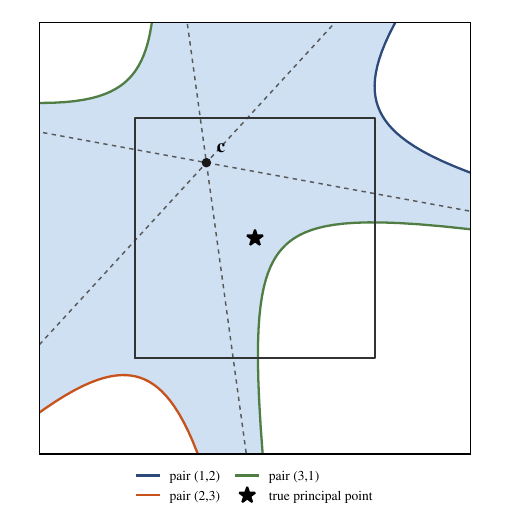}
\caption{The principal points admitted by one junction, for the true focal length of a $90^\circ$ camera (shaded). The region is bounded by three hyperbolas, one for each pair of edges, centered at the junction $\vc$ and with the pair's edge lines (dashed) as asymptotes. The square is the image, and the star is the true principal point. The region is not convex, and changing the focal length scales it about $\vc$ (\cref{prop:pp})}
\label{fig:ppregion}
\end{figure}

\Cref{fig:ppregion} shows the three hyperbolas for one junction.

\paragraph{Both unknown.} With the focal length and the principal point both unknown, each condition $f^2A_{ij}+d_id_j=0$ is homogeneous of degree two in $(\mathbf p-\mathbf c,f)$: a quadric cone in $(p_x,p_y,f)$-space with vertex at $(\mathbf c,0)$. The admitted set is a union of cells of three such cones, invariant under scaling about $(\mathbf c,0)$; \cref{thm:fpp} treats many junctions in this model.

\paragraph{Skew.} \Cref{thm:convex} places no restriction on the camera: skew, aspect ratio and principal point are all free there. Only the restricted models of this section and of \cref{thm:interval} assume zero skew.

\paragraph{Known orientation.} If the three edges of a corner are known either all to approach the camera or all to recede, as for a room corner seen from inside, the admissible set keeps only its fork-pattern part. This restriction never tightens the bound, but it does provide a check on the detection (\OR, Sect.~S4).

\subsection{Certified bounds and model error}\label{sec:noise}

Real junctions come from noisy line fits, so their thresholds move. \OR{} (Sect.~S7) gives the first-order spread of the thresholds and the probability that the true focal length leaves the measured set. Here we give a deterministic guarantee instead: whatever the source of error, it suffices to bound how far each measured angle can be from the true one.

\begin{proposition}[Certified bounds]\label{prop:certified}
Let the position of a junction be exact and its measured angles satisfy $|\hat\psi_i-\psi_i|\le\varphi_i$, and let $\hat A_{ij}$ and $\hat e_i$ be computed from the $\hat\psi_i$ by \eqref{eq:psi}. For each pair let
\[
\begin{aligned}
\mathcal U_{ij}=\bigl\{x>0:\ &|\hat A_{ij}x+\hat e_i\hat e_j|\\
&\le(\varphi_i^2+\varphi_j^2)^{1/2}\,(x\sqrt2+\rho^2)\bigr\},
\end{aligned}
\]
which is an interval or empty, and $\mathcal U=\bigcup_{(i,j)}\mathcal U_{ij}$. Then, as sets of $x=f^2$, the measured and true admissible sets satisfy
\[
\hat F\setminus\mathcal U\;\subseteq\;F\;\subseteq\;\hat F\cup\mathcal U ;
\]
in particular, the square of the true focal length lies in $\hat F\cup\mathcal U$.
\end{proposition}
\begin{proof}
The gradient of the form $\cos(\psi_i-\psi_j)\,x+\rho^2\sin\psi_i\sin\psi_j$ in $(\psi_i,\psi_j)$ is $x\sin(\psi_i-\psi_j)\,(-1,1)+\rho^2(\cos\psi_i\sin\psi_j,\ \sin\psi_i\cos\psi_j)$. Its first term has norm at most $x\sqrt2$, and its second at most $\rho^2$, since $\cos^2\psi_i\sin^2\psi_j+\sin^2\psi_i\cos^2\psi_j\le1$; the form is therefore Lipschitz in $(\psi_i,\psi_j)$ with constant $x\sqrt2+\rho^2$, so at each $x$ the true and measured forms differ by at most $(\varphi_i^2+\varphi_j^2)^{1/2}\,(x\sqrt2+\rho^2)$. Outside $\mathcal U_{ij}$ the measured form exceeds this in absolute value, so the two forms have the same sign; outside $\mathcal U$ all three pairs do, and $x\in F$ exactly when $x\in\hat F$. Both inclusions follow, and the true focal length lies in $F$ because the junction is the image of a right-angle corner. Finally, $\mathcal U_{ij}$ is the set where the two affine functions $\hat A_{ij}x+\hat e_i\hat e_j\pm(\varphi_i^2+\varphi_j^2)^{1/2}\,(x\sqrt2+\rho^2)$ of $x$ do not have the same strict sign, the intersection of two half-lines.
\end{proof}

Each $\mathcal U_{ij}$ is a neighborhood of the measured threshold $\hat x_{ij}$, of width proportional to $(\varphi_i^2+\varphi_j^2)^{1/2}$ unless $\hat A_{ij}$ is small. The certified set $\hat F\cup\mathcal U$ is therefore the measured interval widened by these neighborhoods, and it contains the true focal length for every perturbation of the angles within the bounds $\varphi_i$, not only with high probability. If each endpoint of edge $i$, a segment of length $\ell_{\mathrm{seg}}$, is known to within $\varepsilon$ pixels, its direction is known to within $\varphi_i=\arcsin(2\varepsilon/\ell_{\mathrm{seg}})$, for $\ell_{\mathrm{seg}}>2\varepsilon$, so edges of different lengths carry different bounds. An error in the vertex position enters through $\rho$ and the $e_i$, and the proposition does not cover it.

\paragraph{Model error.} Corners whose angles depart from $90^\circ$, and radial lens distortion, are analyzed in \OR{} (Sect.~S6). A right angle that is off by a few degrees keeps the true camera admitted unless an edge is nearly perpendicular to the viewing ray, that is, unless the vertex is near an edge-on line; and radial distortion acts on each junction by an explicit transformation, which with \cref{prop:certified} gives a guarantee for fitted segments. In simulation at $60^\circ$, edges tilted by up to $2^\circ$ and $5^\circ$ left the true $f$ in $F$ for $92\%$ and $84\%$ of junctions, and distortion that displaces the image corners by $2\%$ left it in $F$ for $96\%$.

\section{Many junctions}\label{sec:many}

A single junction bounds the camera; this section asks what many junctions determine. We treat, in turn, the cameras that one frame rejects (\cref{sec:oneframe}), what several frames determine (\cref{sec:frames}), the focal length alone (\cref{sec:identifiability}) and together with the principal point (\cref{sec:fpp}), and the convex program that combines the constraints (\cref{sec:program}), which closes with a remark on noise.

\subsection{What the junctions of one frame reject}\label{sec:oneframe}

\paragraph{Do many junctions calibrate?} One might count three linear forms per junction against the five degrees of freedom of $\omega^{*}$ and conclude that two junctions calibrate a camera. They do not, because \cref{thm:convex} yields inequalities, not equations. A junction does not place $\omega^{*}$ on a hypersurface; it confines it to a cone, and no finite number of junctions solves for it. What many junctions can do is cut the cone down, and the question is: to what? For the junctions of one Manhattan frame, the answer is exact.

\begin{figure*}[t]
\centering
\begin{tikzpicture}[scale=0.39,font=\small]
\tikzset{img/.style={fill=black!10,draw=black!55},vp/.style={circle,fill,inner sep=1.3pt},
         pass/.style={thick},miss/.style={dashed,black!60},edge/.style={very thick,black!80,line cap=round},
         ttl/.style={font=\footnotesize,align=center,text width=3.8cm}}
\begin{scope}
  \draw[img] (-1.6,-1.2) rectangle (1.6,1.2);
  \coordinate (a1) at (-4.4,0.35); \coordinate (a2) at (4.6,0.05); \coordinate (a3) at (0.2,5.0);
  \draw[pass] ($(a1)!-0.06!(a2)$) -- ($(a2)!-0.06!(a1)$);
  \draw[miss] (a1) -- (a3); \draw[miss] (a2) -- (a3);
  \node[vp,label=below:{$\vv_1$}] at (a1) {}; \node[vp,label=below:{$\vv_2$}] at (a2) {}; \node[vp,label=right:{$\vv_3$}] at (a3) {};
  \node[above] at (3.2,0.1) {$\ell_3$}; \node[left] at (-2.3,2.7) {$\ell_2$}; \node[right] at (2.5,2.7) {$\ell_1$};
  \coordinate (c) at (0.55,0.62);
  \draw[edge] (c) -- ($(c)!0.7cm!(a1)$); \draw[edge] (c) -- ($(c)!0.7cm!(a2)$); \draw[edge] (c) -- ($(c)!0.7cm!(a3)$);
  \fill (c) circle (1.6pt);
  \node[ttl,anchor=south] at (0,6.0) {(a) only the horizon $\ell_3$ passes through the image};
\end{scope}
\begin{scope}[shift={(11,0)}]
  \draw[img] (-1.6,-1.2) rectangle (1.6,1.2);
  \coordinate (b1) at (-3.2,-2.6); \coordinate (b2) at (3.6,-2.2); \coordinate (b3) at (0.0,3.0);
  \draw[miss] (b1) -- (b2);
  \draw[pass] ($(b1)!-0.08!(b3)$) -- ($(b3)!-0.08!(b1)$); \draw[pass] ($(b2)!-0.08!(b3)$) -- ($(b3)!-0.08!(b2)$);
  \node[vp,label=left:{$\vv_1$}] at (b1) {}; \node[vp,label=right:{$\vv_2$}] at (b2) {}; \node[vp,label={[xshift=2pt]right:{$\vv_3$}}] at (b3) {};
  \node[below] at (0.2,-2.45) {$\ell_3$}; \node[left] at (-2.0,0.1) {$\ell_2$}; \node[right] at (2.1,0.1) {$\ell_1$};
  \coordinate (c) at (1.12,0.78);
  \draw[edge] (c) -- ($(c)!0.7cm!(b1)$); \draw[edge] (c) -- ($(c)!0.7cm!(b2)$); \draw[edge] (c) -- ($(c)!0.7cm!(b3)$);
  \fill (c) circle (1.6pt);
  \node[ttl,anchor=south] at (0,6.0) {(b) two edge-on lines, $\ell_1$ and $\ell_2$, pass through the image};
\end{scope}
\begin{scope}[shift={(22,0)}]
  \draw[img] (-1.6,-1.2) rectangle (1.6,1.2);
  \coordinate (c1) at (-4.0,4.0); \coordinate (c2) at (5.0,-0.5); \coordinate (c3) at (0.0,-5.5);
  \draw[pass] (c1) -- (c2); \draw[miss] (c1) -- (c3); \draw[miss] (c2) -- (c3);
  \draw[fill=white] (1.6,1.2) circle (3pt);
  \node[vp,label=left:{$\vv_1$}] at (c1) {}; \node[vp,label=right:{$\vv_2$}] at (c2) {}; \node[vp,label=right:{$\vv_3$}] at (c3) {};
  \node[above right=-1pt] at (1.6,1.25) {$\vc_0$};
  \node[above] at (-2.0,3.05) {$\ell_3$}; \node[left] at (-2.4,-1.4) {$\ell_2$}; \node[right] at (2.6,-2.9) {$\ell_1$};
  \node[ttl,anchor=south] at (0,6.0) {(c) $\ell_3$ touches the image only at its corner $\vc_0$};
\end{scope}
\end{tikzpicture}
\caption{The three vanishing points of a rectilinear structure, the edge-on lines that join them, and the image (gray). The edge-on line $\ell_k$ of axis $k$ passes through the other two vanishing points; solid lines meet the image and dashed lines miss it. Near a line that passes through the image, a junction's two edges toward that line's vanishing points nearly line up, the face they span being seen edge-on, as drawn in (a) and (b); such junctions reject every camera under which the opposite vanishing point is not perpendicular to the other two (\cref{thm:converse}). (a) A nearly level camera, for which few junctions are bounded above. (b) An oblique view, which forces all three perpendicularities. (c) The touching case, which forces the single linear condition \eqref{eq:touch}}
\label{fig:vlines}
\end{figure*}
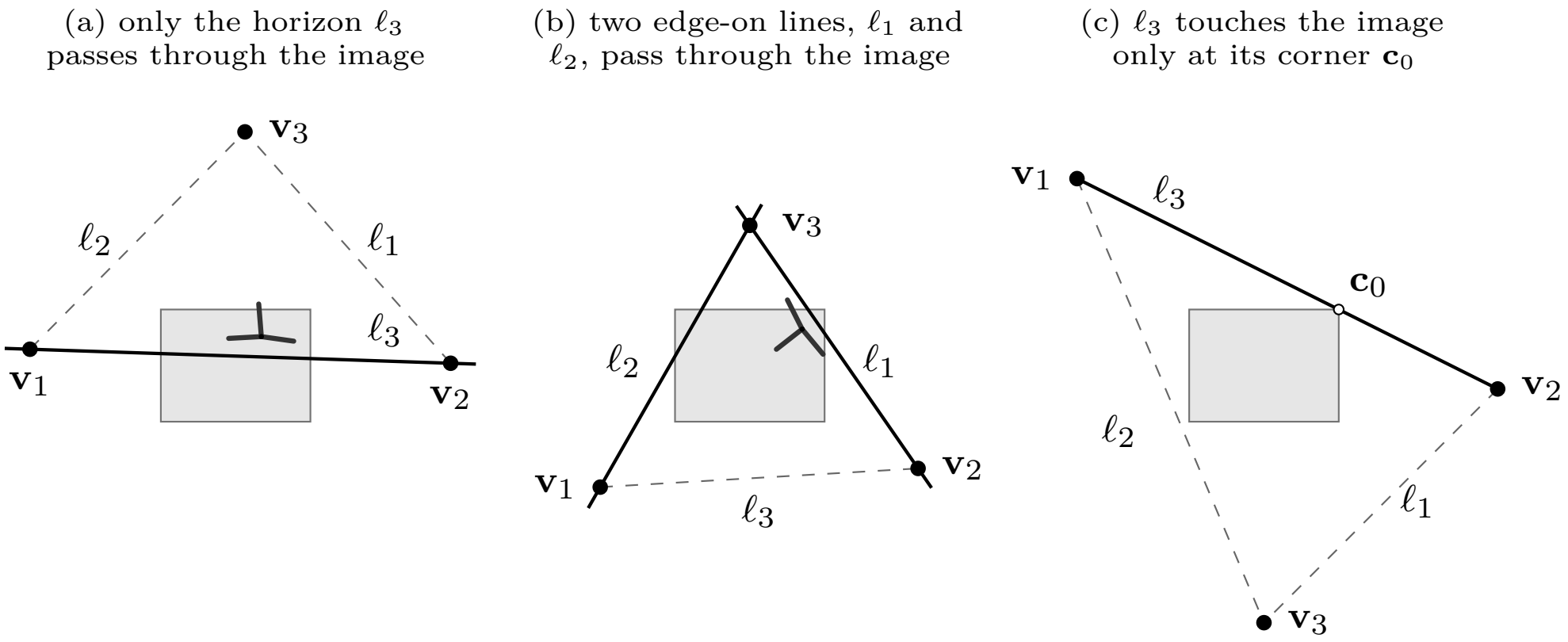

A \emph{Manhattan frame} is a triple of mutually orthogonal scene directions, its axes, and a junction \emph{of the frame} is the image of a corner whose edges lie along them. An image shows one scene, and usually one such frame. Let $\vv_1,\vv_2,\vv_3$ be its vanishing points, as vectors of $\mathbb R^3$ with any fixed scaling, and $V=[\vv_1\ \vv_2\ \vv_3]$. Every line of direction $k$, in every junction of the frame, passes through $\vv_k$, so the junction of the frame with vertex $\vc$ consists of the lines $\vl_k=\vc\times\vv_k$. The natural coordinates are therefore those of the vanishing points, and we write the candidate camera and the vertex in their basis,
\begin{equation}
\omega^{*}=VNV^{\top},\qquad \vc=V\boldsymbol\zeta ,
\label{eq:vpbasis}
\end{equation}
so that $N$ is symmetric and positive-definite and $\boldsymbol\zeta$ gives homogeneous coordinates of the vertex in the triangle of vanishing points. Two facts follow at once. First, the line $\zeta_k=0$ is the \emph{vanishing line} $\ell_k$ through the other two vanishing points. Second, since $V^{\top}\omega V=N^{-1}$ for $\omega=(\omega^{*})^{-1}$, the vanishing points are mutually orthogonal under the candidate camera, $\vv_a^{\top}\omega\vv_b=0$ for $a\neq b$, exactly when $N$ is diagonal. At the true camera $N$ is diagonal, and $N=I$ when the vanishing points are scaled as $\vv_k=K\vd_k$, that is, $V=KR$ with $R=[\vd_1\ \vd_2\ \vd_3]$ the frame. Then $\boldsymbol\zeta=R^{\top}K^{-1}\vc$ has entries $\zeta_k=\vr\cdot\vd_k$, so $\ell_k$ is also the locus where edge $k$ is perpendicular to the viewing ray, the edge-on line of axis $k$. For the vertical axis it is the horizon (\cref{fig:vlines}). The vanishing points serve here only as coordinates for the analysis: the junction constraints never use them, and the results describe what those constraints achieve whether or not any vanishing point could be located.

\begin{lemma}[A junction of the frame]\label{lem:frame}
In the notation of \eqref{eq:vpbasis}, the junction of the frame with vertex $\vc=V\boldsymbol\zeta$, whose lines are $\vl_k=\vc\times\vv_k$, with $\zeta_1\zeta_2\zeta_3\neq0$, is admissible under $\omega^{*}=VNV^{\top}$, with $V$ invertible and $N$ positive-definite, if and only if, for each $a$, with $\{a,p,q\}=\{1,2,3\}$,
\begin{equation}
\nu_a=N_{aa}-\frac{\zeta_a}{\zeta_p}N_{ap}-\frac{\zeta_a}{\zeta_q}N_{aq}+\frac{\zeta_a^2}{\zeta_p\zeta_q}N_{pq}>0 .
\label{eq:nuframe}
\end{equation}
\end{lemma}
\begin{proof}
Here $\vm_k=\vc\times(\vc\times\vv_k)=(\vc\cdot\vv_k)\,\vc-\|\vc\|^2\vv_k$, which acts on $\vc^{\perp}$ as $-\|\vc\|^2\vv_k$. By \cref{thm:convex} the junction is therefore admissible exactly when $\boldsymbol\chi^{\top}\omega^{*}\boldsymbol\chi=\sum_k\nu_k(\vv_k\cdot\boldsymbol\chi)^2$ on $\vc^{\perp}$ with every $\nu_k>0$, these $\nu_k$ being those of \eqref{eq:nuk} multiplied by $\|\vc\|^4$. Put $\mathbf y=V^{\top}\boldsymbol\chi$. Then $\boldsymbol\chi^{\top}\omega^{*}\boldsymbol\chi=\mathbf y^{\top}N\mathbf y$ and $\vc\cdot\boldsymbol\chi=\boldsymbol\zeta\cdot\mathbf y$, so the condition is that $\mathbf y^{\top}N\mathbf y=\mathbf y^{\top}\operatorname{diag}(\nu)\,\mathbf y$ on the plane $\boldsymbol\zeta\cdot\mathbf y=0$. A symmetric matrix vanishes as a quadratic form on that plane exactly when it is $\boldsymbol\zeta\boldsymbol\eta^{\top}+\boldsymbol\eta\boldsymbol\zeta^{\top}$ for some $\boldsymbol\eta$, so $N=\operatorname{diag}(\nu)+\boldsymbol\zeta\boldsymbol\eta^{\top}+\boldsymbol\eta\boldsymbol\zeta^{\top}$. Its off-diagonal entries read $N_{ab}=\zeta_a\eta_b+\zeta_b\eta_a$, three equations with the solution $2\eta_a/\zeta_a=N_{ap}/(\zeta_a\zeta_p)+N_{aq}/(\zeta_a\zeta_q)-N_{pq}/(\zeta_p\zeta_q)$, and its diagonal entries then give $\nu_a=N_{aa}-2\zeta_a\eta_a$, which is \eqref{eq:nuframe}.
\end{proof}

\begin{theorem}[Which cameras one frame's junctions reject]\label{thm:converse}
Let the image $\mathcal I$ be a closed convex polygon with nonempty interior and no vanishing point on its boundary, and consider the junctions of one Manhattan frame with vertices in $\mathcal I$, off the vanishing lines. For an axis $k$, let $i$ and $j$ be the other two.
\begin{enumerate}
\item If $\vc_0\in\ell_k\cap\mathcal I$ is not a vanishing point, every positive-definite $\omega^{*}$ that violates
\begin{equation}
\zeta_j(\vc_0)\,N_{ik}=\zeta_i(\vc_0)\,N_{jk}
\label{eq:touch}
\end{equation}
is rejected by every junction whose vertex is close enough to $\vc_0$, how close depending on $\omega^{*}$. Hence, if $\ell_k$ meets $\mathcal I$ in a segment, every camera with $(N_{ik},N_{jk})\neq0$ is rejected by some junction; if it meets $\mathcal I$ only at a corner $\vc_0$, every camera that violates \eqref{eq:touch} is.
\item There is a constant $C$, depending only on the frame and on $\mathcal I$, such that every $\omega^{*}$ that satisfies the conditions of part (1) and has $\max_{b\neq c}|N_{bc}|<\min_aN_{aa}/C$ passes every junction.
\item If two vanishing lines meet $\mathcal I$ in segments, the cameras that pass every junction are exactly those that make the three vanishing points mutually orthogonal,
\begin{equation}
\omega^{*}=\sum_{k=1}^{3}\theta_k\,\vv_k\vv_k^{\top},\qquad \theta_k>0 .
\label{eq:vpfamily}
\end{equation}
\end{enumerate}
\end{theorem}
\begin{proof}
(1) Let $\vc_0\in\ell_k\cap \mathcal I$ not be a vanishing point, so that $\zeta_i,\zeta_j\neq0$ there, and let the vertex tend to $\vc_0$ from the interior of $\mathcal I$, off the vanishing lines, so that $\zeta_k\to0$. Put $B=(\zeta_i/\zeta_j)N_{jk}-N_{ik}$. By \eqref{eq:nuframe},
\[
\begin{aligned}
\nu_i&=N_{ii}-\frac{\zeta_i}{\zeta_j}N_{ij}+\frac{\zeta_i}{\zeta_k}\,B,\\
\nu_j&=N_{jj}-\frac{\zeta_j}{\zeta_i}N_{ij}-\frac{\zeta_j^2}{\zeta_i\zeta_k}\,B .
\end{aligned}
\]
The terms without $B$ stay bounded, and the two coefficients of $B$ are unbounded with opposite signs, whichever side of $\ell_k$ the vertex is on. Unless $B\to0$, one of $\nu_i$, $\nu_j$ is therefore negative at junctions arbitrarily near $\vc_0$, and by \cref{lem:frame} those are not admitted. So $B(\vc_0)=0$, which is \eqref{eq:touch}. Along a segment the ratio $\zeta_i/\zeta_j$ is a projective coordinate and is not constant, so \eqref{eq:touch} at two of its points gives $N_{ik}=N_{jk}=0$. A line meets a convex polygon in a single point only at a corner of the polygon, which by hypothesis is not a vanishing point.

(2) By \cref{lem:frame}, a camera passes every junction exactly when every $\nu_a$ of \eqref{eq:nuframe} is positive throughout $\mathcal I$. Suppose first that a vanishing point lies in $\mathcal I$. It is interior, the two vanishing lines through it meet $\mathcal I$ in segments, and the conditions of part (1) make $N$ diagonal, so that $\nu_a=N_{aa}>0$. Otherwise, in the expression \eqref{eq:nuframe} for $\nu_a$ the terms that are singular near $\ell_p\cap \mathcal I$ combine, exactly as above, into a bounded multiple of $B/\zeta_p$, with $B$ the quantity belonging to the line $\ell_p$, the other two coordinates being bounded away from zero there. If $\ell_p$ meets $\mathcal I$ in a segment, $B$ vanishes identically. If it meets $\mathcal I$ at a corner $\vc_0$, then, in the notation of part (1) with $p$ in place of $k$, $B=(\zeta_i/\zeta_j-t_0)N_{jp}$, with $t_0$ the value of the ratio at $\vc_0$; and $\mathcal I$ lies in a wedge at $\vc_0$ whose sides make positive angles with the line, so that the distance from $\vc_0$ is at most a constant times $|\zeta_p|$ on $\mathcal I$ and $B/\zeta_p$ is bounded by a constant times $|N_{jp}|$. Hence there is a constant $C$, depending only on the frame and on $\mathcal I$, with $|\nu_a-N_{aa}|\le C\max_{b\neq c}|N_{bc}|$ throughout $\mathcal I$, so that $\nu_a\ge N_{aa}-C\max_{b\neq c}|N_{bc}|>0$ whenever $\max_{b\neq c}|N_{bc}|<\min_aN_{aa}/C$.

(3) Two segments annihilate all three off-diagonal entries, so $N$ is diagonal, with positive entries because it is positive-definite, and $\omega^{*}=VNV^{\top}$ is \eqref{eq:vpfamily} with $\theta_k=N_{kk}$. Conversely, for diagonal $N$, \eqref{eq:nuframe} gives $\nu_a=N_{aa}>0$ at every junction.
\end{proof}

In part (1), $(N_{ik},N_{jk})\neq0$ says that $\vv_k$ is not orthogonal to the other two vanishing points under the camera. A vanishing line that misses $\mathcal I$ rejects no camera on this account, although the inequalities \eqref{eq:nuframe}, which a camera must satisfy throughout $\mathcal I$ to pass every junction (\cref{lem:frame}), still bound it. If no edge-on line touches $\mathcal I$, the constant of part (2) can be taken as $C=\sup_{\vc\in\mathcal I}\max_a\sum|\kappa_{a,bc}|$, the sum running over the off-diagonal entries $N_{bc}$ that part (1) does not force to vanish, with $\kappa_{a,bc}$ their coefficients in \eqref{eq:nuframe}; each coefficient is bounded on $\mathcal I$, because its denominator vanishes only on an edge-on line that misses $\mathcal I$. Near the true camera, the cameras that pass every junction therefore form a family whose dimension is that of the linear space cut out by the conditions of part (1). A camera that violates \eqref{eq:touch} only slightly is rejected only by junctions close to $\ell_k$. The theorem guarantees that such junctions exist; \OR{} (Sect.~S9) estimates how many junctions placed at random it takes to meet one. The constant $C$ of part (2) grows as a vanishing line that misses $\mathcal I$, or a vanishing point outside $\mathcal I$, comes close to $\mathcal I$, and as the angle of a touching wedge shrinks. The global admitted set is not characterized in general, and the number of conditions forced is the rank of the system of part (1), not the sum of per-line counts.

In the commonest case, the admitted set can be described globally rather than only near the true camera.

\begin{proposition}[One edge-on line through the image]\label{prop:oneline}
Under the hypotheses of \cref{thm:converse}, let $\ell_k$ meet $\mathcal I$ in a segment and let $\ell_i$ and $\ell_j$ miss $\mathcal I$. Let $r_-\le r_+$ be the extremes over $\mathcal I$ of $r=\zeta_i/\zeta_j$, which have one sign and are attained at corners of $\mathcal I$, and let $u^{\star}$ be the extreme over $\mathcal I$ of $u=\zeta_k^2/(\zeta_i\zeta_j)$, which has the sign of $r$. Up to its boundary, the set of positive-definite $\omega^{*}=VNV^{\top}$ that admit every junction of the frame with vertex in $\mathcal I$ is
\begin{equation}
\begin{aligned}
&N_{ik}=N_{jk}=0,\qquad N_{kk}+u^{\star}N_{ij}>0,\\
&N_{ii}>r_{\pm}N_{ij},\qquad N_{jj}>N_{ij}/r_{\pm},
\end{aligned}
\label{eq:oneline}
\end{equation}
five linear inequalities on the one free off-diagonal entry and the diagonal.
\end{proposition}
\begin{proof}
By \cref{thm:converse}(1) every admitted camera has $N_{ik}=N_{jk}=0$, and then \eqref{eq:nuframe} reduces to $\nu_i=N_{ii}-rN_{ij}$, $\nu_j=N_{jj}-N_{ij}/r$ and $\nu_k=N_{kk}+uN_{ij}$. Since $\ell_i$ and $\ell_j$ miss $\mathcal I$, $\zeta_i$ and $\zeta_j$ do not vanish on $\mathcal I$, so $r$ and $u$ have constant signs there, $r$ is a ratio of two affine functions and attains its extremes at corners of the convex polygon $\mathcal I$, and $u$ ranges from values arbitrarily close to $0$, near $\ell_k$, to $u^{\star}$. For fixed $N$, $\nu_i$ is affine in $r$, $\nu_j$ in $1/r$ and $\nu_k$ in $u$, so each is positive over its range exactly when it is positive at the ends of the range, up to the boundary; at the end $u\to0$ the condition is $N_{kk}>0$, which positive-definiteness already gives.
\end{proof}

For the camera of \OR{} (Sect.~S11), tilted by $4^\circ$, \eqref{eq:oneline} with unit diagonal gives the free entry the interval $(-0.882,\,0.286)$, which the numerical interval $[-0.88,\,0.28]$ found there matches to its grid step. For the same camera the constant of \cref{thm:converse}(2) is $C=3.50$, and its sufficient condition $|N_{12}|<1/C=0.286$ reproduces the upper end of this interval exactly. With the frame oriented differently, a dense sample of $1.5$ million junctions confirms both ends of the formula to within its grid step of $0.005$.

\begin{proposition}[One frame cannot calibrate a general camera]\label{prop:oneframe}
Every camera of the family \eqref{eq:vpfamily} admits every junction of the frame, wherever its vertex lies. The family has three parameters, two once $\omega^{*}$ is normalized, so the junctions of one frame, however many, never determine a general camera.
\end{proposition}
\begin{proof}
For diagonal $N$, \eqref{eq:nuframe} gives $\nu_a=N_{aa}>0$ for every $\boldsymbol\zeta$.
\end{proof}

The family \eqref{eq:vpfamily} is the classical one: the cameras under which the three vanishing points of the frame are mutually orthogonal \citep{hartley2004}. The reason is that every line of direction $k$, in every junction of the frame, passes through the same vanishing point $\vv_k$, so any number of junctions encodes only the frame's vanishing points. \Cref{thm:converse} determines how much of that information they retain, and the answer depends on the image. 

By \cref{thm:converse}, each vanishing line contributes two conditions when it meets the image in a segment, one when it touches the image only at a corner, and none when it misses the image. Lines in general position either pass through the image or miss it, so the number of conditions one frame forces is $0$, $2$ or $3$: three when at least two lines are in view, and two when one is, as when only the horizon passes through the image. The touching case is exceptional but real; \OR{} (Sect.~S11) gives an example. One frame can therefore determine a camera only if the camera model has no more parameters than the frame forces conditions. A single focal length needs one, and \cref{thm:identframe} says when one frame supplies it; a general camera needs five, which one frame never supplies.

\subsection{Several frames}\label{sec:frames}

One frame never determines a general camera (\cref{prop:oneframe}). Several frames can, provided their axes point in enough different directions, and the next theorem says exactly when.

Let Manhattan frames $R_1,\dots,R_n$ be seen by one camera $K$, possibly in different images, and write a candidate camera as $\omega^{*}=K\Xi K^{\top}$, so that $\Xi=I$ is the true one. An axis is a direction without regard to sign; call it \emph{visible} if its vanishing line meets the image area $\mathcal I$ in a segment, and \emph{touching} if the line meets $\mathcal I$ in a single point $\vc_0$, whose ray is $\vr_0=K^{-1}\vc_0$.

\begin{theorem}[Several frames]\label{thm:frames}
Consider all the junctions of the frames with vertices in an image area $\mathcal I$ that satisfies the hypotheses of \cref{thm:converse} for every frame.
\begin{enumerate}
\item Every positive-definite $\omega^{*}=K\Xi K^{\top}$ that admits all these junctions has each visible axis $\vd$ as an eigenvector of $\Xi$ and, for each touching axis $\vd$, $\Xi\vd$ in the plane spanned by $\vd$ and $\vr_0$; conversely, every $\Xi$ near $I$ that satisfies these conditions is admitted. If two axes of each frame are visible, the admitted cameras are exactly those with $\Xi$ positive-definite and diagonal in each frame.
\item The junctions determine $\omega^{*}$, up to scale, if and only if the only symmetric $\Xi$ that satisfy the conditions of (1) are the multiples of $I$. If no axis is touching, this holds if and only if the visible axes do not all lie in the union of a plane through the origin and the line normal to it.
\end{enumerate}
\end{theorem}
\begin{proof}
For a frame $R$ scale the vanishing points as $V=KR$. The matrix of \cref{lem:frame} is then $N=R^{\top}\Xi R$, so that $N_{ab}=\vd_a^{\top}\Xi\vd_b$, and at a point with ray $\vr$ the coordinates are $\zeta_a=\vd_a\cdot\vr$. By \cref{thm:converse}(1) a visible axis $\vd_k$ forces $N_{ik}=N_{jk}=0$, that is, $\Xi\vd_k$ is perpendicular to $\vd_i$ and $\vd_j$ and hence parallel to $\vd_k$. A touching axis forces \eqref{eq:touch}, which reads $\vd_k^{\top}\Xi\,\mathbf w=0$ with $\mathbf w=\zeta_j(\vc_0)\,\vd_i-\zeta_i(\vc_0)\,\vd_j$. Since $\zeta_k(\vc_0)=0$, $\vr_0=\zeta_i\vd_i+\zeta_j\vd_j$, so $\mathbf w$ is perpendicular to both $\vd_k$ and $\vr_0$; it is nonzero, because $\vc_0$ is not a vanishing point, and so it is a nonzero multiple of $\vd_k\times\vr_0$. The condition therefore says that $\Xi\vd_k$ lies in the plane spanned by $\vd_k$ and $\vr_0$. Conversely, \cref{thm:converse}(2) admits every $\Xi$ near $I$ that satisfies these conditions for one frame, uniformly over its junctions through the constant $C$, and so, the frames being finitely many, every $\Xi$ near $I$ that satisfies them for all. If two axes of each frame are visible, $N$ is diagonal by \cref{thm:converse}(3), which is the last statement of (1).

For (2), the conditions are linear in $\Xi$ and are satisfied by $I$. If their only solutions are the multiples of $I$, every admitted $\omega^{*}$ is a multiple of $KK^{\top}$, which is admitted. If they have a solution $\Xi_1$ that is not a multiple of $I$, they have $I+t\,\Xi_1$ for every $t$; it is positive-definite and admitted for $t$ near $0$ by (1), and it is not a multiple of $I$ when $t\neq0$. Suppose now that no axis is touching. A symmetric matrix that is not a multiple of $I$ has either three distinct eigenvalues, and then its eigenvectors are three mutually orthogonal lines, or exactly two equal, and then they are a plane and its normal; in either case they lie in the union of a plane and its normal. If the visible axes do not, the conditions therefore have only the multiples of $I$ as solutions. If they lie in a plane with normal $\mathbf n$, each of them is an eigenvector of $I+t\,\mathbf n\mathbf n^{\top}$.
\end{proof}

If the visible axes do lie in a plane and along its normal $\mathbf n$, the matrix $\mathbf n\mathbf n^{\top}$ satisfies the conditions of part (1), so the cameras $K(I+t\,\mathbf n\mathbf n^{\top})K^{\top}$ are admitted for all $t$ near $0$, by the proof of part (2). When every frame shows two of its vanishing lines, part (1) describes the admitted cameras globally, and the condition of part (2) becomes that no axis is common to all the frames. Indeed, an orthonormal triple contained in a plane and its normal must contain the normal, since the plane holds at most two orthonormal vectors, so every frame then has the normal as an axis. Conversely, if all the frames share an axis $\vd$, the matrices $I+t\,\vd\vd^{\top}$, $t>-1$, are positive-definite and diagonal in each frame. They are the only ones as soon as two of the frames have no other axis in common, because a symmetric form that is diagonal in two orthonormal bases of the plane $\vd^{\perp}$ sharing no direction is a multiple of the identity there.

At the other extreme, an image may show only its horizon, the vanishing line of the vertical. It then contributes a single visible axis, its vertical direction in camera coordinates.

\begin{corollary}[Images that show only their horizons]\label{cor:horizons}
Let three or more images taken with one camera each show only the horizon of their frame, passing through the image, and let no two of their vertical directions be perpendicular. Their junctions determine a general camera if and only if the vertical vanishing points of the images are not collinear. If the camera only pitches between them, the cameras $K(I+t\,\mathbf e_1\mathbf e_1^{\top})K^{\top}$, which differ from the true one only in the aspect ratio, are admitted for $t$ near $0$; if it only rolls, the cameras $K(I+t\,\mathbf e_3\mathbf e_3^{\top})K^{\top}$ are, a focal-length ambiguity.
\end{corollary}
\begin{proof}
The visible axes are the vertical directions $\mathbf g_m$. When no two are perpendicular, they lie in the union of a plane and its normal only if they lie in the plane, since one along the normal would be perpendicular to the rest; and directions are coplanar exactly when their vanishing points $K\mathbf g_m$ lie on one line, the vanishing line of their plane. \Cref{thm:frames}(2) gives the first statement. Pitching keeps every $\mathbf g_m$ perpendicular to $\mathbf e_1$, and rolling keeps it in the image plane, perpendicular to $\mathbf e_3$, so \cref{thm:frames}(2) gives the two families.
\end{proof}

At least three images are therefore needed, and the camera must do more than rotate about one axis between them.

A touching axis adds one condition where a visible one adds two. Applied to the cameras $\operatorname{diag}(f',f',1)$, these conditions reproduce \cref{thm:identframe}, including its exceptional point (\OR, Sect.~S11).

\subsection{The focal length: identifiability}\label{sec:identifiability}

For a camera whose only unknown is the focal length, \cref{lem:frame} can be solved explicitly, and the joint admissible set of one frame's junctions takes a closed form.

\begin{proposition}[The joint admissible set in closed form]\label{prop:gform}
Let $K=\operatorname{diag}(f,f,1)$, scale the vanishing points of a frame $R$ as $V=KR$, so that $\zeta_a=\vr\cdot\vd_a$ for the viewing ray $\vr=K^{-1}\vc$, and put $z_a=\vd_a\cdot\hat{\mathbf z}$, with $\hat{\mathbf z}$ the optical axis, and $w_a=z_a/\zeta_a$. For a candidate focal length $f'$, with $s=f'/f$, the quantities of \cref{lem:frame} at the junction with vertex $\boldsymbol\zeta$ are
\begin{equation}
\nu_a=s^2+(1-s^2)\,\gamma_a,\qquad \gamma_a=(z_a-\zeta_aw_p)(z_a-\zeta_aw_q).
\label{eq:gform}
\end{equation}
Consequently, for any set of junctions of the frame, writing $\Gamma^{+}$ and $\Gamma^{-}$ for the suprema of $\gamma_a$ and of $-\gamma_a$ over the junctions and over $a$,
\begin{equation}
\bigcap F=\Bigl\{f':\ \frac{\Gamma^{-}}{1+\Gamma^{-}}<\frac{f'^2}{f^2}<\frac{\Gamma^{+}}{\Gamma^{+}-1}\Bigr\}
\label{eq:capF}
\end{equation}
up to its endpoints, the upper bound being absent when $\Gamma^{+}\le1$. Here $\Gamma^{-}$ is positive as soon as one junction has no edge line through the principal point, by \cref{prop:bound}(1); when $\Gamma^{-}\le0$ there is no lower bound.
\end{proposition}
\begin{proof}
A focal length $f'$ lies in every $F$ exactly when $\omega^{*}=\operatorname{diag}(x,x,1)$, $x=f'^2$, admits every junction, and then
\[
\begin{aligned}
N&=R^{\top}K^{-1}\operatorname{diag}(x,x,1)K^{-\top}R\\
&=s^2I+(1-s^2)\,\mathbf z\mathbf z^{\top}.
\end{aligned}
\]
In \eqref{eq:nuframe}, $\nu_a$ is linear in $N$ and equals $1$ at $N=I$; at the rank-one $N=\mathbf z\mathbf z^{\top}$ its four terms factor as $(z_a-\zeta_az_p/\zeta_p)(z_a-\zeta_az_q/\zeta_q)=\gamma_a$, so $\nu_a=s^2+(1-s^2)\gamma_a$, which is \eqref{eq:gform}. At $s=1$ every $\nu_a$ equals $1$. For $s<1$, $\nu_a>0$ exactly when $-\gamma_a<s^2/(1-s^2)$, and this holds at every junction exactly when $\Gamma^{-}\le0$ or $s^2>\Gamma^{-}/(1+\Gamma^{-})$; for $s>1$, $\nu_a>0$ exactly when $\gamma_a<s^2/(s^2-1)$, which holds at every junction exactly when $\Gamma^{+}\le1$ or $s^2<\Gamma^{+}/(\Gamma^{+}-1)$.
\end{proof}

Each junction therefore bounds $f$ from below when some $\gamma_a$ is negative and from above when some $\gamma_a$ exceeds one, and the bounds approach $f$ as $\gamma$ becomes unbounded. Since $w_b$ is unbounded only where $\zeta_b\to0$ with $z_b\neq0$, the bounds can approach $f$ only near the edge-on line of an axis that is not parallel to the image plane; the question is what happens there. For a single junction, \eqref{eq:capF} gives the interval as an explicit function of the junction's position.

\begin{theorem}[Where a junction bounds the focal length]\label{thm:boundmap}
Let $K=\operatorname{diag}(f,f,1)$, let a frame have axes $\vd_a$, and for a vertex $\vc$ off the vanishing lines let $\vr=K^{-1}\vc$. For $\{a,p,q\}=\{1,2,3\}$,
\begin{equation}
\gamma_a=-\,\frac{\bigl(\vr\cdot(\hat{\mathbf z}\times\vd_p)\bigr)\bigl(\vr\cdot(\hat{\mathbf z}\times\vd_q)\bigr)}{(\vr\cdot\vd_p)(\vr\cdot\vd_q)},
\label{eq:boundmap}
\end{equation}
and the junction of the frame with vertex $\vc$ admits exactly the focal lengths $f'$ with $\Gamma^-/(1+\Gamma^-)<f'^2/f^2<\Gamma^+/(\Gamma^+-1)$, where $\Gamma^+=\max_a\gamma_a$ and $\Gamma^-=\max_a(-\gamma_a)$, the lower bound being absent when $\Gamma^-\le0$ and the upper when $\Gamma^+\le1$. It therefore bounds $f$ from above exactly when some $\gamma_a>1$, and from below by a positive value exactly when some $\gamma_a<0$. If the axes $p$ and $q$ are tilted, then
\begin{equation}
\gamma_a=1+\frac{z_pz_q\,(\vc-\vv_p)\cdot(\vc-\vv_q)}{f^2\,(\vr\cdot\vd_p)(\vr\cdot\vd_q)},
\label{eq:thales}
\end{equation}
with $\vc$ and the vanishing points in centered pixel coordinates. Hence $\{\gamma_a=1\}$ is the circle with diameter $\vv_p\vv_q$, on which the junction's edges toward $\vv_p$ and $\vv_q$ are perpendicular in the image, and the region $\gamma_a>1$ is bounded by this circle and the edge-on lines $\ell_p$ and $\ell_q$.
\end{theorem}
\begin{proof}
Write $R_b(\vr)=\vr\cdot(\hat{\mathbf z}\times\vd_b)$ and $\ell_b(\vr)=\vr\cdot\vd_b$; their zero sets are the line $R_b$ through the principal point and $\vv_b$ and the edge-on line $\ell_b$, and the claim is $\gamma_a=-R_pR_q/(\ell_p\ell_q)$. With $\zeta_b=\vr\cdot\vd_b$ and $z_b=\hat{\mathbf z}\cdot\vd_b$, \eqref{eq:gform} reads $\gamma_a=(z_a\zeta_p-\zeta_az_p)(z_a\zeta_q-\zeta_az_q)/(\zeta_p\zeta_q)$. By the expansion of the triple product, $z_a\zeta_p-\zeta_az_p=\vr\cdot\bigl(\hat{\mathbf z}\times(\vd_p\times\vd_a)\bigr)$, and $\vd_p\times\vd_a=-\varepsilon\vd_q$ with $\varepsilon=\det[\vd_1\,\vd_2\,\vd_3]=\pm1$; likewise $z_a\zeta_q-\zeta_az_q=\varepsilon\,\vr\cdot(\hat{\mathbf z}\times\vd_p)$. Since $\varepsilon^2=1$, this is \eqref{eq:boundmap}. One junction is a set of junctions, so \eqref{eq:capF} gives its interval, and the two bounds follow. For \eqref{eq:thales}, write $\vr=(\vc/f,1)$. The quadratic $Q=R_pR_q+\ell_p\ell_q$ in $\vc$ has no term in $xy$ and the coefficient $\bar\vd_p\cdot\bar\vd_q/f^2=-z_pz_q/f^2\neq0$ on $x^2$ and on $y^2$, so its zero set is a circle. The circle passes through $\vv_p$ and $\vv_q$, where $\vr$ is proportional to $\vd_p$ or $\vd_q$, and through $\vc^{\star}_p$, where $R_p$ meets $\ell_p$. There $R_p$, through $\vv_p$, meets $\ell_p$, through $\vv_q$, at a right angle, since $\ell_p$ has normal $\bar\vd_p$, so by Thales' theorem the circle is the one with diameter $\vv_p\vv_q$; it passes likewise through $\vc^{\star}_q$. The difference between $Q$ and $-(z_pz_q/f^2)(\vc-\vv_p)\cdot(\vc-\vv_q)$ has no quadratic terms and vanishes at $\vv_p$, $\vv_q$ and $\vc^{\star}_p$, which are not collinear, because $\vv_p$ is not on $\ell_p$, unless $\vc^{\star}_p=\vv_q$; so it is zero, by continuity in the frame also in that case, and \eqref{eq:thales} follows from \eqref{eq:boundmap}. The edge of the junction along $\vd_p$ lies on the line through $\vc$ and $\vv_p$, so on this circle the edges along $\vd_p$ and $\vd_q$ are perpendicular in the image. Finally, $\gamma_a-1$ changes sign only across this circle and the lines $\ell_p$ and $\ell_q$.\end{proof}

\begin{figure*}[t]
\centering
\includegraphics[width=0.92\linewidth]{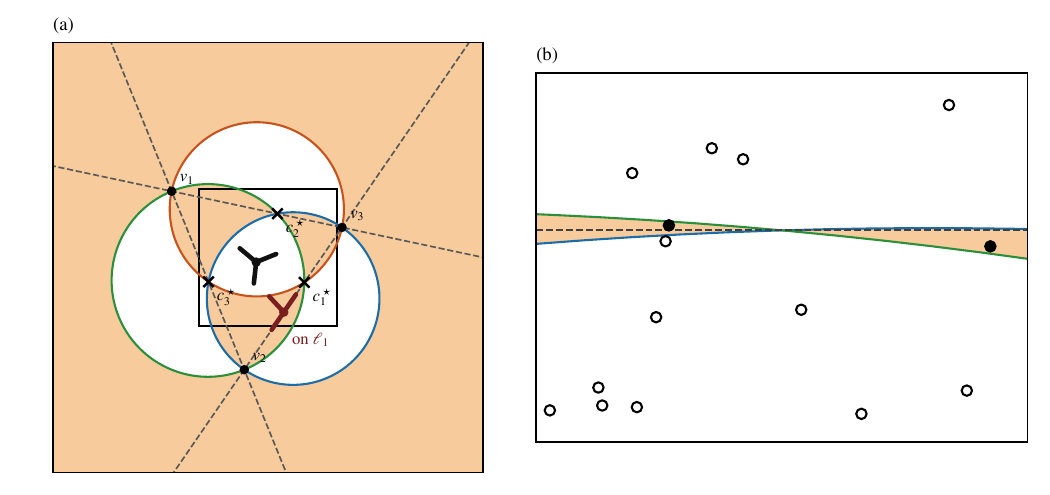}
\caption{Where a junction bounds the focal length from above (\cref{thm:boundmap}). (a) A camera seen obliquely, the image outlined in black: junctions with vertices in the shaded region bound $f$ from above. Each axis contributes the circle with diameter between the other two vanishing points (dots), which passes through the points $\vc^{\star}$ (crosses), and the edge-on lines (dashed) complete the boundary. Of the two junctions drawn, the one far from every edge-on line has three distinct edges; the one marked ``on $\ell_1$'' lies on that line, where its two edges along the other axes fall on one image line, the face they span being seen edge-on. (b) A simulated nearly level camera, matched to York Urban ($640\times480$\,px, $f=675$\,px, a $3^\circ$ tilt): the region is a thin wedge along the horizon (dashed), covering $4\%$ of the image, and only $2$ of $15$ junctions at random positions (filled) lie in it, which is why upper bounds are rare for ordinary, nearly level cameras}
\label{fig:boundmap}
\end{figure*}

The circle of axis $a$ passes through the points $\vc^{\star}_p$ and $\vc^{\star}_q$ of the edge-on lines $\ell_p$ and $\ell_q$ nearest the principal point. The region in which a junction bounds the focal length from above is thus bounded by three circles, one for each axis, and by the edge-on lines (\cref{fig:boundmap}). On the circle of axis $a$, the junction's edges along the other two axes appear perpendicular: this is the raw right angle at which the threshold of that pair passes through infinity (\cref{thm:interval}). The circle is also the section by the image plane of the \emph{calibration sphere} with diameter $\vv_p\vv_q$, on which the center of projection lies because it sees $\vv_p$ and $\vv_q$ in perpendicular directions; in photogrammetry, this section is known as the locus of the isocenter \citep{grammatikopoulos2004}. The region hugs the edge-on lines of the tilted axes, where $\gamma$ is unbounded, and \cref{thm:identframe} states that it reaches $f$ from both sides when such a line passes through the image. On the York Urban images of \OR{} (Sect.~S12), the region covers a median $9.4\%$ of the image and a mean $12.3\%$; yet when the $181$ real junctions are replayed without noise, only $4$ of them, $2.2\%$, lie in it. Real corners avoid the neighborhoods of the edge-on lines, where two of their edges would look nearly collinear.

\begin{theorem}[Identifiability of the focal length from one frame]\label{thm:identframe}
Let the camera be $K=\operatorname{diag}(f,f,1)$ and the image a closed convex polygon with nonempty interior, such as a rectangle, with no vanishing point on its boundary; consider all the junctions of one Manhattan frame with vertices in the image and off its vanishing lines. Call an axis \emph{tilted} if it is not parallel to the image plane, and for a tilted axis $\vd_k$ let $\vc^{\star}_k$ be the point of its edge-on line $\ell_k$ nearest the principal point; if $\vd_k$ is the optical axis, that line is at infinity, never meets the image, and plays no part. Then $\bigcap F=\{f\}$ if and only if, for some tilted axis, $\ell_k$ meets the image at a point other than $\vc^{\star}_k$.
\end{theorem}
\begin{proof}
By \cref{prop:gform}, $\bigcap F=\{f\}$ exactly when $\gamma$ is unbounded above and below on the image. In the notation of the proof of \cref{thm:boundmap}, $\gamma_a=-R_pR_q/(\ell_p\ell_q)$, where the line $R_b$ through the principal point and $\vv_b$ is perpendicular to $\ell_b$ and meets it at $\vc^{\star}_b$. So $\gamma$ is bounded away from the edge-on lines, and the line $\ell_k$ of an untilted axis contributes nothing: it then passes through the principal point and coincides with $R_i$ and $R_j$, which cancel it.

Let $\ell_k$, with $z_k\neq0$, meet the image at a point $\vc_0$ that is not a vanishing point. The components with $\ell_k$ in their denominators are $\gamma_i=-(R_j/\ell_j)(R_k/\ell_k)$ and $\gamma_j=-(R_i/\ell_i)(R_k/\ell_k)$. On $\ell_k$, where $\vr=\zeta_i\vd_i+\zeta_j\vd_j$ and $\vd_i\times\vd_j=\varepsilon\vd_k$, we have $R_i=\varepsilon z_k\zeta_j$ and $R_j=-\varepsilon z_k\zeta_i$, so $(R_i/\ell_i)(R_j/\ell_j)=-z_k^2<0$ at $\vc_0$. As the vertex tends to $\vc_0$ from within the image, $R_k/\ell_k\to\pm\infty$ unless $R_k(\vc_0)=0$, that is, unless $\vc_0=\vc^{\star}_k$; then $\gamma_i$ and $\gamma_j$ diverge with opposite signs, whichever side of $\ell_k$ the vertex is on, and $\bigcap F=\{f\}$. A line that meets the image in more than one point meets it in a segment, which contains points other than $\vc^{\star}_k$ and the vanishing points; one that meets it in a single point meets it at a boundary point that is not a vanishing point. This proves sufficiency.

Conversely, if every edge-on line of a tilted axis that meets the image meets it only at its point $\vc^{\star}_k$, that point is a corner of the image, which lies in a wedge whose sides make positive angles with $\ell_k$. In the wedge, $|R_k|$, proportional to the distance from $R_k$, is at most a constant times $|\ell_k|$, so $R_k/\ell_k$ and hence $\gamma$ are bounded on the image, and $\bigcap F$ contains a neighborhood of $f$.
\end{proof}

A vanishing line need not pass through the image for its junctions to determine $f$: it suffices that the line reach the image, because a single junction near the line bounds $f$ from both sides at once, $\gamma_i$ and $\gamma_j$ having opposite signs. The exceptional point $\vc^{\star}_k$ is real, and \OR{} (Sect.~S10) shows its cost.

\begin{corollary}[A wide field of view suffices]\label{cor:wide}
Under the hypotheses of \cref{thm:identframe}, if no axis of the frame is parallel to the image plane and the image contains a cone of half-angle greater than $\arcsin(1/\sqrt3)\approx35.26^\circ$ about the optical axis, a circular field of view above $70.53^\circ$, then $\bigcap F=\{f\}$. The constant is sharp: for the frame seen down a cube diagonal, a circular field of view of at most $70.53^\circ$ does not determine $f$.
\end{corollary}
\begin{proof}
Since $\sum_a(\vd_a\cdot\hat{\mathbf z})^2=1$ for an orthonormal frame, some axis has $0<|z_a|\le1/\sqrt3$. Its edge-on line consists of the rays perpendicular to $\vd_a$, a plane that makes the angle $\arcsin|z_a|\le\arcsin(1/\sqrt3)$ with $\hat{\mathbf z}$. The line therefore meets the open cone, and so the interior of the image, in a segment, which contains points other than $\vc^{\star}_a$; \cref{thm:identframe} applies. For the cube diagonal every axis has $|z_a|=1/\sqrt3$, so each edge-on line keeps an angle of exactly $\arcsin(1/\sqrt3)$ from the optical axis; a circular image of at most that half-angle meets each line at most at its point $\vc^{\star}_a$, and \cref{thm:identframe} shows that $f$ is then not determined.
\end{proof}

The restriction to tilted axes matters in practice, because the commonest camera pose is nearly level. For a level camera, the vertical axis is parallel to the image plane and its edge-on line is the horizon, so $w_k=0$ and the horizon contributes nothing to $\gamma$. Concretely, if $u$ is the abscissa of the junction, the thresholds of the two pairs containing the vertical edge are $-u$ times the abscissae of the two horizontal vanishing points, independent of the junction's distance from the horizon. No junction near the horizon is therefore bounded above through its vertical edge. The horizontal axes of a level camera are tilted, however, and the edge-on line of each is the vertical image line through the other horizontal vanishing point. If neither horizontal vanishing point lies within the horizontal extent of the image, these lines miss it and upper bounds are rare; if one does, as in corridor or street views, its line passes through the image and \cref{thm:identframe} applies. Tilting the camera restores identifiability only gradually. Take $f=900$\,px and a $960$\,px square image, with vertices uniform over it and the horizontal vanishing points outside its horizontal extent, so that only the horizon is in view. The share of junctions bounded above is then $0.4\%$, $1.8\%$, $4.4\%$, $10.8\%$ and $22.2\%$ at tilts of $0.5^\circ$, $2^\circ$, $5^\circ$, $12^\circ$ and $25^\circ$, against none for a level camera, for which the intersection of $40\,000$ admissible sets is $[0.80f,\infty)$. \OR{} (Sect.~S12) finds the consequence in real images.

\paragraph{The mechanism at a single junction.} At the true camera, the interpretation-plane normal $\vn_m$ is parallel to $\vr\times\vd_m$, so for two orthogonal edges
\begin{equation}
\begin{aligned}
(\vr\times\vd_i)\cdot(\vr\times\vd_j)&=\|\vr\|^2(\vd_i\cdot\vd_j)-(\vr\cdot\vd_i)(\vr\cdot\vd_j)\\
&=-(\vr\cdot\vd_i)(\vr\cdot\vd_j).
\end{aligned}
\label{eq:edgeon}
\end{equation}
The pair form therefore vanishes at the true $f$ exactly when one of its two edges is \emph{perpendicular to the viewing ray}. Geometrically, $\vr\perp\vd_i$ places the ray in the face spanned by the other two edges, so that face is seen edge-on and those two edges project onto one image line: the configuration is the degeneracy of \cref{sec:sign}, approached rather than reached. The condition $\vr\perp\vd_i$ is the vanishing line $\ell_i$ of \cref{sec:oneframe} seen from a single junction: as the vertex approaches $\ell_i$, the thresholds of the two pairs that contain edge $i$ both tend to $f$, one from each side. The junctions that pin $f$ most tightly are therefore those that see a face nearly edge-on, which are also those closest to the degeneracy of two edges sharing an image line. 

\subsection{Focal length and principal point}\label{sec:fpp}

\Cref{thm:identframe} assumes the principal point known. When it is unknown as well, \cref{thm:frames}(1) still applies, and its conditions can be solved completely for a camera with square pixels and zero skew. For an axis $\vd$, write $\bar{\vd}\in\mathbb R^2$ for its first two coordinates, its direction in the image plane, and $z=\vd\cdot\hat{\mathbf z}$, as in \cref{prop:gform}; the axis is tilted when $z\neq0$.

\begin{theorem}[Focal length and principal point]\label{thm:fpp}
Let the camera have square pixels and zero skew, with its focal length and principal point unknown, and let the frames and visible axes be as in \cref{thm:frames}. The junctions determine the focal length and the principal point unless
\begin{enumerate}
\item no visible axis is tilted; or
\item the visible tilted axes are a single axis, or two perpendicular axes, lying in a plane $\Theta$ that contains the optical axis, and every other visible axis is normal to $\Theta$.
\end{enumerate}
If no axis is touching, then in case (1) the focal length is not determined, and in case (2) the admitted cameras near the true one form a one-parameter family, in which the principal point moves along the image of $\Theta$, the line through the principal point with direction $\bar{\vd}$, for any tilted axis $\vd$.
\end{theorem}
\begin{proof}
Write a candidate camera as $K'=KT$ with
\[
T=\begin{pmatrix} sI & \mathbf p\\ \mathbf 0^{\top} & 1\end{pmatrix},\qquad
\Xi=TT^{\top}=\begin{pmatrix} s^2I+\mathbf p\mathbf p^{\top} & \mathbf p\\ \mathbf p^{\top} & 1\end{pmatrix},
\]
where $s>0$ is the ratio of the focal lengths and $\mathbf p\in\mathbb R^2$ the displacement of the principal point divided by $f$. By \cref{thm:frames}(1) each visible axis $\vd$ satisfies $\Xi\vd=\lambda\vd$ with $\lambda>0$, that is,
\[
s^2\bar{\vd}+(\mathbf p\cdot\bar{\vd}+z)\,\mathbf p=\lambda\bar{\vd},\qquad \mathbf p\cdot\bar{\vd}+z=\lambda z .
\]
A visible axis is not the optical axis, whose vanishing line is at infinity, so $\bar{\vd}\neq0$. If $z=0$, the second equation gives $\mathbf p\cdot\bar{\vd}=0$, and the first then holds with $\lambda=s^2$. If $z\neq0$, substituting the second equation into the first gives $\lambda z\,\mathbf p=(\lambda-s^2)\,\bar{\vd}$, so $\mathbf p=m\bar{\vd}$ for some $m$, and then
\begin{equation}
\lambda=1+m\,\|\bar{\vd}\|^2/z,\qquad s^2=\lambda\,(1-mz).
\label{eq:fppaxis}
\end{equation}

Suppose some visible axis is tilted and case (2) does not hold. If two tilted axes have directions $\bar{\vd}$ that are not parallel, $\mathbf p=0$. Otherwise these directions are all multiples of one unit vector $\mathbf u$, the tilted axes lie in the plane $\Theta$ spanned by $(\mathbf u,0)$ and $\hat{\mathbf z}$, and $\mathbf p=h\mathbf u$. An untilted visible axis $\mathbf e$ with $\bar{\mathbf e}\cdot\mathbf u\neq0$ then gives $h=0$. If there is none, every untilted visible axis is normal to $\Theta$, and since case (2) does not hold, $\Theta$ contains two visible tilted axes that are neither equal nor perpendicular. Write them as $\vd_a=(c_a\mathbf u,z_a)$ with $c_a^2+z_a^2=1$, $a=1,2$, so that $m_a=h/c_a$. Then \eqref{eq:fppaxis} becomes
\[
s^2=1+\tau_ah-h^2,\qquad \tau_a=\frac{c_a}{z_a}-\frac{z_a}{c_a}=2\cot2\theta_a ,
\]
where $c_a=\cos\theta_a$ and $z_a=\sin\theta_a$. Now $\tau_1=\tau_2$ only if $\theta_1-\theta_2$ is a multiple of $\pi/2$, that is, only if the axes are equal or perpendicular, so $h=0$. In every case therefore $\mathbf p=0$; for a tilted axis $\lambda=1$, and the first equation gives $s=1$. The only camera that satisfies the conditions is the true one, and it is admitted.

Suppose finally that no axis is touching. In case (1) every visible axis has $z=0$, and $\mathbf p=0$ with any $s$ satisfies the conditions; by \cref{thm:frames}(1) the cameras with $s$ near $1$ are admitted, so the focal length is not determined. In case (2) all the tilted axes have the same $\tau_a=\tau$, for a single axis trivially and for two perpendicular ones by the remark above. Then $\mathbf p=h\mathbf u$ with $s^2=1+\tau h-h^2$ satisfies the conditions of every tilted axis, and of every untilted one, which is normal to $\Theta$ so that $\bar{\mathbf e}\cdot\mathbf u=0$; by \cref{thm:frames}(1) these cameras are admitted for $h$ near $0$.
\end{proof}

The exceptional configurations are classical. One frame with all three vanishing lines in view determines the camera unless one of its axes is parallel to the image plane, which is case (2), the failure of the orthocenter construction when a vanishing point is at infinity. A single image from a tilted camera showing only its horizon is also case (2): the principal point can slide along the line joining it to the vertical vanishing point; for a level camera the same image is case (1). 

\OR{} (Sect.~S5) gives the case of two images that show only their horizons. A touching axis adds one condition, which may remove the family of case (2); for the focal length alone, this is the statement of \cref{thm:identframe} that a vanishing line touching the image away from $\vc^{\star}_k$ determines $f$.

\subsection{The convex program}\label{sec:program}

\Cref{thm:convex} turns calibration from junctions into a concrete convex program, because each junction's sign pattern is read from the junction alone, so the constraints can be written down without knowing $K$.

For a camera with known principal point and zero skew, $\omega^{*}=\operatorname{diag}(X,Y,1)$ and positive-definiteness reduces to $X,Y>0$, so the cameras consistent with $M$ junctions form the feasible set of a linear program, whose extent in $X$ and in $Y$ is again a linear program. For the full camera, with skew and principal point free, positive-definiteness is a matrix inequality, which we impose with cutting planes; this requires nothing beyond a linear-programming solver. From noiseless junctions of independent frames, every parameter is recovered to within about one pixel (\OR, Sect.~S11). From the junctions of a single frame, by \cref{prop:oneframe}, the program cannot close on a general camera; \OR{} (Sect.~S12) applies a robust form of it to real junctions.

\paragraph{Procedure.} (i) For each junction, read the three rank-one generators and their sign pattern from the junction alone (\cref{thm:convex}); each condition $\nu_k>0$ is then a linear inequality in the entries of $\omega^{*}$. (ii) Normalize $\omega^{*}_{33}=1$ and solve the linear program, with any objective, in a bounded box. In practice the box comes from loose prior bounds, for instance focal lengths between a tenth and ten times the image width and a principal point inside the image; it need only contain the true camera. (iii) If the solution is not positive-definite, add the cut $\mathbf v^{\top}\omega^{*}\mathbf v\ge0$ for the eigenvector $\mathbf v$ of its most negative eigenvalue and solve again, until the smallest eigenvalue is nonnegative to tolerance. (iv) The extent of each entry of $\omega^{*}$ is the value of a further linear program over the final feasible set. (v) Under noise the strict feasible set empties as junctions accumulate; replace it by a depth-$\kappa$ consensus, or weight each junction by its miss probability, as in \OR{} (Sects.~S7 and S8).

\begin{proposition}[The convex program]\label{prop:program}
Let $M$ junctions be admitted by a camera $K_0$, and normalize $\omega^{*}_{33}=1$.
\begin{enumerate}
\item The set of $\omega^{*}$ that admit all $M$ junctions is convex: the intersection of the positive-definite cone with the $3M$ open half-spaces $\nu^{(m)}_k(\omega^{*})>0$ of \cref{thm:convex}, one for each image line of each junction.
\item Run the procedure with the closed constraints $\nu^{(m)}_k\ge0$ in a compact box containing $K_0K_0^{\top}$. Then every linear program is feasible, the iterates $\omega_t$ satisfy $\liminf_t\lambda_{\min}(\omega_t)\ge0$, and every accumulation point satisfies all the junction constraints and is positive semidefinite. In particular, for every $\tau>0$ the stopping rule $\lambda_{\min}(\omega_t)\ge-\tau$ is met after finitely many cuts.
\end{enumerate}
\end{proposition}
\begin{proof}
(1) By \cref{thm:convex}, a junction is admitted exactly when its three coefficients $\nu_k$, each linear in $\omega^{*}$, are positive.

(2) The matrix $K_0K_0^{\top}$ satisfies every junction constraint and every cut $\mathbf v^{\top}\omega^{*}\mathbf v\ge0$, so every linear program is feasible, and, the box being compact, it has a solution $\omega_t$. Let $\mathbf v_s$ be the unit eigenvector whose cut is added at step $s$, so that $\mathbf v_s^{\top}\omega_s\mathbf v_s=\lambda_{\min}(\omega_s)$. Every later iterate satisfies that cut, so for $t>s$
\[
\lambda_{\min}(\omega_s)=\mathbf v_s^{\top}(\omega_s-\omega_t)\mathbf v_s+\mathbf v_s^{\top}\omega_t\mathbf v_s\ge-\|\omega_s-\omega_t\| .
\]
If $\omega_{t_m}\to\bar\omega$ along a subsequence, taking $s=t_m$ and $t=t_{m+1}$ gives $\lambda_{\min}(\omega_{t_m})\ge-\|\omega_{t_m}-\omega_{t_{m+1}}\|\to0$, so $\lambda_{\min}(\bar\omega)\ge0$ by continuity; and $\bar\omega$ satisfies the junction constraints, which hold at every iterate and are closed. Finally, if $\lambda_{\min}(\omega_t)<-\tau$ for infinitely many $t$, these iterates have, by compactness, a convergent subsequence whose limit has $\lambda_{\min}\le-\tau<0$, a contradiction. Hence $\liminf_t\lambda_{\min}(\omega_t)\ge0$, and for every $\tau>0$ the stopping rule $\lambda_{\min}(\omega_t)\ge-\tau$ is met after finitely many cuts.
\end{proof}

\paragraph{Under noise.} The strict intersection of admissible sets loses coverage as junctions accumulate, because a single wrong interval suffices to exclude the truth. \OR{} (Sect.~S8) gives a depth-$\kappa$ consensus whose coverage follows a binomial prediction, and a log-plausibility built on the noise analysis of Sect.~S7. In simulation, its maximizer needs ten to twenty junctions to come within $10\%$ of $f$ at $\sigma=0.5$\,px, several images' worth at the $1.8$ full junctions per image of York Urban (\OR, Sect.~S12).

\section{Experiments}\label{sec:experiments}

We check \cref{thm:interval} in simulation and then test the predictions on real images. \OR{} (Sect.~S10) gives the full protocol, the checks of the noise model of Sect.~S7, the identifiability transition behind \cref{thm:identframe}, and the robustness of the bound to the protocol's idealizations, among them other noise levels, a tilted camera and a misplaced principal point; the dependence on edge length is analyzed in Sect.~S7, lens distortion in Sect.~S6, and further experiments are in Sect.~S13. Offsetting the assumed principal point by up to $20$\,px leaves the coverage of the true focal length at $94$ to $98\%$ at all three fields of view.

\subsection{The focal-length bound}\label{sec:e2}
Corners are sampled uniformly over orientations and over a $1000\times1000$\,px image, at fields of view of $40^\circ$, $60^\circ$ and $90^\circ$ ($f=1374$, $866$ and $500$\,px), and each edge is observed as a $150$\,px segment whose endpoints carry Gaussian noise of standard deviation $\sigma$ (\OR, Sect.~S10). Over $1000$ corners per field of view, with $\sigma=0.5$\,px and $150$\,px edges, the admissible set $F$ contained the true $f$ in 98\%, 97\% and 97\% of cases at $40^\circ$, $60^\circ$ and $90^\circ$, and it was a single interval in all $3000$ cases, as \cref{thm:interval} requires. It was unbounded above for 90\%, 81\% and 66\% of corners, and a check at $\sigma=0$ confirmed in $3000$ of $3000$ cases that unboundedness coincides with the raw image angles passing the orthographic fork-or-arrow rule, as \cref{prop:bound}(2) states.

The lower bound rises with distance from the center: the median of $\fmin/f$ is $0.08$, $0.15$ and $0.21$ in the inner third of the image and $0.29$, $0.41$ and $0.52$ in the outer third, at the three fields of view (\OR, Sect.~S10). An upper bound appears for 13\%, 25\% and 43\% of outer-third corners, at a median of $1.3$ to $1.4$ times $f$. The similarity law of \cref{cor:similar} accounts for most of the variation with position. Measured in units of $\rho$ at $\sigma=0$, the median of $\fmin/\rho$ stays within a factor of two across bins of distance from the center and fields of view, whereas $\fmin/f$ ranges over a factor of seven (\OR, Sect.~S10). The growth of the bound with radius is therefore mostly the factor $\rho$ of \cref{cor:similar}; the remainder reflects how the angles $\psi_i$ are distributed over junctions.

\subsection{Real images}\label{sec:realsummary}

On the $102$ images of the York Urban Database \citep{denis2008}, the admissible set contained the true focal length for $179$ of $181$ real junctions, and the certified sets of \cref{prop:certified} for all of them. Only two junctions bounded the focal length from above: real corners avoid the neighborhoods of the edge-on lines, so the bounds do not calibrate this nearly level camera. Calibration from three vanishing points does better where line families are available. \OR{} (Sect.~S12) gives the data, the junction extraction and every number.

\section{Conclusion}\label{sec:conclusion}

A right-angle corner constrains an uncalibrated camera through a single identity: three image lines through a point are the image of such a corner under a given camera exactly when a product of three forms, each linear in the dual image of the absolute conic, is negative. Everything in this paper follows from it. One junction admits a convex cone of cameras, the same at every junction up to a linear change of coordinates, and a closed-form interval when only the focal length is unknown. With the vertex position known, these bounds can be certified against any bounded error in the measured edge directions. When the corners belong to one rectilinear structure, their junctions reject, for each axis whose edge-on line passes through the image, exactly the cameras under which that axis's vanishing point is not orthogonal to the other two, and no other cameras near the truth: junctions carry at most the information of the vanishing points. For the focal length, a junction bounds it from above exactly inside circles whose diameters join pairs of vanishing points, and the corners of one structure determine it when the edge-on line of a tilted axis passes through the image. Since each junction's constraint is computed from the image alone, calibration from junctions is a convex feasibility problem.

The experiments agree with the single-junction results, and on real images they mark the limit of the approach. Real corners avoid the neighborhoods of the edge-on lines, so a nearly level camera with an ordinary field of view obtains a near-tight lower bound on its focal length rather than an estimate, and line families, where they exist, remain the better route. The experiments do not reach the cluttered, close-range regime that motivates the theory; whether cluttered interiors and automatic junction detection change this picture is the open empirical question, and \OR{} (Sect.~S12) discusses this and the other limitations of the evaluation. Several theoretical questions remain open as well. Radial distortion acts on each junction by an explicit transformation (\OR, Sect.~S6), which suggests estimating the distortion coefficient from junctions rather than treating it as a nuisance. The admitted cameras are described globally when one edge-on line passes through the image (\cref{prop:oneline}), but not in general. Nor is the trihedral with known non-right angles characterized.

\backmatter

\section*{Declarations}

\paragraph{Funding.} No funding was received for conducting this study.

\paragraph{Competing interests.} The author has no competing interests to declare that are relevant to the content of this article.

\paragraph{Ethics approval and consent.} Not applicable.

\paragraph{Data availability.} The York Urban Database \citep{denis2008} analyzed in this study is publicly available from its authors. No new image data were generated.

\paragraph{Code availability.} The code that verifies every identity symbolically and regenerates every number and figure in this paper is submitted with this manuscript as Online Resource 2 and will be released in a public GitHub repository, whose address will be given in the published version.

\paragraph{Author contributions.} The author is the sole author of this work, with the assistance described below.

\paragraph{Use of AI tools.} This work was carried out with substantial assistance from a large language model, Claude (Anthropic), in the versions available during 2026, including Claude Opus 5.5. The author posed the research question, chose the problems to pursue, decided what to prove, what to include and what to set aside, and directed every revision, including the responses to referee reports. Under that direction, the tool proposed and developed derivations and proofs, including those of the main theorems; wrote the symbolic verification code and the experiment and figure code; ran the experiments on data that the author obtained; drafted and revised the text; and located and checked references. Every algebraic identity in the paper is checked symbolically by the supplied code, every reported number is regenerated by it, and every cited reference was checked against its publisher or an authoritative index. The author checked every theorem and proof personally, reviewed all the results, and takes full responsibility for the content of the paper.

\bibliography{refs}


\begin{thebibliography}{39}
\ifx \bisbn   \undefined \def \bisbn  #1{ISBN #1}\fi
\ifx \binits  \undefined \def \binits#1{#1}\fi
\ifx \bauthor  \undefined \def \bauthor#1{#1}\fi
\ifx \batitle  \undefined \def \batitle#1{#1}\fi
\ifx \bjtitle  \undefined \def \bjtitle#1{#1}\fi
\ifx \bvolume  \undefined \def \bvolume#1{\textbf{#1}}\fi
\ifx \byear  \undefined \def \byear#1{#1}\fi
\ifx \bissue  \undefined \def \bissue#1{#1}\fi
\ifx \bfpage  \undefined \def \bfpage#1{#1}\fi
\ifx \blpage  \undefined \def \blpage #1{#1}\fi
\ifx \burl  \undefined \def \burl#1{\textsf{#1}}\fi
\ifx \doiurl  \undefined \def \doiurl#1{\url{https://doi.org/#1}}\fi
\ifx \betal  \undefined \def \betal{\textit{et al.}}\fi
\ifx \binstitute  \undefined \def \binstitute#1{#1}\fi
\ifx \binstitutionaled  \undefined \def \binstitutionaled#1{#1}\fi
\ifx \bctitle  \undefined \def \bctitle#1{#1}\fi
\ifx \beditor  \undefined \def \beditor#1{#1}\fi
\ifx \bpublisher  \undefined \def \bpublisher#1{#1}\fi
\ifx \bbtitle  \undefined \def \bbtitle#1{#1}\fi
\ifx \bedition  \undefined \def \bedition#1{#1}\fi
\ifx \bseriesno  \undefined \def \bseriesno#1{#1}\fi
\ifx \blocation  \undefined \def \blocation#1{#1}\fi
\ifx \bsertitle  \undefined \def \bsertitle#1{#1}\fi
\ifx \bsnm \undefined \def \bsnm#1{#1}\fi
\ifx \bsuffix \undefined \def \bsuffix#1{#1}\fi
\ifx \bparticle \undefined \def \bparticle#1{#1}\fi
\ifx \barticle \undefined \def \barticle#1{#1}\fi
\bibcommenthead
\ifx \bconfdate \undefined \def \bconfdate #1{#1}\fi
\ifx \botherref \undefined \def \botherref #1{#1}\fi
\ifx \url \undefined \def \url#1{\textsf{#1}}\fi
\ifx \bchapter \undefined \def \bchapter#1{#1}\fi
\ifx \bbook \undefined \def \bbook#1{#1}\fi
\ifx \bcomment \undefined \def \bcomment#1{#1}\fi
\ifx \oauthor \undefined \def \oauthor#1{#1}\fi
\ifx \citeauthoryear \undefined \def \citeauthoryear#1{#1}\fi
\ifx \endbibitem  \undefined \def \endbibitem {}\fi
\ifx \bconflocation  \undefined \def \bconflocation#1{#1}\fi
\ifx \arxivurl  \undefined \def \arxivurl#1{\textsf{#1}}\fi
\csname PreBibitemsHook\endcsname

\bibitem[\protect\citeauthoryear{Perkins}{1968}]{perkins1968}
\begin{botherref}
\oauthor{\bsnm{Perkins}, \binits{D.N.}}:
Cubic corners.
Quarterly Progress Report~89,
MIT Research Laboratory of Electronics
(1968).
\url{http://hdl.handle.net/1721.1/55745}
\end{botherref}
\endbibitem

\bibitem[\protect\citeauthoryear{Kanade}{1980}]{kanade1980}
\begin{barticle}
\bauthor{\bsnm{Kanade}, \binits{T.}}:
\batitle{A theory of {O}rigami world}.
\bjtitle{Artificial Intelligence}
\bvolume{13}(\bissue{3}),
\bfpage{279}--\blpage{311}
(\byear{1980})
\doiurl{10.1016/0004-3702(80)90004-1}
\end{barticle}
\endbibitem

\bibitem[\protect\citeauthoryear{Barnard}{1983}]{barnard1983}
\begin{barticle}
\bauthor{\bsnm{Barnard}, \binits{S.T.}}:
\batitle{Interpreting perspective images}.
\bjtitle{Artificial Intelligence}
\bvolume{21}(\bissue{4}),
\bfpage{435}--\blpage{462}
(\byear{1983})
\doiurl{10.1016/S0004-3702(83)80021-6}
\end{barticle}
\endbibitem

\bibitem[\protect\citeauthoryear{Horaud}{1987}]{horaud1987}
\begin{barticle}
\bauthor{\bsnm{Horaud}, \binits{R.}}:
\batitle{New methods for matching {3-D} objects with single perspective views}.
\bjtitle{IEEE Transactions on Pattern Analysis and Machine Intelligence}
\bvolume{9}(\bissue{3}),
\bfpage{401}--\blpage{412}
(\byear{1987})
\doiurl{10.1109/TPAMI.1987.4767922}
\end{barticle}
\endbibitem

\bibitem[\protect\citeauthoryear{Kanatani}{1986}]{kanatani1986}
\begin{barticle}
\bauthor{\bsnm{Kanatani}, \binits{K.-i.}}:
\batitle{The constraints on images of rectangular polyhedra}.
\bjtitle{IEEE Transactions on Pattern Analysis and Machine Intelligence}
\bvolume{8}(\bissue{4}),
\bfpage{456}--\blpage{463}
(\byear{1986})
\doiurl{10.1109/TPAMI.1986.4767809}
\end{barticle}
\endbibitem

\bibitem[\protect\citeauthoryear{Kanatani}{1988}]{kanatani1988}
\begin{barticle}
\bauthor{\bsnm{Kanatani}, \binits{K.-i.}}:
\batitle{Constraints on length and angle}.
\bjtitle{Computer Vision, Graphics, and Image Processing}
\bvolume{41}(\bissue{1}),
\bfpage{28}--\blpage{42}
(\byear{1988})
\doiurl{10.1016/0734-189X(88)90115-6}
\end{barticle}
\endbibitem

\bibitem[\protect\citeauthoryear{Shakunaga and Kaneko}{1986}]{shakunaga1986}
\begin{bchapter}
\bauthor{\bsnm{Shakunaga}, \binits{T.}},
\bauthor{\bsnm{Kaneko}, \binits{H.}}:
\bctitle{Perspective angle transform and its application to 3-{D} configuration recovery}.
In: \bbtitle{Proc. IEEE Conference on Computer Vision and Pattern Recognition (CVPR)},
pp. \bfpage{594}--\blpage{601}
(\byear{1986})
\end{bchapter}
\endbibitem

\bibitem[\protect\citeauthoryear{Shakunaga and Kaneko}{1989}]{shakunaga1989}
\begin{barticle}
\bauthor{\bsnm{Shakunaga}, \binits{T.}},
\bauthor{\bsnm{Kaneko}, \binits{H.}}:
\batitle{Perspective angle transform: Principle of shape from angles}.
\bjtitle{International Journal of Computer Vision}
\bvolume{3},
\bfpage{239}--\blpage{254}
(\byear{1989})
\doiurl{10.1007/BF00133033}
\end{barticle}
\endbibitem

\bibitem[\protect\citeauthoryear{Huffman}{1971}]{huffman1971}
\begin{bchapter}
\bauthor{\bsnm{Huffman}, \binits{D.A.}}:
\bctitle{Impossible objects as nonsense sentences}.
In: \beditor{\bsnm{Meltzer}, \binits{B.}},
\beditor{\bsnm{Michie}, \binits{D.}} (eds.)
\bbtitle{Machine Intelligence 6},
pp. \bfpage{295}--\blpage{323}.
\bpublisher{Edinburgh University Press},
\blocation{Edinburgh}
(\byear{1971})
\end{bchapter}
\endbibitem

\bibitem[\protect\citeauthoryear{Clowes}{1971}]{clowes1971}
\begin{barticle}
\bauthor{\bsnm{Clowes}, \binits{M.B.}}:
\batitle{On seeing things}.
\bjtitle{Artificial Intelligence}
\bvolume{2},
\bfpage{79}--\blpage{116}
(\byear{1971})
\doiurl{10.1016/0004-3702(71)90005-1}
\end{barticle}
\endbibitem

\bibitem[\protect\citeauthoryear{Sugihara}{1986}]{sugihara1986}
\begin{bbook}
\bauthor{\bsnm{Sugihara}, \binits{K.}}:
\bbtitle{Machine Interpretation of Line Drawings}.
\bpublisher{MIT Press},
\blocation{Cambridge, MA}
(\byear{1986})
\end{bbook}
\endbibitem

\bibitem[\protect\citeauthoryear{Wilczkowiak et~al.}{2001}]{wilczkowiak2001}
\begin{bchapter}
\bauthor{\bsnm{Wilczkowiak}, \binits{M.}},
\bauthor{\bsnm{Boyer}, \binits{E.}},
\bauthor{\bsnm{Sturm}, \binits{P.}}:
\bctitle{Camera calibration and {3D} reconstruction from single images using parallelepipeds}.
In: \bbtitle{Proc. IEEE International Conference on Computer Vision (ICCV)},
pp. \bfpage{142}--\blpage{148}
(\byear{2001}).
\doiurl{10.1109/ICCV.2001.937510}
\end{bchapter}
\endbibitem

\bibitem[\protect\citeauthoryear{Wilczkowiak et~al.}{2002}]{wilczkowiak2002}
\begin{bchapter}
\bauthor{\bsnm{Wilczkowiak}, \binits{M.}},
\bauthor{\bsnm{Boyer}, \binits{E.}},
\bauthor{\bsnm{Sturm}, \binits{P.}}:
\bctitle{{3D} modelling using geometric constraints: A parallelepiped based approach}.
In: \bbtitle{Proc. European Conference on Computer Vision (ECCV)}.
\bsertitle{Lecture Notes in Computer Science},
vol. \bseriesno{2353},
pp. \bfpage{221}--\blpage{236}
(\byear{2002}).
\doiurl{10.1007/3-540-47979-1_15}
\end{bchapter}
\endbibitem

\bibitem[\protect\citeauthoryear{Wilczkowiak et~al.}{2005}]{wilczkowiak2005}
\begin{barticle}
\bauthor{\bsnm{Wilczkowiak}, \binits{M.}},
\bauthor{\bsnm{Sturm}, \binits{P.}},
\bauthor{\bsnm{Boyer}, \binits{E.}}:
\batitle{Using geometric constraints through parallelepipeds for calibration and {3D} modeling}.
\bjtitle{IEEE Transactions on Pattern Analysis and Machine Intelligence}
\bvolume{27}(\bissue{2}),
\bfpage{194}--\blpage{207}
(\byear{2005})
\doiurl{10.1109/TPAMI.2005.40}
\end{barticle}
\endbibitem

\bibitem[\protect\citeauthoryear{Coughlan and Yuille}{1999}]{coughlan1999}
\begin{bchapter}
\bauthor{\bsnm{Coughlan}, \binits{J.M.}},
\bauthor{\bsnm{Yuille}, \binits{A.L.}}:
\bctitle{{M}anhattan world: Compass direction from a single image by {B}ayesian inference}.
In: \bbtitle{Proc. IEEE International Conference on Computer Vision (ICCV)},
vol. \bseriesno{2},
pp. \bfpage{941}--\blpage{947}
(\byear{1999}).
\doiurl{10.1109/ICCV.1999.790349}
\end{bchapter}
\endbibitem

\bibitem[\protect\citeauthoryear{Schindler and Dellaert}{2004}]{schindler2004}
\begin{bchapter}
\bauthor{\bsnm{Schindler}, \binits{G.}},
\bauthor{\bsnm{Dellaert}, \binits{F.}}:
\bctitle{{A}tlanta world: An expectation maximization framework for simultaneous low-level edge grouping and camera calibration in complex man-made environments}.
In: \bbtitle{Proc. IEEE Conference on Computer Vision and Pattern Recognition (CVPR)},
vol. \bseriesno{1},
pp. \bfpage{203}--\blpage{209}
(\byear{2004}).
\doiurl{10.1109/CVPR.2004.1315033}
\end{bchapter}
\endbibitem

\bibitem[\protect\citeauthoryear{Caprile and Torre}{1990}]{caprile1990}
\begin{barticle}
\bauthor{\bsnm{Caprile}, \binits{B.}},
\bauthor{\bsnm{Torre}, \binits{V.}}:
\batitle{Using vanishing points for camera calibration}.
\bjtitle{International Journal of Computer Vision}
\bvolume{4}(\bissue{2}),
\bfpage{127}--\blpage{139}
(\byear{1990})
\doiurl{10.1007/BF00127813}
\end{barticle}
\endbibitem

\bibitem[\protect\citeauthoryear{Cipolla et~al.}{1999}]{cipolla1999}
\begin{bchapter}
\bauthor{\bsnm{Cipolla}, \binits{R.}},
\bauthor{\bsnm{Drummond}, \binits{T.}},
\bauthor{\bsnm{Robertson}, \binits{D.}}:
\bctitle{Camera calibration from vanishing points in image of architectural scenes}.
In: \bbtitle{Proc. British Machine Vision Conference (BMVC)},
pp. \bfpage{38}--\blpage{13810}
(\byear{1999}).
\doiurl{10.5244/C.13.38}
\end{bchapter}
\endbibitem

\bibitem[\protect\citeauthoryear{Ko{\v{s}}eck{\'a} and Zhang}{2002}]{kosecka2002}
\begin{bchapter}
\bauthor{\bsnm{Ko{\v{s}}eck{\'a}}, \binits{J.}},
\bauthor{\bsnm{Zhang}, \binits{W.}}:
\bctitle{Video compass}.
In: \bbtitle{Computer Vision -- ECCV 2002}.
\bsertitle{Lecture Notes in Computer Science},
vol. \bseriesno{2353},
pp. \bfpage{476}--\blpage{490}.
\bpublisher{Springer},
\blocation{Berlin, Heidelberg}
(\byear{2002}).
\doiurl{10.1007/3-540-47979-1_32}
\end{bchapter}
\endbibitem

\bibitem[\protect\citeauthoryear{Wildenauer and Hanbury}{2012}]{wildenauer2012}
\begin{bchapter}
\bauthor{\bsnm{Wildenauer}, \binits{H.}},
\bauthor{\bsnm{Hanbury}, \binits{A.}}:
\bctitle{Robust camera self-calibration from monocular images of {M}anhattan worlds}.
In: \bbtitle{Proc. IEEE Conference on Computer Vision and Pattern Recognition (CVPR)},
pp. \bfpage{2831}--\blpage{2838}
(\byear{2012}).
\doiurl{10.1109/CVPR.2012.6248008}
\end{bchapter}
\endbibitem

\bibitem[\protect\citeauthoryear{Criminisi et~al.}{2000}]{criminisi2000}
\begin{barticle}
\bauthor{\bsnm{Criminisi}, \binits{A.}},
\bauthor{\bsnm{Reid}, \binits{I.}},
\bauthor{\bsnm{Zisserman}, \binits{A.}}:
\batitle{Single view metrology}.
\bjtitle{International Journal of Computer Vision}
\bvolume{40}(\bissue{2}),
\bfpage{123}--\blpage{148}
(\byear{2000})
\doiurl{10.1023/A:1026598000963}
\end{barticle}
\endbibitem

\bibitem[\protect\citeauthoryear{Tardif}{2009}]{tardif2009}
\begin{bchapter}
\bauthor{\bsnm{Tardif}, \binits{J.-P.}}:
\bctitle{Non-iterative approach for fast and accurate vanishing point detection}.
In: \bbtitle{Proc. IEEE International Conference on Computer Vision (ICCV)},
pp. \bfpage{1250}--\blpage{1257}
(\byear{2009}).
\doiurl{10.1109/ICCV.2009.5459328}
\end{bchapter}
\endbibitem

\bibitem[\protect\citeauthoryear{Jelinek and Taylor}{2001}]{jelinek2001}
\begin{barticle}
\bauthor{\bsnm{Jelinek}, \binits{D.}},
\bauthor{\bsnm{Taylor}, \binits{C.J.}}:
\batitle{Reconstruction of linearly parameterized models from single images with a camera of unknown focal length}.
\bjtitle{IEEE Transactions on Pattern Analysis and Machine Intelligence}
\bvolume{23}(\bissue{7}),
\bfpage{767}--\blpage{773}
(\byear{2001})
\doiurl{10.1109/34.935850}
\end{barticle}
\endbibitem

\bibitem[\protect\citeauthoryear{Ramalingam et~al.}{2013}]{ramalingam2013}
\begin{bchapter}
\bauthor{\bsnm{Ramalingam}, \binits{S.}},
\bauthor{\bsnm{Pillai}, \binits{J.K.}},
\bauthor{\bsnm{Jain}, \binits{A.}},
\bauthor{\bsnm{Taguchi}, \binits{Y.}}:
\bctitle{{M}anhattan junction catalogue for spatial reasoning of indoor scenes}.
In: \bbtitle{Proc. IEEE Conference on Computer Vision and Pattern Recognition (CVPR)},
pp. \bfpage{3065}--\blpage{3072}
(\byear{2013}).
\doiurl{10.1109/CVPR.2013.394}
\end{bchapter}
\endbibitem

\bibitem[\protect\citeauthoryear{Hedau et~al.}{2009}]{hedau2009}
\begin{bchapter}
\bauthor{\bsnm{Hedau}, \binits{V.}},
\bauthor{\bsnm{Hoiem}, \binits{D.}},
\bauthor{\bsnm{Forsyth}, \binits{D.}}:
\bctitle{Recovering the spatial layout of cluttered rooms}.
In: \bbtitle{Proc. IEEE International Conference on Computer Vision (ICCV)},
pp. \bfpage{1849}--\blpage{1856}
(\byear{2009}).
\doiurl{10.1109/ICCV.2009.5459411}
\end{bchapter}
\endbibitem

\bibitem[\protect\citeauthoryear{Lee et~al.}{2009}]{lee2009}
\begin{bchapter}
\bauthor{\bsnm{Lee}, \binits{D.C.}},
\bauthor{\bsnm{Hebert}, \binits{M.}},
\bauthor{\bsnm{Kanade}, \binits{T.}}:
\bctitle{Geometric reasoning for single image structure recovery}.
In: \bbtitle{Proc. IEEE Conference on Computer Vision and Pattern Recognition (CVPR)},
pp. \bfpage{2136}--\blpage{2143}
(\byear{2009}).
\doiurl{10.1109/CVPR.2009.5206872}
\end{bchapter}
\endbibitem

\bibitem[\protect\citeauthoryear{Workman et~al.}{2015}]{workman2015}
\begin{bchapter}
\bauthor{\bsnm{Workman}, \binits{S.}},
\bauthor{\bsnm{Greenwell}, \binits{C.}},
\bauthor{\bsnm{Zhai}, \binits{M.}},
\bauthor{\bsnm{Baltenberger}, \binits{R.}},
\bauthor{\bsnm{Jacobs}, \binits{N.}}:
\bctitle{{DeepFocal}: A method for direct focal length estimation}.
In: \bbtitle{Proc. IEEE International Conference on Image Processing (ICIP)},
pp. \bfpage{1369}--\blpage{1373}
(\byear{2015}).
\doiurl{10.1109/ICIP.2015.7351024}
\end{bchapter}
\endbibitem

\bibitem[\protect\citeauthoryear{Hold-Geoffroy et~al.}{2018}]{holdgeoffroy2018}
\begin{bchapter}
\bauthor{\bsnm{Hold-Geoffroy}, \binits{Y.}},
\bauthor{\bsnm{Sunkavalli}, \binits{K.}},
\bauthor{\bsnm{Eisenmann}, \binits{J.}},
\bauthor{\bsnm{Fisher}, \binits{M.}},
\bauthor{\bsnm{Gambaretto}, \binits{E.}},
\bauthor{\bsnm{Hadap}, \binits{S.}},
\bauthor{\bsnm{Lalonde}, \binits{J.-F.}}:
\bctitle{A perceptual measure for deep single image camera calibration}.
In: \bbtitle{Proc. IEEE Conference on Computer Vision and Pattern Recognition (CVPR)},
pp. \bfpage{2354}--\blpage{2363}
(\byear{2018}).
\doiurl{10.1109/CVPR.2018.00250}
\end{bchapter}
\endbibitem

\bibitem[\protect\citeauthoryear{Jin et~al.}{2023}]{jin2023}
\begin{bchapter}
\bauthor{\bsnm{Jin}, \binits{L.}},
\bauthor{\bsnm{Zhang}, \binits{J.}},
\bauthor{\bsnm{Hold-Geoffroy}, \binits{Y.}},
\bauthor{\bsnm{Wang}, \binits{O.}},
\bauthor{\bsnm{Blackburn-Matzen}, \binits{K.}},
\bauthor{\bsnm{Sticha}, \binits{M.}},
\bauthor{\bsnm{Fouhey}, \binits{D.F.}}:
\bctitle{Perspective fields for single image camera calibration}.
In: \bbtitle{Proc. IEEE/CVF Conference on Computer Vision and Pattern Recognition (CVPR)},
pp. \bfpage{17307}--\blpage{17316}
(\byear{2023}).
\doiurl{10.1109/CVPR52729.2023.01660}
\end{bchapter}
\endbibitem

\bibitem[\protect\citeauthoryear{Veicht et~al.}{2024}]{veicht2024}
\begin{bchapter}
\bauthor{\bsnm{Veicht}, \binits{A.}},
\bauthor{\bsnm{Sarlin}, \binits{P.-E.}},
\bauthor{\bsnm{Lindenberger}, \binits{P.}},
\bauthor{\bsnm{Pollefeys}, \binits{M.}}:
\bctitle{{GeoCalib}: Learning single-image calibration with geometric optimization}.
In: \bbtitle{Computer Vision -- ECCV 2024}
(\byear{2024}).
\doiurl{10.1007/978-3-031-73661-2_1} .
\bcomment{arXiv:2409.06704}
\end{bchapter}
\endbibitem

\bibitem[\protect\citeauthoryear{Hartley}{1997}]{hartley1997}
\begin{barticle}
\bauthor{\bsnm{Hartley}, \binits{R.I.}}:
\batitle{Self-calibration of stationary cameras}.
\bjtitle{International Journal of Computer Vision}
\bvolume{22}(\bissue{1}),
\bfpage{5}--\blpage{23}
(\byear{1997})
\doiurl{10.1023/A:1007957826135}
\end{barticle}
\endbibitem

\bibitem[\protect\citeauthoryear{Sturm}{1997}]{sturm1997}
\begin{bchapter}
\bauthor{\bsnm{Sturm}, \binits{P.}}:
\bctitle{Critical motion sequences for monocular self-calibration and uncalibrated {E}uclidean reconstruction}.
In: \bbtitle{Proc. IEEE Conference on Computer Vision and Pattern Recognition (CVPR)},
pp. \bfpage{1100}--\blpage{1105}
(\byear{1997}).
\doiurl{10.1109/CVPR.1997.609467}
\end{bchapter}
\endbibitem

\bibitem[\protect\citeauthoryear{de~Agapito et~al.}{2001}]{deagapito2001}
\begin{barticle}
\bauthor{\bsnm{Agapito}, \binits{L.}},
\bauthor{\bsnm{Hayman}, \binits{E.}},
\bauthor{\bsnm{Reid}, \binits{I.}}:
\batitle{Self-calibration of rotating and zooming cameras}.
\bjtitle{International Journal of Computer Vision}
\bvolume{45},
\bfpage{107}--\blpage{127}
(\byear{2001})
\doiurl{10.1023/A:1012471930694}
\end{barticle}
\endbibitem

\bibitem[\protect\citeauthoryear{Kahl et~al.}{2000}]{kahl2000}
\begin{barticle}
\bauthor{\bsnm{Kahl}, \binits{F.}},
\bauthor{\bsnm{Triggs}, \binits{B.}},
\bauthor{\bsnm{{\AA}str{\"o}m}, \binits{K.}}:
\batitle{Critical motions for auto-calibration when some intrinsic parameters can vary}.
\bjtitle{Journal of Mathematical Imaging and Vision}
\bvolume{13}(\bissue{2}),
\bfpage{131}--\blpage{146}
(\byear{2000})
\doiurl{10.1023/A:1026524030731}
\end{barticle}
\endbibitem

\bibitem[\protect\citeauthoryear{Hartley and Zisserman}{2004}]{hartley2004}
\begin{bbook}
\bauthor{\bsnm{Hartley}, \binits{R.}},
\bauthor{\bsnm{Zisserman}, \binits{A.}}:
\bbtitle{Multiple View Geometry in Computer Vision},
\bedition{2nd} edn.
\bpublisher{Cambridge University Press},
\blocation{Cambridge}
(\byear{2004}).
\doiurl{10.1017/CBO9780511811685}
\end{bbook}
\endbibitem

\bibitem[\protect\citeauthoryear{Liebowitz and Zisserman}{1999}]{liebowitz1999}
\begin{bchapter}
\bauthor{\bsnm{Liebowitz}, \binits{D.}},
\bauthor{\bsnm{Zisserman}, \binits{A.}}:
\bctitle{Combining scene and auto-calibration constraints}.
In: \bbtitle{Proc. IEEE International Conference on Computer Vision (ICCV)},
vol. \bseriesno{1},
pp. \bfpage{293}--\blpage{300}
(\byear{1999}).
\doiurl{10.1109/ICCV.1999.791233}
\end{bchapter}
\endbibitem

\bibitem[\protect\citeauthoryear{Triggs}{1997}]{triggs1997}
\begin{bchapter}
\bauthor{\bsnm{Triggs}, \binits{B.}}:
\bctitle{Autocalibration and the absolute quadric}.
In: \bbtitle{Proc. IEEE Conference on Computer Vision and Pattern Recognition (CVPR)},
pp. \bfpage{609}--\blpage{614}
(\byear{1997}).
\doiurl{10.1109/CVPR.1997.609388}
\end{bchapter}
\endbibitem

\bibitem[\protect\citeauthoryear{Grammatikopoulos et~al.}{2004}]{grammatikopoulos2004}
\begin{bchapter}
\bauthor{\bsnm{Grammatikopoulos}, \binits{L.}},
\bauthor{\bsnm{Karras}, \binits{G.}},
\bauthor{\bsnm{Petsa}, \binits{E.}}:
\bctitle{Camera calibration combining images with two vanishing points}.
In: \bbtitle{International Archives of the Photogrammetry, Remote Sensing and Spatial Information Sciences},
vol. \bseriesno{XXXV-B5}
(\byear{2004}).
\bcomment{ISPRS Congress, Istanbul}.
\burl{https://www.isprs.org/proceedings/XXXV/congress/comm5/papers/530.pdf}
\end{bchapter}
\endbibitem

\bibitem[\protect\citeauthoryear{Denis et~al.}{2008}]{denis2008}
\begin{bchapter}
\bauthor{\bsnm{Denis}, \binits{P.}},
\bauthor{\bsnm{Elder}, \binits{J.H.}},
\bauthor{\bsnm{Estrada}, \binits{F.J.}}:
\bctitle{Efficient edge-based methods for estimating {M}anhattan frames in urban imagery}.
In: \bbtitle{Proc. European Conference on Computer Vision (ECCV)},
pp. \bfpage{197}--\blpage{210}
(\byear{2008}).
\doiurl{10.1007/978-3-540-88688-4_15}
\end{bchapter}
\endbibitem

\end{thebibliography}

\end{document}


\maketitle
\renewcommand{\thesection}{S\arabic{section}}
\renewcommand{\theequation}{S\arabic{equation}}
\renewcommand{\thefigure}{S\arabic{figure}}
\renewcommand{\thetable}{S\arabic{table}}
\renewcommand{\thetheorem}{S\arabic{theorem}}

\noindent This resource collects material that supports the article but is not needed to follow it, in three parts. Part~I gives further theory: the constraint from a corner and a parallel line (Sect.~S1); mirror resolution from two corners, counts and classical restatements for two corners, and what opacity adds (Sects.~S2--S4); further results on several frames (Sect.~S5); model error, for corners that are not right-angled and for lens distortion (Sect.~S6); the first-order noise analysis and the rules that combine bounds under noise (Sects.~S7 and S8); and junctions placed at random (Sect.~S9). Part~II reports the computations and experiments: the simulation protocol and the checks of the noise model (Sect.~S10); numerical checks of the proved results and worked examples of the convex program (Sect.~S11); the experiments on real images, with the limitations of the evaluation (Sect.~S12); and further experiments (Sect.~S13). Part~III lists directions for further work (Sect.~S14). Sections, equations, figures, tables and propositions of this resource carry the prefix S; references without it are to the article.

\part{Further theory}

\section{A corner and a parallel line}\label{sec:solver}

A single additional image line, the image of a scene line parallel to one edge of the corner, determines $f$ up to a twofold ambiguity, as derived below. The resulting constraint is classical; \cref{M-prop:general} shows how it arises.

\begin{proposition}[A corner and a parallel line]\label{prop:quadsolver}
Let a right-angle corner with image lines $\vl_1,\vl_2,\vl_3$, and one more image line $\vl_3'$ of a scene line parallel to the edge $\vd_3$, be seen by a camera, and let $\vv_3=\vl_3\times\vl_3'$. For any candidate camera $K$, with $\omega^{*}=KK^{\top}$ and $\omega=(\omega^{*})^{-1}$, the two sides of
\begin{equation}
\bigl(\vl_1^{\top}\omega^{*}\vl_2\bigr)\bigl(\vv_3^{\top}\omega\,\vv_3\bigr)=(\vl_1\cdot\vv_3)(\vl_2\cdot\vv_3),
\label{eq:cfsolver}
\end{equation}
differ by $\det(K)^2\,\vv_1^{\top}\omega\vv_2$, where $\vv_i=\vl_i\times\omega\vv_3$ is the point in which edge $i$ meets the polar of $\vv_3$. Hence \eqref{eq:cfsolver} holds exactly when the candidate camera makes $\vv_1$ and $\vv_2$ orthogonal, as the true camera does. For $K=\operatorname{diag}(f,f,1)$ the condition, multiplied by $f^2$, is a quadratic in $f^2$, so there are at most two solutions unless it vanishes identically.
\end{proposition}
\begin{proof}
The lines $\vl_3$ and $\vl_3'$ are images of parallel scene lines, so they meet at the vanishing point $\vv_3$ of $\vd_3$. Its polar $\vh=\omega\vv_3$ is the vanishing line of the planes perpendicular to $\vd_3$, and the edges $\vd_1$ and $\vd_2$ lie in such a plane, so their vanishing points are $\vv_i=\vl_i\times\vh$, and the two edges are orthogonal exactly when $\vv_1^{\top}\omega\vv_2=0$. For a symmetric invertible matrix,
\[
\begin{aligned}
(\mathbf a\times\mathbf b)^{\top}\operatorname{adj}(\omega^{*})(\mathbf c\times\mathbf d)={}&(\mathbf a^{\top}\omega^{*}\mathbf c)(\mathbf b^{\top}\omega^{*}\mathbf d)\\
&-(\mathbf a^{\top}\omega^{*}\mathbf d)(\mathbf b^{\top}\omega^{*}\mathbf c),
\end{aligned}
\]
and $\operatorname{adj}(\omega^{*})=\det(K)^2\omega$. Taking $\mathbf a=\vl_1$, $\mathbf c=\vl_2$ and $\mathbf b=\mathbf d=\vh$, and using $\omega^{*}\vh=\vv_3$, gives $\det(K)^2\,\vv_1^{\top}\omega\vv_2=(\vl_1^{\top}\omega^{*}\vl_2)(\vv_3^{\top}\omega\vv_3)-(\vl_1\cdot\vv_3)(\vl_2\cdot\vv_3)$. For $\omega^{*}=\operatorname{diag}(f^2,f^2,1)$ the first factor is linear in $f^2$ and $f^2\,\vv_3^{\top}\omega\vv_3$ is linear in $f^2$, which gives the quadratic.
\end{proof}

This is the three-direction case of \citet{wildenauer2012}, who state the constraint as the orthogonality of $\vv_1$ and $\vv_2$; we claim only the equivalence. The corner's own sign test cannot select between the two roots. At a root, $\vv_1$, $\vv_2$ and $\vv_3$ are mutually orthogonal under the candidate camera, and $K^{-1}\vv_i$ lies in the interpretation plane of edge $i$; the candidate therefore admits the corner, and \eqref{M-eq:sign} holds automatically. Only the sign test at a second corner can reject a root (Sect.~S13). The derivation in the corner parameterization, together with its degenerate configurations and its conditioning, follows.

\subsection{Derivation in the corner parameterization}\label{sec:Ssolver}

Two right-angle corners that share only their vertical direction, as two objects standing on one floor do, are determined with one equation to spare when $f$ is unknown (\cref{tab:counts}). This section reduces that problem to a quadratic in $f^2$. The reduction shows that the second corner's horizontal lines enter only through the spare equation, so the minimal data are one corner and one more image line parallel to one of its edges; a second corner is one convenient source of that line. The constraint itself is not new. \citet{wildenauer2012} solve for three orthogonal vanishing points and the focal length from four lines aligned with either two or three orthogonal directions, and their three-direction case is the present one: four lines, one direction seen twice and two seen once, yielding a quadratic in $f^2$. In vanishing-point form (\cref{sec:elimination}) our \eqref{eq:quadratic} is that constraint. We derive it in the corner parameterization rather than in the vanishing points for one reason: it keeps the junction structure, which makes the sign test of \cref{M-sec:sign} available as a root filter at a second corner (at the corner that defines the roots it always passes, as \cref{sec:solver} of the article shows), and that is what \cref{sec:e8} tests. We claim no new algebra: \cref{prop:quadsolver} of the article states the condition in coordinate-free form and proves it identical to the classical vanishing-point constraint. The problem is small enough to solve by hand; the general machinery for minimal problems \citep{kukelova2008,larsson2017} is not needed here.

\subsubsection{Setup}

Label each corner's edges so that edge 3 is the shared vertical direction $\vd_3$ and edges 1 and 2 are horizontal. The second corner's horizontal edges are those of the first rotated by an unknown yaw $\psi$ about $\vd_3$. Let $\vl_1,\vl_2,\vl_3$ be the first corner's image lines and $\vl_1',\vl_2',\vl_3'$ the second's, with $\vn_i=K^{\top}\vl_i=(fa_i,fb_i,e_i)$ and likewise $\vn_i'$, all linear in $f$. The six constraints are $\vn_i\cdot\vd_i=0$ and $\vn_i'\cdot\vd_i'=0$ with $\vd_3'=\vd_3$, $\vd_1'=\cos\psi\,\vd_1+\sin\psi\,\vd_2$ and $\vd_2'=-\sin\psi\,\vd_1+\cos\psi\,\vd_2$.

\subsubsection{Elimination}\label{sec:elimination}

The vertical direction lies in both vertical interpretation planes, so
\begin{equation}
\vd_3\propto\vw=\vn_3\times\vn_3',
\end{equation}
a function of $f$ alone, quadratic in $f$. The horizontal edges of the first corner lie in the plane perpendicular to $\vd_3$; write $\vd_1=\cos\theta\,\vh_1+\sin\theta\,\vh_2$ and $\vd_2=-\sin\theta\,\vh_1+\cos\theta\,\vh_2$ for an orthonormal basis $\vh_1,\vh_2$ of that plane. The constraints $\vn_1\cdot\vd_1=0$ and $\vn_2\cdot\vd_2=0$ each give $\tan\theta$, namely $-(\vn_1\cdot\vh_1)/(\vn_1\cdot\vh_2)$ and $(\vn_2\cdot\vh_2)/(\vn_2\cdot\vh_1)$, and the two agree if and only if $(\vn_1\cdot\vh_1)(\vn_2\cdot\vh_1)+(\vn_1\cdot\vh_2)(\vn_2\cdot\vh_2)=0$, that is, $\vn_1^{\top}(I-\vd_3\vd_3^{\top})\vn_2=0$. Substituting $\vd_3=\vw/|\vw|$ and clearing the denominator,
\begin{equation}
E(f)=(\vn_1\cdot\vn_2)\,|\vw|^2-(\vn_1\cdot\vw)(\vn_2\cdot\vw)=0 .
\label{eq:E}
\end{equation}
$E$ is a polynomial of degree six in $f$, and even: $f\mapsto-f$ reverses the first two components of every $\vn$ and of $\vw$, so every term is even. It is therefore a cubic in $x=f^2$, but one root is spurious. Write $\vv_3=\vl_3\times\vl_3'=(p,q,s)$, the vertical vanishing point, which does not depend on $f$; then $\vw=(fp,\,fq,\,f^2s)$, and with $m_i=\vl_i\cdot\vv_3$ and $A_{12}=a_1a_2+b_1b_2$ as in \eqref{M-eq:lin}, direct expansion gives
\begin{equation}
\begin{aligned}
E&=x\,\bigl(c_2x^2+c_1x+c_0\bigr),\\
c_2&=A_{12}\,s^2,\\
c_1&=A_{12}(p^2+q^2)+e_1e_2\,s^2-m_1m_2,\\
c_0&=e_1e_2\,(p^2+q^2)
\end{aligned}
\label{eq:coeffs}
\end{equation}
(verified symbolically). The factor $x$ is forced: at $f=0$ every $\vn_i$ is $(0,0,e_i)$, so $\vw=0$ and $E=0$. That root carries no information, and the focal length is a root of the quadratic
\begin{equation}
c_2x^2+c_1x+c_0=0,
\label{eq:quadratic}
\end{equation}
in closed form, with at most two positive solutions. The yaw follows from $\theta$ and one horizontal line of the second corner, and the equation with the roles of the corners exchanged,
\begin{equation}
E'(f)=(\vn_1'\cdot\vn_2')\,|\vw|^2-(\vn_1'\cdot\vw)(\vn_2'\cdot\vw),
\end{equation}
is the spare equation that selects between the roots.

\paragraph{A geometric form.} The coefficients \eqref{eq:coeffs} are opaque as written, but the same reduction that simplified \cref{M-sec:bound} applies here too and leaves every term a distance between a point and a line. Let $\vv_3=\vl_3\times\vl_3'$ be the vertical vanishing point, with $|\vv_3|$ its distance from the principal point, and write
\[
e_i=\vl_i\cdot(0,0,1),\qquad d_i=\vl_i\cdot(\vv_3/s,\,1),
\]
the signed distances from the principal point and from the vanishing point to the image line $\vl_i$, which with $a_i^2+b_i^2=1$ are distances in pixels. Then $d_i=(a_i,b_i)\cdot(\vv_3/s-\vc)$, and \eqref{eq:E} becomes
\begin{equation}
\bigl(A_{12}f^2+e_1e_2\bigr)\bigl(f^2+|\vv_3|^2\bigr)=d_1d_2\,f^2 .
\label{eq:geosolver}
\end{equation}
The left factor $A_{12}f^2+e_1e_2$ is the pair factor of \eqref{M-eq:Lprod} for the two horizontal edges, and $f^2+|\vv_3|^2$ is the companion of the $f^2+\rho^2$ that cancelled there, with the vanishing point in place of the corner. Dividing through by $f^2+\rho^2$ and using \eqref{M-eq:lin} gives the equivalent form
\begin{equation}
g_{12}\,\bigl(f^2+|\vv_3|^2\bigr)=d_1d_2\,\bigl(f^2+\rho^2\bigr).
\label{eq:geosolver2}
\end{equation}
Written this way the degenerate configurations are immediate. $A_{12}=0$, a raw right angle between the horizontal edges, kills the leading coefficient and leaves a linear equation; $e_1e_2=0$, a horizontal edge line through the principal point, kills the constant term; $s=0$ sends $\vv_3$ to infinity and is treated separately below. One case ties the solver back to \cref{M-sec:bound}: if $d_1d_2=0$, so that the vertical vanishing point lies on one of the two horizontal edge lines, \eqref{eq:geosolver} reduces to $A_{12}f^2+e_1e_2=0$ and the solver returns the single-corner threshold $f_{12}$ of \cref{M-thm:interval}. The second corner then adds nothing the first did not already give, and the configuration should be rejected rather than solved.

\paragraph{The coordinate-free form.} The geometric form generalizes directly to an arbitrary calibration matrix: it is the condition \eqref{eq:cfsolver} of the article, which for $\omega^{*}=\operatorname{diag}(f^2,f^2,1)$ is \eqref{eq:geosolver}. Precisely, the polynomial \eqref{eq:E} equals $\det(K)^2$ times the difference of the two sides of \eqref{eq:cfsolver}, for every upper-triangular $K$. The article shows that this is the classical orthogonality of two vanishing points; the degeneracy $d_1d_2=0$ is the vanishing point falling on an edge line, which is classical too.

\begin{proposition}[Corner-plus-line solver]\label{prop:Squad}
Let a right-angle corner with image lines $\vl_1,\vl_2,\vl_3$, and one more image line $\vl_3'$ of a scene line parallel to the edge $\vd_3$, be seen by a camera with square pixels, zero skew and known principal point. Then $f^2$ is a root of the quadratic \eqref{eq:quadratic}, whose coefficients \eqref{eq:coeffs} are polynomial in the four lines, so there are at most two solutions. If $\vl_3'$ belongs to a second corner that shares only $\vd_3$, one of that corner's horizontal lines fixes the yaw $\gamma$, and the two together enter $E'$, the polynomial \eqref{eq:E} formed from the second corner's two horizontal lines in place of the first's, which is the one check.
\end{proposition}

Geometrically, $E=0$ says that the two horizontal edge directions, each the intersection of its interpretation plane with the horizontal plane determined by $\vw$, are orthogonal. In vanishing-point terms, $\vv_3$ is known without $f$; the horizon is its polar line $\omega\vv_3\propto(p,q,f^2s)$, which depends on $f$; the horizontal vanishing points $\vv_1,\vv_2$ are the intersections of $\vl_1$ and $\vl_2$ with the horizon; and \eqref{eq:quadratic} is their orthogonality condition $\vv_1^{\top}\omega\vv_2=0$. If the horizon were known in pixels, the same condition would be linear in $f^2$; here its one free parameter is paid for by the second vertical line. A gravity sensor does not give the horizon in pixels. It gives the vertical $\vg$ in camera coordinates, and the image of the horizon, $K^{-\top}\vg\propto(g_x,g_y,f g_z)$, still depends on $f$ (\cref{sec:variants}).

\paragraph{Degenerate configurations.} Each corresponds to a vanishing coefficient in \eqref{eq:coeffs}. $\vv_3=0$, the two vertical lines coincide: $\vw\equiv0$, $E\equiv0$, no information. $s=0$, the vertical lines parallel and the vertical vanishing point at infinity (camera exactly level): $c_2=0$ and \eqref{eq:quadratic} is linear with the single root $x=-c_0/c_1$; this is the limit of the ill-conditioning measured in \cref{sec:conditioning}. $A_{12}=0$, a raw right angle at corner 1: again $c_2=0$. $e_1e_2=0$, an edge of corner 1 through the principal point: $c_0=0$, so $x=0$ becomes a double root and the one informative root is $-c_1/c_2$. $c_1^2=4c_0c_2$: a double root, with its usual sensitivity. The single-corner degeneracies of \cref{M-sec:sign} apply to each corner as before.

\paragraph{Root validation.} For each positive root $x$ of \eqref{eq:quadratic}, set $f=\sqrt{x}$ and $\vd_3=\vw/|\vw|$; recover $\theta$ from $\vn_1$ (its agreement with $\vn_2$ is \eqref{eq:E} itself and holds to rounding); when a second corner is present, recover $\psi$ from $\vn_1'$; evaluate the residual of $\vn_2'\cdot\vd_2'=0$, equivalently $E'$; and run the sign test of \cref{M-sec:sign} at both corners with the recovered $K$. Keep the root with the smaller residual, and reject the pair when both residuals exceed a threshold set by the line noise or when either corner fails the sign test. In a RANSAC loop (\cref{sec:ransac}) the residual is the per-sample score. How often the second root is removed by the sign test alone, and how often $E'$ is needed, is reported in \cref{sec:e8}.

\subsubsection{Variants}\label{sec:variants}

\paragraph{Gravity known.} With an inertial sensor giving the vertical $\vg$ in camera coordinates, $\vd_3=\vg$ is known without a second vertical line. The corner's own vertical line then gives $f$ linearly, from $\vn_3\cdot\vg=f\,(a_3g_x+b_3g_y)+e_3g_z=0$, unless that line passes through the principal point. The horizontal pair gives $\vn_1^{\top}(I-\vg\vg^{\top})\vn_2=0$, which is a quadratic in $f$ rather than in $f^2$, because $\vg$, unlike $\vw$, does not scale with $f$. It serves as an independent estimate or as a check. This is a one-corner focal-length solver, at the price of a sensor whose rotation to the camera is calibrated.

\paragraph{Shared frame.} If the two corners share all three axes, each direction has two lines, the three vanishing points are known, and the problem is the linear one of \cref{sec:vp}; the quadratic is not needed.

\subsubsection{Conditioning}\label{sec:conditioning}

The solver is exact at zero noise, and its accuracy depends on the angle between the two vertical image lines, which governs the conditioning of $\vw$. When the camera is nearly level the vertical lines are nearly parallel, $\vv_3$ is far away, and small direction errors move $\vw$ a great deal; the horizon is then nearly independent of $f$ and most of the information about $f$ comes from the horizontal pair. \Cref{sec:e8} measures this: with $0.5$\,px endpoint noise on $150$\,px edges the median relative error in $f$ is 1.9\% when the vertical lines differ by more than $15^\circ$ and 13\% when they differ by less than $2^\circ$. The fraction of pairs for which the quadratic has no positive root rises in the same direction. A RANSAC loop over detected junctions should therefore prefer corner pairs with well-separated vertical lines, or fall back to the gravity variant where a sensor is available.


\section{Mirror resolution from two corners}\label{sec:mirror}\label{sec:two}

The results so far concern the camera. The orientation recovered in \cref{M-sec:closed} remains ambiguous up to the mirror of \cref{M-sec:mirror1}, and no information about the camera can resolve it, since the mirror acts within a fixed $K$. This section shows that, under perspective and with the camera calibrated, a second corner of the same frame does resolve it. Degree-of-freedom counts for the other two-corner configurations, and the classical calibration from three vanishing points that two corners of one frame make available, are given in Sect.~S3.

Two corners of one box or room share one Manhattan frame, but in general not one \emph{oriented} frame. When edges are matched by direction, as they are when grouped by vanishing point, the edge of corner 2 along a given axis may point opposite to the edge of corner 1, as happens at the two ends of one box edge. Let $R$ be the orthonormal matrix with columns $\vd_1,\vd_2,\vd_3$ of corner 1; corner 2 then has frame $RD$ for a diagonal sign matrix $D$ that the image does not reveal. With $K$ known, \cref{M-sec:closed} returns each corner's frame up to its own mirror: the alternative for corner $i$ is $H_i$ times the truth, where $H_i=I-2\hat\vr_i\hat\vr_i^{\top}$ is the reflection in the plane through $O$ perpendicular to its ray. We call two hypotheses, one per corner, \emph{consistent} if the two frames agree up to a diagonal sign matrix, which is all that unoriented matching can require. Each corner on its own leaves two candidate frames, its true frame and its mirror image, so a pair of corners leaves four combinations; the question is which of them are consistent.

\begin{theorem}[Mirror resolution from two corners]\label{thm:mirror}
Let two nondegenerate right-angle corners (\cref{M-sec:sign}) of one Manhattan frame be seen by a calibrated camera along rays $\vr_1,\vr_2$, with the correspondence between the two corners' edges known and edges matched as unoriented axes, and let $R$, $D$, $H_i$ and consistency be as above. If $\vr_1\nparallel\vr_2$, then of the four hypothesis pairs $(R,RD)$, $(H_1R,RD)$, $(R,H_2RD)$ and $(H_1R,H_2RD)$, only the true pair $(R,RD)$ is consistent. If $\vr_1\parallel\vr_2$, the double flip $(H_1R,H_2RD)$ is consistent as well.
\end{theorem}
\begin{proof}
A pair $(G_1R,\,G_2RD)$ with $G_i\in\{I,H_i\}$ is consistent when the two frames agree up to a diagonal sign matrix, that is when $G_2^{-1}G_1=RD''R^{\top}$ for some diagonal $D''$ with entries in $\{+1,-1\}$. The three nontrivial pairs are classified by the determinant.

\emph{Single flips.} $\det H_1=-1$, so $(H_1R,RD)$ is consistent only if $\det D''=-1$. A matrix $RD''R^{\top}$ of determinant $-1$ is either the reflection in the plane perpendicular to a frame axis $\vd_k$, when $D''$ has one entry $-1$, or $-I$, when it has three. The latter has eigenvalues $(-1,-1,-1)$ and is not a reflection, so the only possibility is $H_1=I-2\hat\vr_1\hat\vr_1^{\top}=I-2\vd_k\vd_k^{\top}$, which forces $\hat\vr_1=\pm\vd_k$: an edge along the ray, excluded for a nondegenerate corner. The same argument rules out $(R,H_2RD)$.

\emph{The double flip.} Here $\det(H_2H_1)=+1$, so $D''$ has an even number of $-1$ entries and $RD''R^{\top}$ is either $I$ or the rotation by $180^\circ$ about a frame axis. As a composition of two reflections, $H_2H_1$ is the rotation about $\vr_1\times\vr_2$ by twice the angle between the rays. It equals $I$ exactly when $\vr_1\parallel\vr_2$, which is the stated exception. It equals a $180^\circ$ rotation about $\vd_k$ exactly when the rays are $90^\circ$ apart and $\vr_1\times\vr_2\parallel\vd_k$; but then both rays and the other two axes lie in the plane perpendicular to $\vd_k$, so at each corner those two edges share one image line and both corners are degenerate.
\end{proof}

The correspondence between the edges is essential: if $\hat\vr_1\propto\vd_1-\vd_2$, the reflection $H_1$ exchanges two axes, and without correspondence a single flip is consistent. With correspondence, one matched axis suffices.

\begin{remark}[One matched axis]\label{cor:oneaxis}
Under the hypotheses of \cref{thm:mirror}, let only one axis $\vd_k$ be matched between the corners. The single flips are inconsistent, and the double flip is consistent if and only if the rays are parallel, or $90^\circ$ apart with $\vd_k$ in their plane.
\end{remark}
\begin{proof}
A hypothesis pair is now consistent when it maps $\vd_k$ to $\pm\vd_k$. A single flip does so only if $H_i\vd_k=\vd_k-2(\hat\vr_i\cdot\vd_k)\hat\vr_i$ is parallel to $\vd_k$, that is, if $\hat\vr_i\parallel\vd_k$ or $\hat\vr_i\perp\vd_k$: edge $k$ along the ray, or the other two edges in a face that contains the ray, so that they share an image line; both are degenerate. The double flip $H_2H_1$ is the rotation about $\vr_1\times\vr_2$ by twice the angle between the rays. It maps $\vd_k$ to $\pm\vd_k$ exactly when it is the identity, so that the rays are parallel; when $\vd_k$ lies along its axis, which makes $\vd_k$ perpendicular to both rays and both corners degenerate; or when it is a half-turn, the rays $90^\circ$ apart, and $\vd_k$ is perpendicular to its axis, that is, in the plane of the rays.
\end{proof}

 The theorem assumes the camera known. For a candidate camera, the rays and reflections are those that the candidate defines; the experiments of \cref{sec:e7} use the true camera.

In vanishing-point terms the conclusion is unsurprising: the two image lines of each axis meet at its vanishing point, which fixes the axis. Under orthography, by contrast, they are parallel, the mirror is the image plane for every corner, and the skeleton of a polyhedron keeps exactly two interpretations \citep[Proposition~4]{kanatani1986}. What the proof adds is a classification of the failures: a single flip survives only when an edge points along a ray, the double flip only when the rays coincide, and the remaining case, rays $90^\circ$ apart, is excluded because the junctions then degenerate in the image. \citet{shakunaga1989} used this consistency in practice. Exact uniqueness, however, is not stability: the double-flip hypothesis differs from the true pair by an angle that shrinks as the rays approach each other, so the resolution is weakest for corners near each other (\cref{sec:e7}).


\subsection{Simulation}\label{sec:e7}
\begin{figure}[h]
\centering
\includegraphics[width=0.55\linewidth]{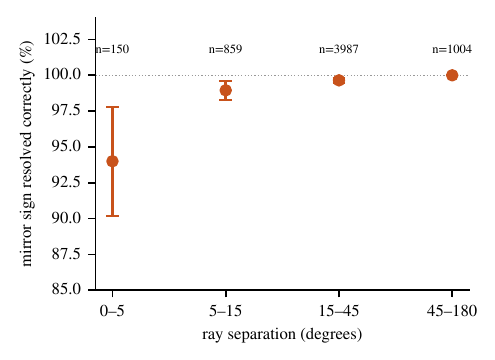}
\caption{Fraction of two-corner pairs of one frame whose mirror sign is resolved correctly, by ray separation, with 95\% binomial intervals and bin counts ($60^\circ$, $\sigma=0.5$\,px, $6000$ pairs); the vertical axis starts at 85\%}
\label{fig:Smirror}
\end{figure}

Two hypothesized frames are scored by their disagreement, the mean over the three axes of the angle between corresponding columns taken as unoriented lines; the disagreement is zero exactly when the frames agree up to a diagonal sign matrix. The hypothesis pair with the smallest disagreement is selected, and its margin is the gap in that score to the second-best pair. Each of $6000$ pairs consists of one random frame seen at two vertices placed independently in the image at depths $2$ to $6$, the second with its edges reoriented at random, at $60^\circ$ and $\sigma=0.5$\,px. The selected pair was the true pair in 99.5\% of cases (Wilson 95\% interval 99.3 to 99.6\%). Accuracy and margin both fell as the rays approached each other, accuracy to $94.0\%$ below $5^\circ$ of separation (89.0 to 96.8\%, $n=150$) (\cref{fig:Smirror}). The residual failures are not sign errors but pairs in which a corner's own closed-form frame is already off by more than the margin: the degradation at small separation reflects the stability limit of \cref{thm:mirror}, not a failure of the theorem.

\section{Two corners: counts, configurations and classical restatements}\label{sec:Stwo}

A second corner changes what is determined in three distinct ways, depending on whether it shares the first corner's Manhattan frame, whether an edge connects them, and whether $f$ is known. This section counts at the level of image lines, below the six-vertex threshold at which the parallelepiped method applies.

\subsection{The counting rule}

The projection $\vc_i$ of a corner point fixes its ray but not its depth, and the image line of an edge through $\vc_i$ does not depend on that depth. Corner positions therefore drop out of the count, and the observations that constrain rotation and intrinsics are the image line directions: three per unconnected corner. The unknowns are the rotations and whatever intrinsics are free. \Cref{tab:counts} lists the configurations and \cref{fig:configs} illustrates three of them.

The counts are generic: they hold for configurations outside the algebraic subsets on which an observation becomes dependent on the others (coincident lines, a corner at the principal point, the degeneracies named in each section). A net count of zero or more says that the unknowns are locally identifiable from the observations, not that the solution is unique or well conditioned. The number of algebraic solutions, and which of them survive the sign test, is established separately where a solver is given (two solutions in \cref{M-sec:closed}, at most two in \cref{sec:solver}), and conditioning is measured in \cref{M-sec:experiments}. A negative count says that a one-parameter family survives, which is the admissible-set case of \cref{M-sec:bound}.

\begin{table*}[t]
\centering\small
\caption{Degree-of-freedom count for one and two right-angle corners. Observations are image line directions, and net is observations minus unknowns. For the connected pair the shared edge is a single line, and the corner positions enter only through the relative depth.}
\label{tab:counts}
\begin{tabularx}{\linewidth}{>{\raggedright\arraybackslash}X c c c >{\raggedright\arraybackslash}X}
\toprule
Configuration & Unknowns & Observations & Net & Determined \\
\midrule
One corner, $K$ known & 3 & 3 & 0 & $R$ up to mirror \citep{kanatani1988} \\
One corner, $f$ unknown & 4 & 3 & $-1$ & lower bound on $f$, upper bound if raw angles fail the orthographic test; $R(f)$ (\cref{M-sec:bound}) \\
Three nonconcurrent lines of one frame, $K$ known & 3 & 3 & 0 & $R$, at most two solutions (\cref{prop:threelines}) \\
Two corners, same frame, $K$ known & 3 & 6 & $+3$ & $R$; mirror resolved; 3 checks \\
Two corners, same frame, $f$ unknown & 4 & 6 & $+2$ & $f$ and $R$; 2 checks \\
Two corners, same frame, $f$ and principal point unknown & 6 & 6 & 0 & $f$, principal point, $R$ \citep{caprile1990} \\
One corner and one line parallel to one of its edges, $f$ unknown & 4 & 4 & 0 & $f$ (quadratic in $f^2$, at most two roots) and $R$ (\cref{sec:solver}) \\
Two corners sharing only the up axis, $f$ unknown & 5 & 6 & $+1$ & $f$ from four of the lines, both rotations; 1 check (\cref{sec:solver}) \\
Two corners, independent frames, $f$ unknown & 7 & 6 & $-1$ & intersection of two admissible sets \\
Two corners connected by an edge, same frame, $f$ unknown & 4 & 5 & $+1$ & $f$, $R$; relative depth of the corners from their positions \\
Six or more matched vertices of one box & \multicolumn{3}{c}{linear in a 5-parameter upgrade} & $f$, principal point, box shape \citep{wilczkowiak2001} \\
\bottomrule
\end{tabularx}
\end{table*}

\begin{figure}[t]
\centering
\begin{tikzpicture}[scale=0.86,font=\footnotesize]
  \def\boxdraw#1#2{%
    \draw (0,0) rectangle (#1,#2);
    \draw (0,#2) -- ++(0.42,0.30) -- ++(#1,0) -- (#1,#2);
    \draw (#1,0) -- ++(0.42,0.30) -- ++(0,#2);}
  \begin{scope}[shift={(0,0)}]
    \boxdraw{1.7}{1.15}
    \draw[very thick] (0,1.15) -- (1.7,1.15);
    \fill (0,1.15) circle (2pt) (1.7,1.15) circle (2pt);
    \node at (1.05,-0.45) {(a)};
  \end{scope}
  \begin{scope}[shift={(3.0,0)}]
    \boxdraw{0.95}{1.25}
    \fill (0.95,1.25) circle (2pt);
    \begin{scope}[shift={(1.6,0.05)}]
      \boxdraw{0.8}{0.7}
      \fill (0.8,0.7) circle (2pt);
    \end{scope}
    \node at (1.4,-0.45) {(b)};
  \end{scope}
  \begin{scope}[shift={(6.6,0)}]
    \boxdraw{0.95}{1.2}
    \fill (0.95,1.2) circle (2pt);
    \begin{scope}[shift={(1.75,0.35)},rotate=27]
      \boxdraw{0.85}{0.65}
      \fill (0.85,0.65) circle (2pt);
    \end{scope}
    \node at (1.3,-0.45) {(c)};
  \end{scope}
\end{tikzpicture}
\caption{Two-corner configurations, the two corners used being the marked vertices: (a) connected by an edge, (b) one Manhattan frame, (c) independent frames. With $f$ unknown and the principal point known, the net counts of \cref{tab:counts} are $+1$, $+2$ and $-1$ respectively}
\label{fig:configs}
\end{figure}

\begin{proposition}[Three nonconcurrent lines]\label{prop:threelines}
Three image lines of one Manhattan frame, one per axis and not concurrent, with $K$ known, admit at most two orthonormal frames.
\end{proposition}
\begin{proof}
Write $\vn_i=K^{\top}\vl_i$. Then $\vd_1$ lies in the plane perpendicular to $\vn_1$, so $\vd_1=\cos\alpha\,\vh_1+\sin\alpha\,\vh_2$ for a fixed orthonormal basis $\vh_1,\vh_2$ of that plane, the choice of which affects only the origin of $\alpha$. Since $\vd_2\perp\vn_2$ and $\vd_2\perp\vd_1$, we have $\vd_2\parallel\vn_2\times\vd_1$ provided $\vd_1\nparallel\vn_2$. The excluded orientations form at most two values of $\alpha$ in $[0,\pi)$, namely those with $\vd_1=\pm\vn_2/|\vn_2|$, which require $\vn_2\perp\vn_1$; at such an $\alpha$ the axis $\vd_2$ is unconstrained by $\vn_2$ and the configuration is degenerate in the sense of \cref{M-sec:sign}. Away from them,
\[
\vd_3=\vd_1\times\vd_2\parallel\vd_1\times(\vn_2\times\vd_1)=\vn_2-\vd_1(\vd_1\cdot\vn_2).
\]
The remaining constraint $\vn_3\cdot\vd_3=0$ is therefore
\[
\vn_2\cdot\vn_3-(\vn_3\cdot\vd_1)(\vn_2\cdot\vd_1)=0 ,
\]
a quadratic form in $(\cos\alpha,\sin\alpha)$ and hence a trigonometric polynomial of degree two, with at most four roots in $[0,2\pi)$. Replacing $\alpha$ by $\alpha+\pi$ sends $\vd_1$ to $-\vd_1$ and gives the same frame, so at most two are distinct.
\end{proof}
Over $20\,000$ random configurations the count was two in all but six, and never more, so the bound is attained generically. \Citet{shakunaga1989} report the same count.

\subsection{Same frame, \texorpdfstring{$f$}{f} unknown: vanishing points (restated)}\label{sec:vp}

When two corners share a frame, each direction is seen as two lines whose intersection is its vanishing point $\vv_i=K\vd_i$, and orthogonality becomes $\vv_i^{\top}\omega\,\vv_j=0$ with $\omega=(KK^{\top})^{-1}$ \citep{hartley2004}. With square pixels and zero skew $\omega$ has three free parameters, so three orthogonal vanishing points calibrate $f$ and the principal point exactly. The principal point $\mathbf{x}_0$ is the orthocenter of the triangle $\vv_1\vv_2\vv_3$ and $f^2=-(\vv_i-\mathbf{x}_0)\cdot(\vv_j-\mathbf{x}_0)$ for any pair, with the vanishing points in pixel coordinates \citep{caprile1990}; see \cref{fig:vp}. Six lines, two per direction, are the minimum for which this works, and two corners provide that. The $+2$ of \cref{tab:counts} is the pair of checks left over when the principal point is known.

\begin{figure}[t]
\centering
\begin{tikzpicture}[scale=1.02,font=\footnotesize]
  \coordinate (v1) at (-3.3,-0.55);
  \coordinate (v2) at (3.5,-0.85);
  \coordinate (v3) at (0.45,3.55);
  \coordinate (f1) at ($(v2)!(v1)!(v3)$);
  \coordinate (f2) at ($(v1)!(v2)!(v3)$);
  \coordinate (f3) at ($(v1)!(v3)!(v2)$);
  \coordinate (H) at (intersection of v1--f1 and v2--f2);
  \coordinate (ca) at (-0.75,0.35);
  \coordinate (cb) at (1.15,0.95);
  \draw[densely dotted] (v1) -- (f1) (v2) -- (f2) (v3) -- (f3);
  \foreach \c in {ca,cb} \foreach \v in {v1,v2,v3} \draw[black!45] (\c) -- (\v);
  \foreach \c in {ca,cb} {
    \foreach \v in {v1,v2,v3} \draw[very thick] (\c) -- ($(\c)!0.55cm!(\v)$);
    \fill (\c) circle (1.5pt); }
  \fill (v1) circle (1.6pt) (v2) circle (1.6pt) (v3) circle (1.6pt);
  \draw[fill=white] (H) circle (1.7pt);
  \node[below=1pt] at (v1) {$\vv_1$};
  \node[below=1pt] at (v2) {$\vv_2$};
  \node[above=1pt] at (v3) {$\vv_3$};
  \node[below=2pt] at (H) {$\mathbf x_0$};
\end{tikzpicture}
\caption{Two corners of one frame give three vanishing points whose orthocenter is the principal point. One corner alone fixes only the line on which each vanishing point lies.}
\label{fig:vp}
\end{figure}

The figure also shows why one corner is not enough: each of its lines passes through one vanishing point, but a single line only says where on that line it lies. The three unknown positions along the lines are the three $k_i$ of \cref{M-sec:single}, and the spare degree of freedom is $f$.

\subsection{Independent frames}

Two boxes rotated arbitrarily share only the camera. Each contributes three lines against its own three-parameter rotation, so neither helps the other and $f$ stays under-determined by one. What the pair gives is the intersection of two admissible sets from \cref{M-sec:bound}, typically tighter than either because the bounds depend on image position and angles differently. If the two frames share one axis, the usual case of two objects on one floor, the second rotation has one free parameter and the count becomes $+1$: $f$ is determined with one equation spare. \Cref{sec:solver} gives the solver and shows that four of the six lines already determine $f$.

\subsection{Connected by an edge}

When the two corners are the ends of one edge, the shared edge appears as a single image line through both $\vc_1$ and $\vc_2$ rather than as one line per corner, and its direction is in any case fixed by $\vc_1$ and $\vc_2$. There are therefore five line observations, the shared line and the four non-shared ones, against the four unknowns $R$ and $f$, and the calibration count drops from $+2$ to $+1$: $f$ is still determined, with one check instead of two. The corner positions do not enter this count; they determine the relative depth, as described below. In vanishing-point terms, the shared direction has only one line, so its vanishing point is known only to lie on that line. Three different things are determined, and should be kept apart. \emph{Calibration}: $f$ and $R$, with one check. \emph{Relative position up to scale}: write $P_1=\lambda\vr_1$ with $\lambda$ the depth gauge and $P_2=P_1+\mu\,\vd_{12}$; the ray condition $P_2\parallel\vr_2$ reads $\vr_1\times\vr_2+(\mu/\lambda)\,\vd_{12}\times\vr_2=0$, a consistent vector equation because $\vr_1$, $\vr_2$ and $\vd_{12}$ are coplanar (all lie in the interpretation plane of the shared line), so $\mu/\lambda$ is determined and the corners are located relative to each other. \emph{Metric scale}: one known length, of the shared edge or of any edge whose far end is also seen, fixes $\lambda$. A connected pair also yields the shared direction from one long segment, in practice the most accurate of the three. This row is the smallest fragment of the parallelepiped method: the shared edge is the first piece of the box's affine structure, and with four more matched vertices the canonic projection matrix of \citet{wilczkowiak2001} becomes available.

\subsection{Corners across images}\label{sec:multiview}

A corner, or any corner aligned to the same Manhattan frame, seen in two images gives each camera's absolute rotation into that frame, so the relative rotation between cameras follows without point correspondences and without the degeneracy that essential-matrix estimation suffers under pure rotation. \citet{kanatani1986} used the relative rotation between two corner images to track an object under orthography; this is the same computation with the camera moving. Two caveats apply. The absolute rotations are defined only up to the 24 rotational symmetries of the cube unless the axes are matched across the views, for instance by gravity or by appearance. Translation needs more than one corner. Once both rotations are known, the corner point gives one epipolar equation on the two-parameter baseline direction, and the edge lines add nothing, since in each view an edge line is fixed by the corner point and by the vanishing point that the rotation already predicts. A connected pair gives two point matches, which fix the baseline direction, and one known edge length then fixes its length.


\section{What opacity adds}\label{sec:opacity}

\Cref{M-sec:single,M-sec:bound,sec:two} treat a corner as three lines, as if its faces were glass. \citet{kanatani1986} analyzed opaque rectangular polyhedra under orthography: his Proposition~7 lists the visibility conditions a corner must satisfy, his Propositions~5 and~6 tie the states of adjacent corners together along a shared edge, and his examples use both to detect hidden lines and reject interpretations. This section restates the part of that analysis that bears on one corner under perspective, then adds what he did not treat.

\paragraph{Visibility is mirror-symmetric (restated).} For a convex corner the face spanned by edges $i$ and $j$ has outward normal $-\vd_k$, so it is visible exactly when the third edge recedes, $\alpha_k>0$. At a fork all three $\alpha$ are positive and all three faces show; at an arrow the shaft is the one edge that approaches the camera, and the face opposite it, between the barbs, is hidden. For a concave corner the normals reverse and the condition is $\alpha_k<0$. The mirror flip of \cref{M-sec:mirror1} negates every $\alpha_i$, so both hypotheses predict the same visible faces and the same junction type. Opacity alone leaves the mirror ambiguity where it was; this is why the hollow-mask illusion works.

\paragraph{The camera's side resolves it.} What breaks the symmetry is knowing on which side of the faces the camera is. Assume the camera is inside the room, the three faces at the junction are two walls and the floor or ceiling, the junction has been correctly classified as a room corner, and its three visible edges are the wall--wall and wall--floor edges rather than occluding contours of nearer objects. Then the camera lies in the open octant bounded by three opaque faces, so all three faces are visible, all $\alpha_i<0$, and every room corner is a fork with the concave sign. A box seen from outside may be a fork or an arrow. Hence an arrow junction is never a wall corner seen from inside, and the prior ``this image was taken indoors'' fixes the sign of every fork that is a wall corner. What remains is the choice between a box corner and a room corner, a scene-level question rather than a geometric one.

\paragraph{A corner known to be a room corner.} When the concave sign is given, four things follow. With $K$ known the rotation from one corner is unique: the closed form of \cref{M-sec:closed} is taken with all $k_i$ negative. The junction must be a fork, which tests the detection. With $f$ unknown, \cref{prop:concave} applies: the admissible set keeps only its fork-pattern components, and the known sign then selects the concave solution on them. The three faces are the floor or ceiling and two walls, so identifying the floor region identifies the vertical edge, which fixes which axis of the recovered frame is up, and a known camera height then makes the floor plane metric by single-view metrology \citep{criminisi2000}. Room corners are also almost never alone: the floor--wall edge runs to the next corner, so the connected-pair row of \cref{tab:counts} is the normal case indoors. A wide-angle view showing the four corners of an end wall and two of an adjacent wall reaches the six vertices at which the parallelepiped method applies to the room itself, seen from inside.

\paragraph{Edge labels (restated, then combined).} With opaque faces every image edge is convex, concave or occluding \citep{huffman1971,clowes1971,waltz1975}. Under the convex reading of an arrow the barbs are occluding edges and the surface beyond each belongs to something farther away; under the concave reading they are convex edges between faces of one object. Occluding edges have observable signatures: T-junctions where other edges vanish behind them, unrelated texture on the two sides, parallax under motion. For a connected pair the shared edge carries one label at both ends, so the two corners' signs are linked \citep[Proposition~6]{kanatani1986}, and this link holds under orthography, where \cref{sec:mirror} fails. The two mechanisms are complementary: the label link propagates one known sign across a polyhedron, and the perspective argument of \cref{sec:mirror} supplies that sign.

\paragraph{Sidedness in several views.} Each visible face places the camera center on its outer side. For the face spanned by $\vd_i,\vd_j$ of a convex corner at $P$ the condition is $(O-P)\cdot\vd_k<0$, with the inequality reversed for a concave corner; in the corner frame, where $P$ is the origin, each is a sign condition on one coordinate of the camera center, one per visible face. In the two-view setting of \cref{sec:multiview} these inequalities, one set per view, bound each camera center to an orthant of the corner frame and hence the baseline direction to a cone, complementing the single epipolar equation that one corner point provides.


\subsection{A corner whose edges share one sign}\label{sec:concave}

Suppose the corner is known to be a room corner seen from inside, so that all three edges approach the camera and all three $k_i$ are negative. Then every product $k_ik_j$ is positive, and by \eqref{M-eq:kk} each $g_{ij}(f)$ must be negative on its own: the junction must be a fork in the canonical view. The same holds for a box corner known to have all three edges receding, since then all $k_i$ are positive and the products are again positive. The prior is therefore a prior on the fork pattern. It cannot tell the room corner from the box corner, because that choice is the mirror of \cref{M-sec:mirror1}.

\begin{proposition}[Uniform-sign prior]\label{prop:concave}
If the three edges of a corner are known to have the same sign, its admissible set is $F^{c}=\{f>0:\,A_{ij}f^2+e_ie_j<0\text{ for all three pairs}\}$, an intersection of three sets each of which is an open half-line, all of $(0,\infty)$, or empty, and hence an interval. By \cref{M-thm:interval} and \cref{M-prop:bound}(3), $F^c=F$ whenever the sign pattern on $F$ is $(-,-,-)$, and $F^c=\emptyset$ otherwise.
\end{proposition}
\begin{proof}
$F^c\subseteq F$ since three negatives have a negative product. Each $\{x>0:A_{ij}x+e_ie_j<0\}$ is an interval: a half-line when $A_{ij}\neq0$, and either $(0,\infty)$ or $\emptyset$ when $A_{ij}=0$, according to the sign of $e_ie_j$. An intersection of intervals is an interval, so $F^c$ is one. By \cref{M-prop:bound}(3) the pattern is constant on $F$, which is one interval by \cref{M-thm:interval}, so $F$ is either contained in $F^c$ or disjoint from it.
\end{proof}

The prior therefore never tightens the bound; in \cref{M-sec:e2} the two sets coincided for every corner whose edges all approach the camera. Its value is as a test: it rejects a junction whose admissible set carries the arrow pattern, which under the stated prior is either a false detection or a sign that $K$ is wrong.

In the experiments of \cref{M-sec:e2} of the article, for corners whose three edges all approach the camera in truth, the set $F^c$ of \cref{prop:concave} coincided with $F$ in every case.

\section{Further results on several frames}\label{sec:Sframes}

Shared axes are common outdoors: buildings standing at different angles on level ground have different horizontal axes but the same vertical. This is the Atlanta world of \citet{schindler2004}. However many frames such a scene contains, it leaves one camera parameter undetermined, an ambiguity that amounts to a stretch of the scene along the vertical. In classical terms, the frames contribute the pole--polar relation between the vertical vanishing point and the horizon, and the conjugacy of two pairs of points on the horizon. These amount to four conditions on five parameters if each frame shows two of its edge-on lines; for a nearly level camera only the horizon passes through the image, and only the two pole--polar conditions are forced. Further pairs on the same horizon add nothing. These are the pole--polar and circular-point constraints of \citet{liebowitz1999}.

\subsection{A camera that only pitches}

When the camera only pitches, the admitted family is an explicit interval, and the description is global.

\begin{remark}[A camera that only pitches]\label{cor:pitch}
Let the images of \cref{M-cor:horizons} show the other edge-on lines missing the image, let the camera only pitch between them, and let two of their vertical directions be distinct and not perpendicular. Then the cameras that admit all their junctions are exactly $K(I+t\,\mathbf e_1\mathbf e_1^{\top})K^{\top}$ with $t$ in an interval $(t_-,t_+)\ni0$, up to its ends, which is the intersection over the images of the conditions \eqref{M-eq:oneline} with $N=I+t\,\mathbf a\mathbf a^{\top}$, $\mathbf a=R_m^{\top}\mathbf e_1$ for the frame $R_m$ of image $m$.
\end{remark}
\begin{proof}
Each vertical direction is an eigenvector of $\Xi$ by \cref{M-thm:frames}(1), and they all lie in the plane perpendicular to $\mathbf e_1$. Two of them, distinct and not perpendicular, make $\Xi$ a multiple of the identity on that plane, so that $\Xi=\alpha I+\beta\,\mathbf e_1\mathbf e_1^{\top}$, which up to scale is $I+t\,\mathbf e_1\mathbf e_1^{\top}$ with $t>-1$. For image $m$, $N=R_m^{\top}\Xi R_m=I+t\,\mathbf a\mathbf a^{\top}$, whose entries are affine in $t$ and which has $a_k=0$ for the vertical axis $k$, since $\mathbf e_1$ is perpendicular to that axis. \Cref{M-prop:oneline} turns each image's junctions into five conditions affine in $t$, and the intersection of the resulting half-lines is an interval that contains $0$, the true camera.
\end{proof}

For the three images of Sect.~S11, at pitches of $4^\circ$, $11^\circ$ and $-7^\circ$, the formula gives the interval $(-0.166,\,0.227)$; a dense check of the junctions confirms it to within its grid step of $0.01$, giving $[-0.16,\,0.23]$.

\subsection{Two images that show only their horizons}

\begin{remark}[Two images that show only their horizons]\label{cor:twohorizons}
Two images taken with a camera with square pixels and zero skew, each showing only its horizon, passing through the image, determine its focal length and principal point if and only if their vertical directions are distinct, not perpendicular, and not both parallel to the image plane.
\end{remark}
\begin{proof}
The visible axes are the two vertical directions $\mathbf g_1$, $\mathbf g_2$, and no axis is touching. If neither is tilted, this is case (1) of \cref{M-thm:fpp}. If only $\mathbf g_1$ is, $\mathbf g_2$ has $z=0$, so $\bar{\mathbf g}_1\cdot\bar{\mathbf g}_2=\mathbf g_1\cdot\mathbf g_2$, and $\mathbf g_2$ is normal to the plane of $\mathbf g_1$ and $\hat{\mathbf z}$ exactly when the two are perpendicular, which is case (2). If both are tilted and distinct, either their image directions are not parallel and the camera is determined, or both lie in one plane with $\hat{\mathbf z}$, which is case (2) exactly when they are perpendicular.
\end{proof}

This holds even if the camera only pitches, a motion that leaves the aspect ratio of a general camera undetermined.

\section{Model error: corners that are not right-angled, and lens distortion}\label{sec:robust}

The results so far assume exact right angles and a pinhole camera. The next two propositions describe what happens when either assumption fails.

\begin{proposition}[Corners that are not right-angled]\label{prop:nonright}
Let a corner have unit edge directions $\vd_i$, not necessarily orthogonal, seen by the true camera along the viewing ray $\vr$, with $\hat\vr=\vr/\|\vr\|$, and let $c_{ij}=\vd_i\cdot\vd_j$ and $a_i=\vd_i\cdot\hat\vr$. The true camera admits the junction as the image of a right-angle corner if and only if
\begin{equation}
\prod_{(i,j)}\bigl(c_{ij}-a_ia_j\bigr)<0 .
\label{eq:nonright}
\end{equation}
In particular it does whenever $|c_{ij}|<|a_ia_j|$ for every pair, and then, when only the focal length is unknown, the true $f$ lies in $F$.
\end{proposition}
\begin{proof}
The normal $\vn_i$ of the interpretation plane of edge $i$ is perpendicular to $\vd_i$ and to $\vr$, so it is a nonzero multiple of $\vd_i\times\vr$, and $\vt_i=\vn_i\times\vr$ is a nonzero multiple of $(\vd_i\times\vr)\times\vr=-\|\vr\|^2\bigl(\vd_i-a_i\hat\vr\bigr)$. Hence $g_{ij}=\vt_i\cdot\vt_j=m_im_j\,(c_{ij}-a_ia_j)$ with nonzero $m_i$, and since each $m_i$ appears twice in $g_{12}g_{23}g_{31}$, the product has the sign of \eqref{eq:nonright}; \cref{M-lem:sign} gives the equivalence. If $|c_{ij}|<|a_ia_j|$ for every pair, each factor has the sign of $-a_ia_j$, and the product has the sign of $-(a_1a_2a_3)^2<0$.
\end{proof}

For a right-angle corner, $c_{ij}=0$ and \eqref{eq:nonright} reads $-(a_1a_2a_3)^2<0$, which holds unless an edge is perpendicular to the ray. For a corner whose angles are slightly off, the sufficient condition fails only where an edge is nearly perpendicular to the ray, that is, near an edge-on line, which is precisely where the informative junctions lie. In simulation at $60^\circ$, without noise and with each edge tilted by up to $2^\circ$, $5^\circ$ and $10^\circ$ away from an orthogonal frame, the true $f$ remained in $F$ for $92.3\%$, $84.2\%$ and $74.6\%$ of junctions. A few degrees of model error thus cost more coverage than a pixel of noise (\cref{M-sec:e2}).

\begin{proposition}[Radial distortion]\label{prop:distortion}
Let the image be distorted about the principal point by $\boldsymbol\chi\mapsto\boldsymbol\chi(1+k\|\boldsymbol\chi\|^2)$, with $1+k\rho^2>0$ and $1+3k\rho^2>0$, and let each edge be represented by its tangent at the distorted vertex. A junction at distance $\rho$ from the principal point with angles $\psi_i$ is then observed at distance $\rho'=\rho(1+k\rho^2)$ with angles $\psi'_i$ in the same quadrants as the $\psi_i$ and
\[
\tan\psi'_i=\vartheta\tan\psi_i,\qquad \vartheta=\frac{1+k\rho^2}{1+3k\rho^2}.
\]
Writing $F(\rho,\psi)$ for the admissible set of a junction with these data, the observed junction has the admissible set $F(\rho',\psi')=(1+k\rho^2)\,F(\rho,\psi')$.
\end{proposition}
\begin{proof}
The differential of the distortion at $\boldsymbol\chi$ is $(1+k\|\boldsymbol\chi\|^2)I+2k\,\boldsymbol\chi\boldsymbol\chi^{\top}$, which multiplies the direction away from the principal point by $1+3k\rho^2$ and the perpendicular direction by $1+k\rho^2$. A half-edge whose components along and across that direction are a positive multiple of $(\cos\psi_i,\sin\psi_i)$ is therefore mapped to a positive multiple of $\bigl((1+3k\rho^2)\cos\psi_i,(1+k\rho^2)\sin\psi_i\bigr)$, which gives $\psi'_i$, and the vertex moves along the line through the principal point, to distance $\rho'$. The last equality is \cref{M-cor:similar}.
\end{proof}

Distortion thus acts on each junction by a scaling about the principal point together with a shear of its angles toward or away from the direction of the principal point, and $F$ moves accordingly. Fitting a segment by a chord rather than a tangent adds an error that vanishes as the segments shorten. In simulation at $60^\circ$, without noise and with tangent edges, distortion that displaces the image corners by $1\%$, $2\%$ and $5\%$ left the true $f$ in $F$ for $97.9\%$, $96.1\%$ and $92.2\%$ of junctions. The York Urban segments were used as given (\cref{sec:proposals}).

With \cref{M-prop:certified}, distortion yields a guarantee rather than a coverage figure.

\begin{corollary}[Distortion with fitted segments]\label{cor:chord}
Under the hypotheses of \cref{prop:distortion}, let each edge be measured by the chord of its distorted image over a segment of undistorted length $\ell_{\mathrm{seg}}$ starting at the vertex, and let $\underline J=\min\bigl(|1+k\rho^2|,|1+3k\rho^2|\bigr)$, the smaller singular value of the differential of the distortion at the vertex. If $3|k|(\rho+\ell_{\mathrm{seg}})\ell_{\mathrm{seg}}<\underline J$, each measured angle differs from the angle $\psi_i$ of the undistorted junction by at most
\[
\varphi=\arctan\frac{|1-\vartheta|}{2\sqrt\vartheta}+\arcsin\frac{3|k|(\rho+\ell_{\mathrm{seg}})\ell_{\mathrm{seg}}}{\underline J},
\]
and the certified set of \cref{M-prop:certified}, computed from the observed junction, contains $(1+k\rho^2)f$: the true focal length, scaled by the factor by which distortion moves the vertex away from the principal point.
\end{corollary}
\begin{proof}
By \cref{prop:distortion}, the tangent angles satisfy $\tan\psi'_i=\vartheta\tan\psi_i$, and over all $\psi$ the difference $|\arctan(\vartheta t)-\arctan t|$, with $t=\tan\psi$, is largest at $t=1/\sqrt\vartheta$, where it equals the first term. For the chord, write $D(\boldsymbol\chi)=\boldsymbol\chi(1+k\|\boldsymbol\chi\|^2)$ and let $\mathbf u$ be the unit direction of the undistorted edge from the vertex $\mathbf c$. The second derivative of $D$ along $\mathbf u$ is $k\bigl(2\boldsymbol\chi+4(\boldsymbol\chi\cdot\mathbf u)\mathbf u\bigr)$, of norm at most $6|k|\|\boldsymbol\chi\|\le6|k|(\rho+\ell_{\mathrm{seg}})$ on the segment, so the chord $D(\mathbf c+\ell_{\mathrm{seg}}\mathbf u)-D(\mathbf c)$ differs from $\ell_{\mathrm{seg}}\,DD(\mathbf c)\mathbf u$ by at most $3|k|(\rho+\ell_{\mathrm{seg}})\ell_{\mathrm{seg}}^2$, while $\|DD(\mathbf c)\mathbf u\|\ge\underline J$. The angle between the chord and the tangent is therefore at most the second term. The observed junction has radius $\rho'=\rho(1+k\rho^2)$ and angles within $\varphi$ of the $\psi_i$; by \cref{M-cor:similar}, rescaling the radius from $\rho$ to $\rho'$ multiplies both $\hat F$ and $\mathcal U$ by $(1+k\rho^2)^2$ as sets of $x$, and \cref{M-prop:certified} at radius $\rho$ places $f^2$ in the unscaled set.
\end{proof}


\section{First-order noise analysis}\label{sec:Snoise}

Real junctions come from noisy line fits, so their thresholds move. Two questions follow: how far, and how often the true focal length falls outside the measured admissible set.

\begin{proposition}[First-order spread of a threshold]\label{prop:noise}
Let each edge be observed as a segment of length $\ell_{\mathrm{seg}}$ starting at the junction, both of whose endpoints are perturbed independently by $\mathcal N(0,\sigma^2I)$, let each line be fitted through its two perturbed endpoints, and, as an idealization, let the junction itself be held at its true position, so that $\rho$ and the reference direction $(u,v)/\rho$ are exact and only the three directions $\psi_i$ are perturbed. Write $\hat x_{ij}$ and $\hat F$ for the threshold and the admissible set computed from the perturbed directions. If $\cos(\psi_i-\psi_j)\neq0$ and $\sin2\psi_i$, $\sin2\psi_j$ are not both zero, then as $\sigma\to0$
\begin{equation}
\begin{aligned}
&\frac{\hat x_{ij}-x_{ij}}{\operatorname{sd}(x_{ij})}\longrightarrow\mathcal N(0,1)\ \ \text{in distribution},\\
&\operatorname{sd}(x_{ij})=\frac{\sigma\rho^2\sqrt{\sin^2 2\psi_i+\sin^2 2\psi_j}}{\sqrt2\,\ell_{\mathrm{seg}}\,\cos^2(\psi_i-\psi_j)} .
\end{aligned}
\label{eq:sd}
\end{equation}
Moreover, for every $\sigma$, the true focal length is excluded only if one of the three forms changes sign at $f^2$:
\begin{equation}
\Pr\bigl(f\notin\hat F\bigr)\le\sum_{(i,j)}\Pr\Bigl(\bigl(\hat A_{ij}f^2+\hat e_i\hat e_j\bigr)\bigl(A_{ij}f^2+e_ie_j\bigr)\le0\Bigr).
\label{eq:union}
\end{equation}
If instead the two angles of a pair are perturbed by independent $\mathcal N(0,\tau^2)$ errors, then for every $\tau$
\begin{equation}
\begin{aligned}
&\Pr\bigl(\text{form $(i,j)$ changes sign at $f^2$}\bigr)\\
&\qquad\le\exp\!\left(-\frac{\bigl(A_{ij}f^2+e_ie_j\bigr)^2}{2\tau^2\bigl(\sqrt2\,f^2+\rho^2\bigr)^2}\right),
\end{aligned}
\label{eq:rigorous}
\end{equation}
and to first order the endpoint model is this one with $\tau=\sqrt2\,\sigma/\ell_{\mathrm{seg}}$.
\end{proposition}
\begin{proof}
The direction of a segment of length $\ell_{\mathrm{seg}}$ whose endpoints are displaced by $\sigma\mathbf a$ and $\sigma\mathbf b$ turns through $\sigma(\mathbf b-\mathbf a)\cdot\mathbf e_{\perp}/\ell_{\mathrm{seg}}+O(\sigma^2)$, where $\mathbf e_{\perp}$ is the unit normal to the segment; so $\hat\psi_i=\psi_i+\sigma\xi_i/\ell_{\mathrm{seg}}+O(\sigma^2)$ almost surely, with $\xi_1,\xi_2,\xi_3$ independent $\mathcal N(0,2)$. By \eqref{M-eq:thresh} the threshold is a differentiable function of $(\psi_i,\psi_j)$ where $\cos(\psi_i-\psi_j)\neq0$, with $\partial x_{ij}/\partial\psi_i=-\rho^2\sin2\psi_j/(2\cos^2(\psi_i-\psi_j))$ and symmetrically in $j$, and \eqref{eq:sd} is the delta method, the gradient being nonzero by hypothesis. For \eqref{eq:union}, $f\in\hat F$ exactly when the product $\hat L(f^2)$ of the three perturbed forms is negative; if none of the three changes sign at $f^2$, that product has the sign of $L(f^2)$, which is negative because $f\in F$.
 For \eqref{eq:rigorous}, write the form as $\Phi_{ij}(\psi_i,\psi_j)=\cos(\psi_i-\psi_j)f^2+\rho^2\sin\psi_i\sin\psi_j$. Its gradient is the sum of $f^2\sin(\psi_i-\psi_j)(-1,1)$, of norm at most $\sqrt2 f^2$, and $\rho^2(\cos\psi_i\sin\psi_j,\sin\psi_i\cos\psi_j)$, whose squared norm is $\rho^4\bigl(\sin^2(\psi_i+\psi_j)+\sin^2(\psi_i-\psi_j)\bigr)/2\le\rho^4$; so $\Phi_{ij}$ is Lipschitz with constant $L=\sqrt2 f^2+\rho^2$. A sign change needs $|\Phi_{ij}(\psi+\xi)-\Phi_{ij}(\psi)|\ge|\Phi_{ij}(\psi)|$, hence $\|\xi\|\ge|\Phi_{ij}(\psi)|/L$, and the norm of $\xi\sim\mathcal N(0,\tau^2I_2)$ has the Rayleigh tail $\Pr(\|\xi\|\ge t)=e^{-t^2/(2\tau^2)}$.\end{proof}

Because $\hat x_{ij}$ has heavy tails and no finite moments, $\operatorname{sd}(x_{ij})$ is a first-order scale rather than a standard deviation, and it describes the threshold, not the decision. The distinction matters near a raw right angle: there the threshold and its scale both diverge, while the form $A_{ij}f^2+e_ie_j$ stays near $e_ie_j$, far from zero, so the decision whether $f$ is admitted remains stable even where the threshold is not. Linearizing each form itself in the three angles therefore gives the working formula
\begin{equation}
\begin{aligned}
&\Pr\bigl(f\notin\hat F\bigr)\approx\sum_{(i,j)}\Phi\!\left(-\frac{|A_{ij}f^2+e_ie_j|}{s_{ij}}\right),\\
&s_{ij}^2=\frac{2\sigma^2}{\ell_{\mathrm{seg}}^2}\bigl(\partial_i^2+\partial_j^2\bigr),
\end{aligned}
\label{eq:miss}
\end{equation}
where $\Phi$ is the standard normal distribution function and $\partial_i=-\sin(\psi_i-\psi_j)f^2+\rho^2\cos\psi_i\sin\psi_j$, $\partial_j=\sin(\psi_i-\psi_j)f^2+\rho^2\sin\psi_i\cos\psi_j$ are the derivatives of the form with respect to the two angles. This is an approximation, not a bound. Summed over the three pairs, \eqref{eq:rigorous} gives a bound valid at every noise level; in simulation it exceeded the true probability by a median factor of $5.5$ where that probability was above $1\%$. The Lipschitz constant $\sqrt2 f^2+\rho^2$ is attained for some angles, so the slack stems from replacing the set where the sign changes by a disc, not from the constant. Holding the junction fixed is an idealization; Sect.~S10 measures the effect of each.

Three conclusions follow from \eqref{eq:sd}. First, the spread grows as $\rho^2$: the border junctions that give the tightest bounds (\cref{M-cor:similar}) are also the least certain, while the relative scale $\operatorname{sd}(x_{ij})/x_{ij}$ does not depend on $\rho$ at all. Second, the spread grows as $1/\cos^2(\psi_i-\psi_j)$, so the threshold of a pair close to a raw right angle in the image is unstable, although, by \eqref{eq:miss}, the pair's contribution to the decision is not. Third, the spread falls as $1/\ell_{\mathrm{seg}}$, so edge length buys accuracy linearly. Sect.~S10 checks \eqref{eq:sd} against a numerical gradient and \eqref{eq:miss} against Monte Carlo. These two formulas are what any use of the bound requires: \eqref{eq:miss} indicates which junctions to trust, and \eqref{eq:sd} supplies a per-junction weight, which the log-plausibility of Sect.~S8 uses and its consensus rule does not.

\section{Combining bounds under noise}\label{sec:Scombine}

This section collects the rules that combine the bounds of many junctions under noise. They are heuristics, not results of the article, and the article keeps only their conclusion.

\subsection{The consensus and the log-plausibility}\label{sec:prior}

Every junction that passes a loose classifier yields an admissible set from \cref{M-sec:bound}, including junctions with no visible parallel partner and junctions whose edges are too short to locate a vanishing point. Combining them over an image or a video gives a bound on $f$ with no line families and no calibration object, and a natural prior for a later bundle adjustment of $f$.

Combining them by \emph{intersection}, however, does not survive noise, and the failure gets worse as evidence accumulates. By \cref{M-thm:interval} each corner yields one interval, so the intersection is again an interval and is trivial to compute. But it contains the true $f$ only if every one of the intervals does, and \cref{M-sec:e2} puts single-corner coverage at 97 to 98\% rather than 100\%. In \cref{sec:e6} the coverage of the intersection falls from $95\%$ at $2$ junctions to $25\%$ at $50$. A depth-$\kappa$ consensus, which keeps the set covered by at least $\kappa$ of the $M$ intervals and costs a sort of the endpoints, keeps the coverage at 96 to 100\% from $10$ junctions on at $\sigma=0.5$\,px, and 88 to 97\% at $\sigma=1$\,px, with $\kappa=\lceil0.9M\rceil$, as a binomial model of independent misses predicts (\cref{tab:combine}). 

The consensus rule is a hard cut and a heuristic, however, and \cref{prop:noise} supplies what is needed to replace it. For a candidate $x=f^2$ and a junction with thresholds $x_{ij}$ and spreads $\operatorname{sd}(x_{ij})$, the probability that the form of pair $(i,j)$ carries the wrong sign at $x$ is $\epsilon_{ij}=\Phi(-|x-x_{ij}|/\operatorname{sd}(x_{ij}))$, where $\Phi$ is the standard normal distribution function. Admissibility is a \emph{parity} condition, an odd number of negative forms, so the observed verdict is correct exactly when an even number of signs have flipped. If the three flips are treated as independent (an approximation, since the forms share the angles $\psi_i$), then $\Pr(\text{even})=\tfrac12(1+\prod_{(i,j)}(1-2\epsilon_{ij}))$. Writing $\varsigma=\operatorname{sgn}\prod_{(i,j)}(A_{ij}x+e_ie_j)$ for the sign of the observed product, the probability that the junction really is admissible at $x$ is
\begin{equation}
\pi(x)=\tfrac12\Bigl(1-\varsigma\textstyle\prod_{(i,j)}\bigl(1-2\epsilon_{ij}(x)\bigr)\Bigr),
\label{eq:pi}
\end{equation}
and $M$ junctions give the log-plausibility $\Lambda(x)=\sum_m\log\pi_m(x)$, summed over the junctions $m$; it is a plug-in score, not a likelihood of the data. Its maximizer is a point estimate rather than a set, and a finite maximizer exists only if some junction is bounded above: otherwise $\Lambda$ increases toward its supremum as $x\to\infty$. In the simulations the maximizer is sought on a grid from $(0.3f)^2$ to $(3f)^2$. Here $\epsilon_{ij}$ uses the threshold scale \eqref{eq:sd}. On the same junction samples, $100$ trials per setting, the form-based scale of \eqref{eq:miss} leaves the maximizer's median error unchanged at $\sigma=0.5$\,px ($2.8\%$ and $2.7\%$ at $50$ junctions) and reduces it markedly at $\sigma=2$\,px, from $18.2\%$ to $11.9\%$ at $10$ junctions and from $8.0\%$ to $4.2\%$ at $50$; the form-based scale is the one to use, and \cref{tab:lowcount} uses it. \Cref{tab:lowcount} reports its accuracy. The parity matters: it is what ties $\Lambda$ to the sign condition, since the margins $|x-x_{ij}|$ alone do not say on which side of each threshold the admissible set lies.


\paragraph{Consensus sets.} The depth-$\kappa$ consensus need not be a single interval.

\begin{lemma}\label{lem:Sdepth}
Let $M$ intervals be given and $\kappa>M/2$. The set covered by at least $\kappa$ of them is a finite union of intervals, which need not be connected, and its hull is contained in the set covered by at least $2\kappa-M$ of them.
\end{lemma}
\begin{proof}
Let $x<y<z$, with $x$ and $z$ each covered by at least $\kappa$ intervals. At most $M-\kappa$ intervals miss $x$ and at most $M-\kappa$ miss $z$, so at least $2\kappa-M$ contain both, and each of these contains $y$. For a disconnected example, take $M\ge3$, $\kappa=M-1$, one interval $[0,1]$, one interval $[2,3]$ and $M-2$ copies of $[0,3]$: the points $1/2$ and $5/2$ are covered $M-1$ times and $3/2$ only $M-2$ times.
\end{proof}

In the simulations below, the coverage reported is that of the set itself, and its lower and upper ends are those of its hull, which by the lemma lies at depth at least $2\kappa-M$, about $0.8M$. The set was never disconnected at $M=2$ and $5$; at $M=10$, $25$ and $50$ it was disconnected in $5\%$, $11\%$ and $4\%$ of trials at $\sigma=0.5$\,px, and in $6\%$, $19\%$ and $19\%$ at $\sigma=1$\,px.

\paragraph{Independence of the flips.} The three flips are not independent: the forms of the pairs $(1,2)$ and $(2,3)$ share the error of edge $2$, with covariance $\tau^2\,\partial_{\psi_2}\Phi_{12}\,\partial_{\psi_2}\Phi_{23}$ to first order. By inclusion and exclusion, the probability of an odd number of flips is $\sum p_i-2\sum_{i<j}p_{ij}+4p_{123}$, with $p_{ij}$ and $p_{123}$ the joint flip probabilities, and its independent version replaces each joint probability by the product of the marginals. Each joint probability and each product lies between $0$ and the smaller of the marginals involved, so the two differ by at most $8p_{(1)}+2p_{(2)}$, where $p_{(1)}\le p_{(2)}$ are the two smallest flip probabilities: independence matters only where two forms are near zero at once. Over $600$ random junctions at $\sigma=1$\,px the difference from the exact correlated value was at most $0.0018$.

\subsection{Simulations}\label{sec:e6}
Corners of one scene are drawn at $60^\circ$ and their admissible sets combined, 400 trials per setting. The strict intersection is compared with the depth-$\kappa$ consensus of \cref{sec:prior} at $\kappa=\lceil 0.9M\rceil$ (\cref{tab:combine}).

\begin{table*}[t]
\centering\small
\caption{Combining the bounds of $M$ junctions seen by one camera at $60^\circ$, each from its own random frame, $400$ trials per setting. The consensus is the set covered by at least $\kappa=\lceil0.9M\rceil$ of the $M$ intervals; at $M=2$ and $5$, $\kappa=M$ and it is the intersection. The binomial prediction assumes independent misses at the single-junction rate. Lower and upper ends are medians over all trials}
\label{tab:combine}
\begin{tabular}{lccccc}
\toprule
junctions $M$ & 2 & 5 & 10 & 25 & 50\\
\midrule
intersection, coverage of true $f$, $\sigma=0.5$\,px & 95\% & 89\% & 75\% & 50\% & 25\%\\
consensus, coverage, $\sigma=0.5$\,px & 95\% & 89\% & 96\% & 98\% & 100\%\\
binomial prediction for the consensus, $\sigma=0.5$\,px & 95\% & 88\% & 98\% & 98\% & 100\%\\
consensus, lower end, $/f$, $\sigma=0.5$\,px & 0.57 & 0.76 & 0.71 & 0.81 & 0.80\\
consensus bounded above, $\sigma=0.5$\,px & 34\% & 62\% & 60\% & 88\% & 96\%\\
consensus, upper end, $/f$, $\sigma=0.5$\,px & $\infty$ & 1.72 & 2.26 & 1.41 & 1.40\\
intersection, coverage, $\sigma=1$\,px & 93\% & 76\% & 57\% & 30\% & 8\%\\
consensus, coverage, $\sigma=1$\,px & 93\% & 76\% & 94\% & 88\% & 97\%\\
binomial prediction for the consensus, $\sigma=1$\,px & 92\% & 80\% & 93\% & 91\% & 98\%\\
\bottomrule
\end{tabular}
\end{table*}

The intersection degrades monotonically in the number of corners, as the argument in \cref{sec:prior} predicts, while the coverage of the consensus stays at 96 to 100\% from $10$ junctions on at $\sigma=0.5$\,px. At small $M$ the two coincide because $\lceil 0.9M\rceil=M$.

 The log-plausibility \eqref{eq:pi} gives a point estimator on data of the same kind. \Cref{tab:lowcount} gives the error of its maximizer at the numbers of junctions a real image supplies ($1.8$ full junctions per image in \cref{sec:e4}) and at larger numbers. The consensus interval is no substitute: it is a set, often unbounded above, and the midpoint of such an interval is set largely by where it is clipped.

\begin{table*}[t]
\centering\footnotesize
\caption{The bound and the log-plausibility estimator at the numbers of junctions a real image supplies, at a $60^\circ$ field of view, $200$ trials per setting. Errors are median relative errors in $f$ of the maximizer of $\Lambda$ with the form-based scale of \eqref{eq:miss}, with bootstrap $95\%$ intervals; the maximizer is sought on a grid from $0.3f$ to $3f$, which dominates the error at $M\le3$}
\label{tab:lowcount}
\begin{tabular}{lccccccc}
\toprule
junctions $M$ & 1 & 2 & 3 & 5 & 10 & 20 & 50\\
\midrule
intersection bounded, nonempty, $\sigma=0.5$\,px & 20\% & 29\% & 49\% & 64\% & 82\% & 78\% & 44\%\\
intersection empty, $\sigma=0.5$\,px & 0\% & 1\% & 1\% & 0\% & 5\% & 21\% & 56\%\\
its median width, $/f$ & 0.69 & 0.79 & 0.60 & 0.49 & 0.39 & 0.19 & 0.06\\
error of the maximizer of $\Lambda$, $\sigma=0.5$\,px & 68\% & 37\% & 33\% & 16\% & 8.5\% & 5.2\% & 2.6\%\\
\quad 95\% interval & 64--70 & 33--42 & 24--41 & 12--22 & 7.5--11 & 4.3--6.1 & 2.0--2.9\\
error of the maximizer of $\Lambda$, $\sigma=2$\,px & 68\% & 52\% & 34\% & 23\% & 14\% & 7.4\% & 5.1\%\\
\quad 95\% interval & 61--70 & 40--64 & 30--41 & 18--26 & 11--16 & 6.5--8.6 & 4.2--5.7\\
\bottomrule
\end{tabular}
\end{table*}

\Cref{tab:lowcount} shows the sharpest practical limitation of the method: one junction is nearly useless as an estimator of $f$, two or three are weak, and the maximizer's error falls below 10\% only at about ten to twenty. At $1.8$ junctions per image that means roughly six to eleven images taken with the same camera, not one; they need not show the same scene. The bound is therefore a constraint to carry into a larger estimation problem, or an accumulator over a video, rather than a single-image calibrator. The identifiability of \cref{M-thm:identframe} concerns junctions spread over the whole image and should not be read as a claim about one image.

\section{Junctions placed at random}\label{sec:Srandomall}

\subsection{Random edge directions}\label{sec:Srandom}

The interval of \cref{M-thm:interval} can be empty, a half-line or bounded. For junctions whose edges point in random directions, the three cases have simple probabilities.

\begin{proposition}[Random junctions]\label{prop:random}
If the three half-edge directions of a junction with $\rho>0$ are independent and uniformly distributed, then $F$ is empty with probability $1/2$, a half-line $(f_{ij},\infty)$ with probability $1/4$, and a bounded interval with probability $1/4$.
\end{proposition}
\begin{proof}
The angles $\psi_i$ are then independent and uniform, so the $\tau_i=\cot\psi_i$ are independent standard Cauchy variables. Their signs are independent fair coins, independent of the magnitudes $|\tau_i|$, and $|\tau_i|$ has the same distribution as $1/|\tau_i|$, since $\psi\mapsto\pi/2-\psi$ exchanges $\cot$ and $\tan$. A pair is in $S$ exactly when $\tau_i\tau_j<-1$. If the three signs agree, no product is negative and $S$ is empty; this has probability $1/4$. Otherwise exactly one $\tau_k$ has the minority sign, with probability $3/4$, and only the two pairs containing $k$ can be in $S$: $(i,k)\in S$ exactly when $|\tau_i|>W$, with $W=1/|\tau_k|$. The variables $|\tau_i|$, $|\tau_j|$ and $W$ are independent and identically distributed, so $W$ is the largest, the middle or the smallest of the three with probability $1/3$ each, and $|S|$ is $0$, $1$ or $2$ accordingly. Hence $|S|=0$ with probability $1/4+\tfrac34\cdot\tfrac13=1/2$, and $|S|=1$ and $|S|=2$ each with probability $\tfrac34\cdot\tfrac13=1/4$; \cref{M-thm:interval} turns $|S|$ into the shape of $F$.
\end{proof}

The camera-free test therefore rejects exactly half of all junctions whose edges point in independent uniform directions. Junctions of real corners are not of this kind, and the test accepts nearly all of them (\cref{sec:e4}).

\subsection{Rates for random vertex positions}\label{sec:Srate}

This section collects the rates at which junctions placed at random over the image close the bounds of \cref{M-thm:identframe}. They rest on a model of where junctions lie, independent and spread over the whole image, that real corners do not follow: on York Urban they avoid the neighborhoods of the edge-on lines (\cref{sec:e4}). The article therefore states only the deterministic results, and the rates are given here.

\begin{proposition}[Rate of identifiability]\label{prop:rate}
Under the hypotheses of \cref{M-thm:identframe}, let the image contain no vanishing point, except in part (3), and let $M$ junctions of the frame have vertices drawn independently with a density bounded above and below by positive constants on the image. Let $W_M$ be the width of $\bigcap F$, in units of $f$.
\begin{enumerate}
\item If the edge-on line of a tilted axis meets the image in a segment, whether by passing through its interior or by containing one of its edges, $W_M$ is of exact order $1/M$: for every $\varepsilon>0$ there are constants $0<c<C$ with $\Pr(c/M\le W_M\le C/M)\ge1-\varepsilon$ for all large $M$; we then say that $W_M$ is of \emph{exact order} $1/M$. Moreover $W_M$ is nonincreasing in $M$, and for an infinite sequence of junctions, almost surely $c'/(M\log^2M)\le W_M\le C'\log M/M$ for all large $M$; in particular $W_M\to0$ almost surely.
\item If such lines meet the image only at its corners, at least one of which is not the point $\vc^{\star}_k$ of its line, $W_M$ is of exact order $1/\sqrt M$ in the same sense.
\item If a vanishing point $\vv_a$ lies in the interior of the image and the other two axes are tilted, $W_M$ is of exact order $1/(M\log M)$ in the same sense.
\item (Limit law.) Under the hypotheses of (1), let the vertices have a continuous density $p$ on $\mathcal I$, and let no edge-on line contain an edge of $\mathcal I$. For each axis $k$ whose edge-on line passes through $\mathcal I$, parameterize $\ell_k$ by the signed distance $\varsigma$ from $\vc^{\star}_k$, in pixels, and let $r_k=|\zeta_i/\zeta_j|$ along it. With
\[
\chi=\sum_k|z_k|\int_{\ell_k\cap \mathcal I}p\,|\varsigma|\Bigl(r_k+\frac1{r_k}\Bigr)d\varsigma ,
\]
$y\Pr(\pm\gamma>y)\to\chi$ for both signs, and $M(f_{\max}/f-1)$ and $M(1-f_{\min}/f)$, where $f_{\min}$ and $f_{\max}$ are the ends of $\bigcap F$, each converge in distribution to the exponential law with rate $2\chi$.
\end{enumerate}
\end{proposition}
\begin{proof}
By \eqref{M-eq:capF}, in the variable $s^2$ the width is $1/(\Gamma^{+}_M-1)+1/(1+\Gamma^{-}_M)$ once $\Gamma^{+}_M>1$, where $\Gamma^{\pm}_M$ are the largest values of $\pm \gamma$ among the $M$ junctions, so it suffices to show that $\Gamma^{\pm}_M$ are of exact order $M$, or $\sqrt M$. This is a statement about the tails of $\gamma$ at a random junction. (1) Away from the edge-on lines of tilted axes $\gamma$ is bounded, and near such a line $\ell_k$ the expansion in the proof of \cref{M-thm:identframe} gives $|\gamma|\le C_1/|\zeta_k|+C_2$, because $\zeta_i$ and $\zeta_j$ are bounded and, the image containing no vanishing point, bounded away from zero. The set $\{|\gamma|>y\}$ therefore lies in bands of width $O(1/y)$ about those lines, and $\Pr(|\gamma|>y)\le\chi_2/y$. Conversely, on a segment of $\ell_k$ in the image and away from $\vc^{\star}_k$ the coefficient $\zeta_i^2z_k(w_i-w_j)$ is bounded away from zero, so on each side of the segment that lies in the image one of $\gamma_i$, $\gamma_j$ exceeds $c_1/|\zeta_k|-C_2$ and the other is below its negative; hence $\Pr(\gamma>y)\ge\chi_1/y$ and $\Pr(\gamma<-y)\ge\chi_1/y$ for large $y$. It follows that
\[
\begin{aligned}
\Pr\bigl(\Gamma^{\pm}_M\le\delta M\bigr)&\le\Bigl(1-\frac{\chi_1}{\delta M}\Bigr)^{M}\le e^{-\chi_1/\delta},\\
\Pr\bigl(\Gamma^{\pm}_M>M/\delta\bigr)&\le M\cdot\frac{\chi_2\delta}{M}=\chi_2\delta ,
\end{aligned}
\]
and taking $\delta$ small gives the claim. For the almost-sure statement, $W_M$ is nonincreasing because $\bigcap F$ only shrinks as junctions are added, and $\Gamma^{\pm}_M$ is nondecreasing. With $y=\chi_1M/(2\log M)$, $\Pr(\Gamma^{\pm}_M\le y)\le(1-\chi_1/y)^M\le e^{-\chi_1M/y}=M^{-2}$, which is summable, so by the Borel--Cantelli lemma $\Gamma^{\pm}_M>\chi_1M/(2\log M)$ for all large $M$, almost surely, and the width $1/(\Gamma^+_M-1)+1/(1+\Gamma^-_M)$ is at most $C'\log M/M$. In the other direction, $\Pr(\Gamma^+_{2^n}>2^nn^2)\le2^n\chi_2/(2^nn^2)=\chi_2/n^2$, which is summable, so $\Gamma^+_{2^n}\le2^nn^2$ for all large $n$; for $2^{n-1}<M\le2^n$, monotonicity gives $\Gamma^+_M\le2^nn^2\le2M\log_2^2(2M)$, and the width is at least $1/(\Gamma^+_M-1)\ge c'/(M\log^2M)$. (2) Near a corner of the image at which an edge-on line touches it, the vertices of the junctions lie in a wedge, in which the set $\{|\zeta_k|<1/y\}$ has area of exact order $1/y^2$. At a corner that is the point $\vc^{\star}_k$ of its line $\gamma$ is bounded, by the proof of \cref{M-thm:identframe}; at any other, the expansion there gives $\chi_1/y^2\le\Pr(\pm \gamma>y)\le\chi_2/y^2$ for large $y$. Hence $\Pr(\Gamma^{\pm}_M\le\delta\sqrt M)\le e^{-\chi_1/\delta^2}$ and $\Pr(\Gamma^{\pm}_M>\sqrt M/\delta)\le\chi_2\delta^2$, and $\Gamma^{\pm}_M$ is of exact order $\sqrt M$.
(3) The coordinates $\zeta_b$ are affine functions of the image point, so near $\vv_a$ the pair $(u,v)=(\zeta_p,\zeta_q)$ is an affine coordinate system centered at $\vv_a$, in which the image contains the square $|u|,|v|<\eta$ for some $\eta>0$, and $\zeta_a$ stays between $\zeta_0/2$ and $2\zeta_0$ for some $\zeta_0\neq0$. There
\[
\gamma_a=\frac{c}{uv}-z_a\zeta_a\Bigl(\frac{z_p}{u}+\frac{z_q}{v}\Bigr)+z_a^2,\qquad c=\zeta_a^2z_pz_q\neq0,
\]
so that $|\gamma_a-c/(uv)|\le C(1/|u|+1/|v|)+C$ for a constant $C$. For the upper bound, $|\gamma_a|>y$ with $y>3C$ forces $|uv|<3|c|/y$, $|u|<3C/y$ or $|v|<3C/y$, sets of area $O((\log y)/y)$, $O(1/y)$ and $O(1/y)$ in the square; $\gamma_p$ and $\gamma_q$ obey the same bound, and away from $\vv_a$ the tails are $O(1/y)$ by part (1). For the lower bound, take $a=|c|/4$ and $b=2C$ and the quadrant in which $uv$ has the sign of $\pm c$, which lies in the image because $\vv_a$ is interior. On its subset $\{b/y\le|u|,|v|\le\eta,\ |uv|\le a/y\}$, the main term has modulus at least $4y$ and the rest at most $y+C$, so $\pm\gamma_a>y$ for $y\ge C$; and the subset has area $(a/y)\log(ay/b^2)-O(1/y)$ for large $y$. Hence $\Pr(\pm\gamma>y)$ is of exact order $(\log y)/y$. In the argument of part (1), with $y=\delta M\log M$, $\Pr(\Gamma^{\pm}_M\le\delta M\log M)\le(1-\chi_1\log y/y)^M\le e^{-\chi_1/(2\delta)}$ for large $M$, and with $y=M\log M/\delta$, $\Pr(\Gamma^{\pm}_M>M\log M/\delta)\le M\chi_2\log y/y\le2\chi_2\delta$; so $\Gamma^{\pm}_M$ is of exact order $M\log M$.
 (4) is proved in \cref{app:limit}.\end{proof}

The statement is in probability because $W_M$ has no finite expectation: with a small but positive probability no junction among the $M$ is bounded above, and the width is then infinite. Part (4) makes the rate concrete: for the frame seen down a cube diagonal at a $70^\circ$ field of view, $\chi=0.234$, and over $400$ runs of $20\,000$ uniform junctions $M(f_{\max}/f-1)$ had mean $2.10$ and median $1.51$, against $1/(2\chi)=2.14$ and $\ln2/(2\chi)=1.48$; for an oblique frame at $75^\circ$, $\chi=0.567$ and the means were $0.90$ against $0.88$. A vanishing point inside the image speeds the bounds up by a logarithmic factor, part (3): in simulation, with one vanishing point inside the image and both other axes tilted, $y\Pr(\gamma>y)$ grew linearly in $\log y$, from $1.04$ at $y=10^2$ to $2.00$ at $y=10^5$.

All of the forcing, and these rates, come from junctions close to an edge-on line, where two of a junction's edges are nearly collinear in the image. A detector that discards junctions whose smallest angle is below some $\alpha_{\min}$, and noise of scale $\sigma$ on edges of length $\ell_{\mathrm{seg}}$, remove exactly these, so in practice the width stops shrinking at a level set by $\max(\alpha_{\min},\sigma/\ell_{\mathrm{seg}})$; the identifiability results describe noiseless junctions.

\subsubsection{Proof of the limit law}\label{app:limit}

This subsection proves \cref{prop:rate}(4). Fix an axis $k$ whose edge-on line $\ell_k$ passes through $\mathcal I$, and write $i$ and $j$ for the other two. No vanishing point lies in $\mathcal I$, so the other edge-on lines keep a positive distance from $\ell_k\cap \mathcal I$; and no edge-on line contains an edge of $\mathcal I$, so near each interior point of the segment $\ell_k\cap \mathcal I$ both sides of $\ell_k$ lie in $\mathcal I$. Use coordinates $(\varsigma,\delta)$, with $\varsigma$ the signed distance along $\ell_k$ from $\vc^{\star}_k$ and $\delta$ the signed distance from $\ell_k$, both in pixels.

\emph{Step 1: the exact form near $\ell_k$.} By \cref{M-thm:boundmap}, $\gamma_i=g_i\,\varsigma/\delta$ and $\gamma_j=g_j\,\varsigma/\delta$, with $g_i=-R_j/\ell_j$ and $g_j=-R_i/\ell_i$, because $R_k/\ell_k=\varsigma/\delta$ exactly: $R_k$ is the perpendicular to $\ell_k$ through $\vc^{\star}_k$, and both have normals of length $|\bar\vd_k|/f$. Choose $\eta>0$ so small that $\ell_i$ and $\ell_j$ do not vanish on $N_\eta=\{|\delta|<\eta\}\cap \mathcal I$. There $g_i$ and $g_j$ are smooth, with $|g|\le G$ and $|\partial_\delta g|\le L$, and $\gamma_k$ is bounded, say by $B$. On $\ell_k$, $g_ig_j=-z_k^2$ (proof of \cref{M-thm:identframe}), with $|g_i|=|z_k|r_k$ and $|g_j|=|z_k|/r_k$; shrinking $\eta$, $g_ig_j<0$ on all of $N_\eta$.

\emph{Step 2: the set where $\gamma$ is large.} Let $y>\max(B,1)$ and fix $\varsigma\neq0$. Then $\gamma_i>y$ requires $\delta$ to have the sign of $\varsigma g_i$ and $|\delta|<|\varsigma||g_i(\varsigma,\delta)|/y$. Since $|\varsigma||g_i(\varsigma,\delta)|/y-|\delta|$ decreases in $|\delta|$ once $y>L\sup|\varsigma|$, the $\delta$-section of $\{\gamma_i>y\}$ is an interval from $0$ to $|\varsigma|\bigl(|g_i(\varsigma,0)|+O(1/y)\bigr)/y$, uniformly in $\varsigma$. On that side $\gamma_j<0$, and on the other side the roles of $i$ and $j$ are exchanged, so the section of $\{\gamma^+>y\}\cap N_\eta$ has total length
\[
\begin{aligned}
&\frac{|\varsigma|\bigl(|g_i|+|g_j|\bigr)(\varsigma,0)}{y}+O\!\left(\frac1{y^2}\right)\\
&\qquad=\frac{|z_k||\varsigma|\,(r_k+1/r_k)}{y}+O\!\left(\frac1{y^2}\right).
\end{aligned}
\]
The sections lie in $\mathcal I$ except for $\varsigma$ within $O(1/y)$ of the two ends of the segment, which changes the area by $O(1/y^2)$.

\emph{Step 3: the tail.} The density $p$ is continuous, so it differs from $p(\varsigma,0)$ by $o(1)$, uniformly, on sections of length $O(1/y)$, and
\[
\begin{aligned}
&\Pr\bigl(\gamma^+>y,\ \vc\in N_\eta\bigr)\\
&\quad=\frac1y\int p(\varsigma,0)\,|z_k||\varsigma|\Bigl(r_k+\frac1{r_k}\Bigr)d\varsigma+o\!\left(\frac1y\right).
\end{aligned}
\]
Outside the sets $N_\eta$ of the lines that pass through $\mathcal I$, $\gamma$ is bounded except near lines that only touch $\mathcal I$, where $\Pr(|\gamma|>y)=O(1/y^2)$ by part (2). Summing over $k$ gives $y\Pr(\gamma^+>y)\to\chi$, and the same holds for $-\gamma$, the two sides of each line exchanging roles.

\emph{Step 4: the extremes.} The junctions are independent, so for every $y>0$, $\Pr(\Gamma^+_M\le My)=\bigl(1-\Pr(\gamma^+>My)\bigr)^M\to e^{-\chi/y}$; that is, $M/\Gamma^+_M$ converges in distribution to the exponential law with rate $\chi$, and so does $M/\Gamma^-_M$. By \eqref{M-eq:capF}, $f_{\max}/f=\bigl(\Gamma^+_M/(\Gamma^+_M-1)\bigr)^{1/2}=1+1/(2\Gamma^+_M)+O\bigl((\Gamma^+_M)^{-2}\bigr)$, so $M(f_{\max}/f-1)=M/(2\Gamma^+_M)+O_p(1/M)$, which converges to half an exponential variable of rate $\chi$, that is, to the exponential law with rate $2\chi$; the same argument applies to $1-f_{\min}/f$. Only the two laws separately are claimed: $\Gamma^+_M$ and $\Gamma^-_M$ are driven by the same junctions near the edge-on lines and are not independent.

\subsubsection{Simulations}
Over $24$ random frames at a $70^\circ$ field of view, the median width of $\bigcap F$ fell from $0.18f$ at $20$ noiseless junctions to $9.4\times10^{-5}f$ at $20\,000$, with a least-squares log-log slope of $-0.981$ over $200$ to $20\,000$ junctions, against the $-1$ of \cref{prop:rate}.

The frame seen down a cube diagonal shows that \cref{M-thm:identframe} is sharp (\cref{fig:transition}). The first of its vanishing lines to reach the sampled square (half-width $480$\,px) does so at $f=960$\,px, a $55.0^\circ$ field of view for the $1000$\,px image, and reaches it at the line's exceptional point. Over $30$ replicates per size, the median width at $54^\circ$, below the transition, stays at $0.61f$, $0.55f$ and $0.54f$ at $10^3$, $3\times10^4$ and $3\times10^5$ junctions, a log-log slope of $-0.01$; at $56^\circ$, $58^\circ$, $60^\circ$, $65^\circ$ and $70^\circ$ it falls with least-squares slopes of $-0.92$, $-1.01$, $-0.91$, $-1.03$ and $-1.13$ between $3\times10^4$ and $3\times10^5$ junctions, each $95\%$ bootstrap interval containing $-1$ (Sect.~S11). At the transition itself the width stalls at $0.46f$ over $2\times10^{6}$ junctions. Rotating the frame by $12^\circ$ about the optical axis moves the exceptional point off the corner: the line then first touches the image at $f=939$\,px, and the width falls with a least-squares log-log slope of $-0.52$ over $3\times10^{4}$ to $2\times10^{6}$ junctions, against $-0.99$ once the line passes through the image, as \cref{prop:rate} predicts. Sect.~S11 gives further runs.

\part{Computation and experiments}

\section{Protocol, noise model and identifiability}\label{sec:Sprotocol}

\subsection{Protocol}

Images are $1000\times1000$\,px with the principal point at the center, at fields of view of $40^\circ$, $60^\circ$ and $90^\circ$ ($f=1374$, $866$ and $500$\,px). Corners are sampled uniformly over orientations and over the image, at depths of 2 to 6 units. Each edge is observed as a $150$\,px segment starting at the junction, and both endpoints carry Gaussian noise of standard deviation $\sigma\in\{0,0.5,1,2\}$\,px. Each image line is fitted through its two noisy endpoints, and the junction point is the least-squares intersection of the three fitted lines. Each fitted line is then translated, with its direction kept, to pass through the estimated junction, so that every interpretation plane contains the ray, as the derivation requires; this changes the coverage figures of \cref{M-sec:e2} by at most $0.1$ percentage point. Where corners are binned by position, the binning variable is the distance from the principal point normalized by the image half-width, so that the inner third is $\rho<167$\,px and the outer third $\rho>333$\,px. The code and its random seeds are supplied (see the Declarations), and Sect.~S11 reports the numerical checks of the proved results.

\begin{figure}[t]
\centering
\includegraphics[width=\linewidth]{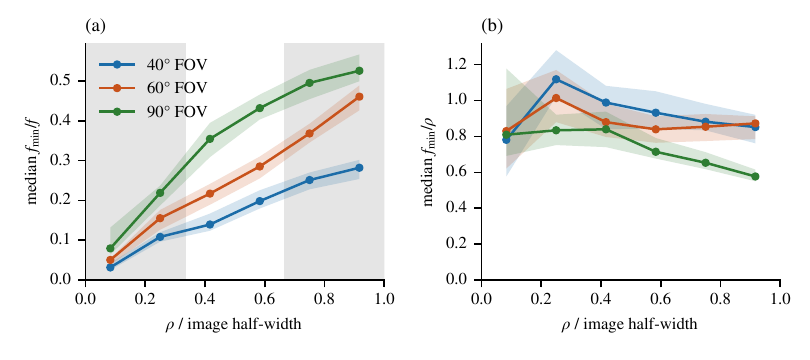}
\caption{(a) Median lower bound $\fmin/f$ from one corner against the corner's distance from the principal point in units of the image half-width, at three fields of view, with bootstrap 95\% intervals on the median ($3000$ corners per field of view, a larger sample than that of \cref{M-sec:e2}, $\sigma=0.5$\,px); the shaded bands are the inner and outer thirds used in \cref{M-sec:e2}, and corners farther than one half-width, toward the image's corners, are omitted. (b) The same lower bounds in units of the corner's distance $\rho$, which vary far less with position than in (a), as \cref{M-cor:similar} predicts.}
\label{fig:boundmirror}
\end{figure}

\Cref{fig:boundmirror} shows the lower bound of the article's \cref{M-sec:e2} against the distance from the center, in units of $f$ and of $\rho$.

\subsection{Robustness of the bound}

The bound is robust to the idealizations of the protocol: coverage stays above $96\%$ with the principal point off by up to $20$\,px and falls to $94\%$ at $40$\,px; it falls to $89\%$ with $20$\,px edges; and tilting the vertical as in York Urban changes neither the coverage nor the share of junctions bounded above (Sect.~S13).

\paragraph{A misplaced principal point.} The focal-length results assume the principal point known, and in practice it is usually taken as the image center. To measure the cost of an error in it, the junctions of the protocol were measured relative to an assumed principal point offset from the true one by $0$, $5$, $10$, $20$ and $40$\,px in a random direction, with the endpoint noise of $0.5$\,px on $150$\,px edges modeled as angle noise, over $1000$ corners per field of view. The admissible set contained the true focal length for $96$, $97$, $96$, $98$ and $96\%$ of corners at $40^\circ$, for $97$, $98$, $97$, $96$ and $94\%$ at $60^\circ$, and for $97$, $97$, $95$, $94$ and $89\%$ at $90^\circ$. An offset of up to $20$\,px therefore leaves the coverage essentially unchanged, and only a wide lens with an offset of $40$\,px loses a noticeable share. The script \texttt{pp\_offset.py} of Online Resource~2 regenerates these numbers.

\subsection{The first-order noise model}\label{sec:e3}
Two checks address \cref{prop:noise}. First, we compared the closed form \eqref{eq:sd} with the standard deviation obtained by differentiating the whole pipeline, from the twelve endpoint coordinates through the fitted lines and the estimated junction to the threshold. Over 468 thresholds, the ratio of the two has median $1.004$, with 10th and 90th percentiles $0.88$ and $1.07$; the junction's own uncertainty therefore contributes little, and the edge angles carry the error. Second, at a $60^\circ$ field of view, with $200$ perturbations of each of $600$ corners, the predicted miss probability \eqref{eq:miss} was $3.3\%$, $4.9\%$ and $8.3\%$ at $\sigma=0.5$, $1$ and $2$\,px, against $3.2\%$, $4.6\%$ and $7.9\%$ measured; the threshold-based approximation it replaces gave $3.7\%$, $5.5\%$ and $7.1\%$. This agreement is consistent with the coverage of 97 to 98\% in \cref{M-sec:e2} and of 98.9\% in \cref{sec:e4}.

\subsection{Identifiability from the bounds alone}\label{sec:e5}
\begin{figure}[t]
\centering
\includegraphics[width=\linewidth]{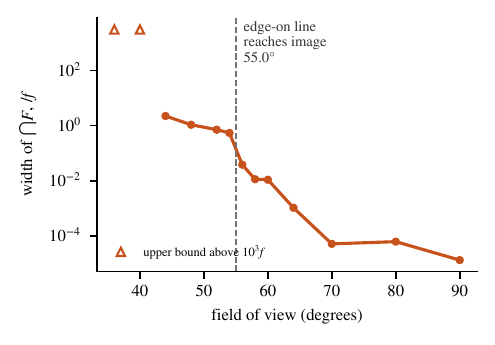}
\caption{Width of the intersection of the admissible sets of $20\,000$ noiseless junctions against field of view, for a frame seen down a cube diagonal on a square image. At $36^\circ$ and $40^\circ$ the tightest upper bound exceeds $10^3f$ (triangles); from $44^\circ$ the intersection is bounded but stalls while no edge-on line passes through the image, and it falls once one does, above $55.0^\circ$ (\cref{M-thm:identframe})}
\label{fig:transition}
\end{figure}

The frame seen down a cube diagonal shows that \cref{M-thm:identframe} is sharp (\cref{fig:transition}). The first of its vanishing lines to reach the sampled square (half-width $480$\,px) does so at $f=960$\,px, a $55.0^\circ$ field of view for the $1000$\,px image, and it reaches the square at the line's exceptional point. Below the transition, the width of the intersection stops shrinking, at $0.54f$ over $3\times10^{5}$ junctions at $54^\circ$; at the transition itself it stalls at $0.46f$ over $2\times10^{6}$ junctions; above it, the width falls toward zero as junctions are added. How fast it falls for junctions placed at random is the subject of Sect.~S9, which also reports the replicated runs.

\section{Numerical checks of the proved results}\label{sec:Schecks}

The results of the article are proved; the checks below test the statements, and the code that implements them, against direct computation. The protocol is that of \cref{M-sec:experiments} of the article unless stated.

\paragraph{\Cref{M-thm:convex}.} A direct check over $2.25$ million sampled cameras, with all five entries of $K$ perturbed, found the predicted cell matching the admissible set exactly, with no admissible camera outside it and none inside it inadmissible. The criterion \eqref{M-eq:nuk} agreed with the definition in each of $200\,000$ further trials.

\paragraph{\Cref{M-lem:frame} and \cref{M-thm:converse}.} \Cref{M-lem:frame} was checked against the definition on $100\,000$ random cameras, frames and junctions, with all five intrinsic parameters varied, without exception. For a camera tilted by $4^\circ$ whose horizontal vanishing points fall outside the image, only the horizon passes through it. There, with $N_{aa}=1$, an entry $N_{13}$ or $N_{23}$ of $10^{-3}$ makes some junction inadmissible, while $N_{12}$ ranges over $[-0.88,\,0.28]$, which is the interval the inequalities \eqref{M-eq:nuframe} give, at each of $301$ values tested. With $f=1440$\,px on a square image of side $960$\,px and a frame whose third vanishing line passes through one corner of the image and nowhere else, the other two missing it, the camera with $N_{aa}=1$, $N_{12}=0$, $N_{23}=0.2$ and $N_{13}$ fixed by \eqref{M-eq:touch} admitted all $600\,000$ junctions sampled, down to $10^{-9}$\,px from the corner, although the cosines between its vanishing points are $-0.14$ and $-0.20$; changing $N_{13}$ by $10^{-6}$ made junctions near the corner inadmissible. One frame accordingly determines a camera model with at most that many parameters. The single focal length needs one, and \cref{M-thm:identframe} says when it gets it. The aspect-ratio camera needs two, and is recovered from a single frame to $f_x\in[899.8,900.2]$ and $f_y\in[699.8,700.0]$ by $4000$ junctions. The general camera needs five, which one frame never supplies. From one frame its feasible region stayed frozen across $100$ to $6000$ junctions and reached degenerate cameras, with smallest eigenvalue near zero. After $4000$ junctions the feasible cameras made the cosine between vanishing-point directions $3.5\times10^{-4}$, against a median of $0.42$ for random cameras.

\paragraph{\Cref{M-thm:frames}.} We checked both cases with the cutting-plane program of \cref{M-sec:program}. For two frames that differ by a rotation of $37^\circ$ about a common axis, each member tested of the family $K(I+t\,\vd\vd^{\top})K^{\top}$, with $t$ from $-0.6$ to $2$, admitted all $8000$ sampled junctions. The feasible region of $3000$ junctions stayed open, with a worst relative extent over the entries of $\omega^{*}$ of $43$. For two frames with no common axis, no member tested other than $t=0$ was admitted, and the extent fell from $0.81$ at $600$ junctions to $0.14$ at $3000$. The criterion of part (2) agreed with the rank of the linear conditions in each of $4000$ random sets of one to five axes, half of them constructed to be degenerate. With $f_x=900$, $f_y=700$, a skew of $15$ and a principal point of $(40,-30)$\,px, on a square image of side $960$\,px, three images at pitches of $4^\circ$, $11^\circ$ and $-7^\circ$, each showing only its horizon, admitted the camera with $t=0.2$ at every junction sampled; once the roll was varied as well, by $6^\circ$, $-5^\circ$ and $9^\circ$, none of the members tested was admitted.

A touching axis adds one condition where a visible one adds two. Applied to the cameras $\operatorname{diag}(f',f',1)$, for which $\Xi=s^2I+(1-s^2)\hat{\mathbf z}\hat{\mathbf z}^{\top}$ with $s=f'/f$, the conditions of \cref{M-thm:frames} reproduce \cref{M-thm:identframe}, its exceptional point included. A visible axis $\vd$ is an eigenvector of this $\Xi$ only if $s=1$, unless $z=\vd\cdot\hat{\mathbf z}=0$ or $\vd=\pm\hat{\mathbf z}$. For a touching axis, $(\Xi\vd)\cdot(\vd\times\vr_0)=(1-s^2)\,z\,\hat{\mathbf z}\cdot(\vd\times\vr_0)$, which vanishes for $s\neq1$ only if $z=0$ or $\hat{\mathbf z}$, $\vd$ and $\vr_0$ are coplanar; since $\vr_0\perp\vd$, the last happens exactly when $\vr_0$ is proportional to $\hat{\mathbf z}-z\vd$, the ray of the point $\vc^{\star}_k$. The condition of part (1) agreed with \eqref{M-eq:touch} to rounding error in each of $3000$ random cases. At the level of junctions, with a longer focal length, $f_x=1800$\,px, so that a frame can have one vanishing line touching a corner of the image and the other two missing it, the three images of a camera that only pitches again admitted the family above; adding such a frame, whose axis satisfies $\mathbf n\cdot\vd\neq0$ and $\mathbf n\cdot(\vd\times\vr_0)\neq0$, removed it, the smallest $\nu_a$ near the corner being $-6\times10^{10}$ at $t=0.05$. For that frame alone, the cameras $K(I+t\,\hat\vr_0\hat\vr_0^{\top})K^{\top}$, which satisfy its condition, were admitted for $t$ up to $0.05$ and rejected only at $t=0.2$, while $K(I+t\,\mathbf e_1\mathbf e_1^{\top})K^{\top}$, which violates it, was rejected at every $t$ tested, down to $10^{-3}$.

\paragraph{\Cref{M-prop:gform}.} Over $40$ random cameras and frames with $3000$ junctions each, \eqref{M-eq:capF} reproduced the intersection of the single-junction intervals of \cref{M-thm:interval} to $10^{-13}$ at both ends.

\paragraph{\Cref{M-thm:fpp}.} The characterization agreed with the rank of the linearized conditions in each of $1500$ sets of axes covering all the configurations above. At the level of junctions, with $f=250$\,px so that all three vanishing lines of a frame with one axis parallel to the image plane are in view, the cameras of the family with $h=0.05$ and $0.12$ admitted every junction sampled, while changing the focal length alone by $5\%$ did not. An image at a pitch of $6^\circ$ showing only its horizon admitted its family, and adding a second at a pitch of $13^\circ$, again showing only its horizon, excluded it.

\paragraph{Identifiability from the bounds alone.} Corners of one scene at $60^\circ$, noiseless, with their admissible sets intersected as they arrive. The intersection closes on the true focal length from both sides (\cref{tab:close}), as \cref{M-thm:identframe} predicts.

\begin{table*}[t]
\centering\small
\caption{The intersection of the admissible sets of the corners of one scene, at $60^\circ$ and without noise, as corners accumulate}
\label{tab:close}
\begin{tabular}{lccccc}
\toprule
corners & 10 & 100 & 1000 & 5000 & 20000\\
\midrule
tightest lower bound $/f$ & 0.9826 & 0.9826 & 0.99960 & 0.99996 & 0.999962\\
tightest upper bound $/f$ & 1.3932 & 1.0060 & 1.00015 & 1.00015 & 1.000020\\
relative width            & 0.411  & 0.0234 & 0.00055 & 0.00019 & 0.000058\\
\bottomrule
\end{tabular}
\end{table*}

Holding the Manhattan frame fixed and varying only the corner's position in the image, $40\,000$ corners give $[0.999965,\,1.000028]\times f$, a width of $6.3\times10^{-5}$; letting the frames vary as well gives $1.2\times10^{-5}$ over the same count. This is the noiseless rate; \cref{sec:e6} is the corresponding statement under noise, where strict intersection fails and a consensus rule is needed.

\paragraph{The condition of \cref{M-thm:identframe} is sharp.} Take the frame seen down a cube diagonal, whose three axes make equal angles with the optical axis, so that \cref{M-thm:identframe} predicts identifiability exactly when the edge-on plane of some axis enters the image. Sampling $30\,000$ noiseless corners at each field of view, with the corner's position as the only variable, the intersection of the admissible sets is given in \cref{tab:sharp}.

\begin{table*}[t]
\centering\small
\caption{Width of the intersection of admissible sets for a frame seen down a cube diagonal, from $30\,000$ noiseless corners at each field of view}
\label{tab:sharp}
\begin{tabular}{lccccccc}
\toprule
field of view & $40^\circ$ & $46^\circ$ & $50^\circ$ & $54^\circ$ & $60^\circ$ & $73^\circ$ & $90^\circ$\\
\midrule
edge-on plane in view & no & no & no & no & yes & yes & yes\\
width of $\bigcap F$, $/f$ & $7\times10^{3}$ & $1.39$ & $0.88$ & $0.55$ & $3.8\times10^{-3}$ & $7.3\times10^{-5}$ & $2.4\times10^{-5}$\\
\bottomrule
\end{tabular}
\end{table*}

The transition falls between $54^\circ$ and $60^\circ$, which is where the edge-on plane first reaches the image's corners: a square image sees further off-axis at its corners than at its edges, so the threshold sits below the $70.53^\circ$ that a circular field would need. That the $54^\circ$ width is a genuine limit and not undersampling is checked by multiplying the sample a hundredfold: it reads $0.551$, $0.537$ and $0.543$ at $3000$, $30\,000$ and $300\,000$ corners, while at $60^\circ$ it keeps falling, through $0.152$, $0.0021$ and $0.0013$. These figures and those of \cref{tab:sharp} come from different random samples of corners; at a given number of corners the width is itself random (\cref{prop:rate}), and at $60^\circ$ and $30\,000$ corners the two samples differ by a factor of about two.

\paragraph{Replicated widths for the cube diagonal.} \Cref{tab:Stransition} gives the median width of $\bigcap F$, in units of $f$, over $30$ replicates per size, for the frame seen down a cube diagonal on a square image of half-width $480$\,px, computed from the closed form of \cref{M-prop:gform} of the article. The slopes are least-squares fits over $3\times10^4$ to $3\times10^5$ junctions, with $95\%$ bootstrap intervals over replicates.

\begin{table}[h]
\centering\footnotesize
\caption{Median width of $\bigcap F$ for the cube-diagonal frame, $30$ replicates per size}
\label{tab:Stransition}
\begin{tabular}{lccccccc}
\toprule
FOV & $10^3$ & $3\cdot10^3$ & $10^4$ & $3\cdot10^4$ & $10^5$ & $3\cdot10^5$ & slope\\
\midrule
$54^\circ$ & 0.61 & 0.61 & 0.57 & 0.55 & 0.55 & 0.54 & $-0.01$ $(-0.01,0.00)$\\
$56^\circ$ & 0.45 & 0.43 & 0.29 & 0.099 & 0.025 & 0.012 & $-0.92$ $(-1.30,-0.64)$\\
$58^\circ$ & 0.27 & 0.099 & 0.037 & 0.0095 & 0.0021 & 0.00093 & $-1.01$ $(-1.25,-0.79)$\\
$60^\circ$ & 0.063 & 0.038 & 0.015 & 0.0043 & 0.0014 & 0.00052 & $-0.91$ $(-1.28,-0.57)$\\
$65^\circ$ & 0.0070 & 0.0019 & 0.00040 & 0.00026 & 0.000055 & 0.000024 & $-1.03$ $(-1.28,-0.84)$\\
$70^\circ$ & 0.0032 & 0.00086 & 0.00028 & 0.00013 & 0.000030 & 0.0000095 & $-1.13$ $(-1.34,-0.92)$\\
\bottomrule
\end{tabular}
\end{table}

\subsection{Worked examples of the convex program}\label{sec:Sprogram}

For a camera with known principal point and zero skew, $\omega^{*}=\operatorname{diag}(X,Y,1)$ and positive-definiteness reduces to $X,Y>0$, so the cameras consistent with $M$ junctions form the feasible set of a linear program. Its extent in $X$ and in $Y$ is the value of a further linear program, so no grid is needed. With a true camera of $f_x=900$, $f_y=700$ and junctions from independent frames, $1000$ junctions pin $f_x$ to $[899.8,\,902.5]$ and $f_y$ to $[699.8,\,701.9]$.

For the full camera, with skew and principal point free, positive-definiteness is a matrix inequality, which we impose with cutting planes: we solve the linear program and, whenever the solution is not positive-definite, add the cut $\mathbf v^{\top}\omega^{*}\mathbf v\ge0$, where $\mathbf v$ is the eigenvector belonging to the solution's most negative eigenvalue. This requires nothing beyond a linear-programming solver. With a true camera of $f_x=900$, $f_y=700$, skew $15$ and principal point $(40,-30)$, and $2000$ junctions from independent frames, every parameter is recovered to within about one pixel: $f_x\in[899.3,901.1]$, $f_y\in[699.5,700.9]$, skew $\in[14.86,15.20]$, $p_x\in[39.52,40.05]$ and $p_y\in[-30.24,-29.83]$. These junctions are noiseless. The linear programs are solved in a box of side $10^8$ about the origin, cuts are added until the smallest eigenvalue is nonnegative to relative tolerance $10^{-9}$, and each interval is the range of the parameter over the ten feasible cameras that extremize the entries of $\omega^{*}$. Since $f_x$, $f_y$ and the skew are not linear in $\omega^{*}$, these are inner estimates of the feasible ranges. From the junctions of a single frame, by \cref{M-prop:oneframe}, the same program cannot close on a general camera; \cref{sec:e4} applies a robust form of it to real junctions.

\section{Real images: the York Urban Database}\label{sec:e4}
For the York Urban camera, whose horizontal field of view is $51^\circ$ (diagonal $61^\circ$), two properties can be checked: that the bound holds, and that upper bounds are rare (\cref{M-prop:bound}(2)). Both hold, and the last implies that the junctions do not calibrate this camera.

The York Urban Database \citep{denis2008}, a standard benchmark for vanishing-point estimation \citep{tardif2009}, supplies 102 images at $640\times480$ with ground-truth intrinsics ($f=674.92$\,px, principal point $(307.55,251.45)$) and hand-labeled line segments assigned to the three Manhattan directions. Junctions are formed by clustering segment endpoints and are kept only if the three fitted lines concur within 3\,px and every near endpoint lies within 12\,px of the junction. From 12\,122 segments, this procedure yields 3450 two-family junctions but only 181 full three-family junctions, 1.8 per image, in 59 of the 102 images; other extraction parameters give 161 to 181 (Sect.~S13). This scarcity matters more than anything else measured here.

\begin{figure*}[t]
\centering
\includegraphics[width=\linewidth]{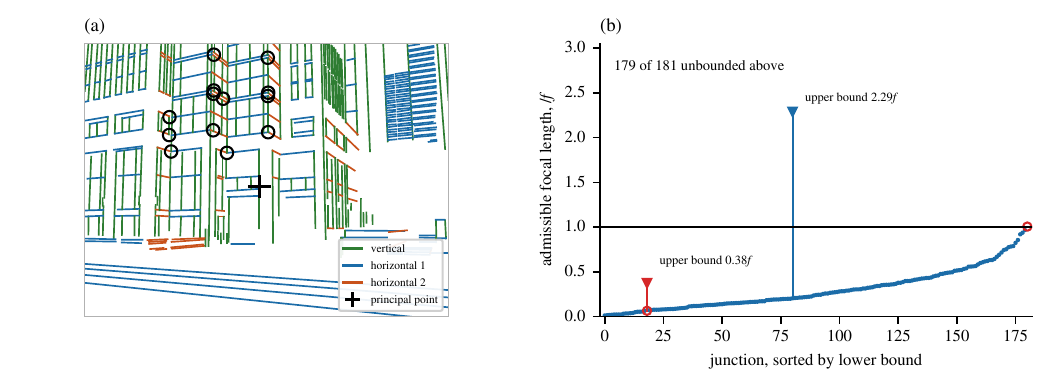}
\caption{The bound on real images, York Urban Database. (a) The richest image, holding $13$ of the $181$ full junctions: hand-labeled segments colored by Manhattan direction (key at lower right), with the junctions circled, all of which admit the true focal length. (b) The admissible interval of each of the $181$ junctions, sorted by lower bound, against the true $f$ (black line). Dots are lower bounds, and $179$ intervals are unbounded above. The two red marks are the misses, a lower bound at $1.006f$ and an interval bounded above at $0.38f$; the only other upper bound is at $2.29f$}
\label{fig:real}
\end{figure*}

\paragraph{Near-duplicate junctions.} Eight pairs of these junctions lie within $8$\,px of each other in the same image, with all three edge directions agreeing to within $3^\circ$: either one corner labeled twice, or parallel edges a few pixels apart, such as the inner and outer edges of a frame. They are kept in the counts below. Dropping one junction of each pair leaves $173$ junctions in the same $59$ images; the admissible set then contains the true focal length for $171$ of them ($98.8\%$), the certified sets of the article's \cref{M-prop:certified} for all of them, still only two junctions bound the focal length from above, and the median lower bound is $0.24f$ instead of $0.23f$. No conclusion of this section changes.

\paragraph{Coverage and the sign test.} The admissible set contained the true $f$ for 179 of the 181 junctions, 98.9\% (Wilson 95\% interval 96.1 to 99.7\%), against 97 to 98\% in simulation (\cref{fig:real}); this is the sign test of \cref{M-sec:sign} under the true $K$, not independent evidence. As a detector, the camera-free test accepted all 181 genuine junctions and $35$ of the $39$ three-family endpoint clusters that failed validation. The median $\fmin/f$ is $0.24$ (quartiles $0.13$ and $0.43$), and the set was unbounded above for 98.9\% of junctions, against 66 to 90\% in simulation (\cref{M-sec:e2}).

\paragraph{Certified intervals.} Bounding each edge's angle error by $\varphi_i=\arcsin(2\varepsilon/\ell_i)$, with $\ell_i$ the length of its own segment and $\varepsilon$ an assumed endpoint accuracy, \cref{M-prop:certified} gives a certified set for every junction. For $\varepsilon=0.25$\,px this set contains the true focal length for all $181$ junctions, including the two that the measured intervals miss; these are covered from $\varepsilon=0.10$ and $0.20$\,px, and the second has an edge only $15$\,px long. The guarantee costs little from below, lowering the median lower bound by $12\%$ at $\varepsilon=0.25$\,px and by $35\%$ at $1$\,px, but it removes both upper bounds. The vertices are estimated from the fitted lines rather than known exactly, as the proposition assumes, so this is a consistency check of the certified bounds on real data rather than a guarantee.

\paragraph{Against vanishing points.} Classical calibration from three vanishing points, fitted by least squares to the labeled families with the principal point taken as their orthocenter, reaches a median error in $f$ of $2.1\%$ (75th percentile $6.9\%$) on $89$ of $97$ images, and $1.7\%$ on $56$ of the $59$ junction images; a RANSAC comparison of junction sampling with the line sampling of \citet{wildenauer2012} is in \cref{sec:Ssampling}. Junctions are useful where line families are not available, not where they are.

\paragraph{The bounds do not calibrate this camera.} The 181 junctions come from one camera and can be pooled. In principle, the junctions of any two of these images determine the focal length and principal point (Sect.~S5), and those of all $59$ determine a general camera (\cref{M-thm:frames}). The vertical directions, however, lie within $18.6^\circ$ of their mean, so the conditions are nearly dependent (Sect.~S13), and we asked the junctions only for the focal length. Only two of the $181$ junctions bound it from above, one wrongly at $0.38f$ and the other at $2.29f$, while the lower bounds climb to $0.97f$ and $1.006f$. The strict intersection is therefore empty. Rules fixed in advance do no better: the depth-$(M-1)$ consensus, which tolerates one wrong interval, is $[1.006f,\,2.29f]$ and misses $f$, while the depth-$(M-2)$ consensus is $[0.97f,\infty)$, tight from below and uninformative from above, as noiseless simulations of level scenes predict (the largest of $181$ lower bounds has median $0.982f$). Combined by the log-plausibility of Sect.~S8, the junctions overestimate $f$ by $47$ to $67\%$.

\paragraph{Why upper bounds are rare.} The camera's tilt alone does not explain it: tilting a simulated vertical as in York Urban, by a median of $3.2^\circ$, still leaves about $20\%$ of junctions bounded above at $60^\circ$, against $1.1\%$ here (Sect.~S13). The explanation is where the corners lie. Keeping each junction at its position but replacing its measured lines by the noiseless lines of the image's ground-truth frame bounds only four junctions from above, at $2.06f$, $2.43f$, $3.59f$ and $4.40f$. Likewise, the region of \cref{M-thm:boundmap} in which a junction bounds $f$ from above covers a median $9.4\%$ of these images but holds only $2.2\%$ of their junctions, even though in $22$ of the $59$ images it reaches into the image along the edge-on line of a tilted horizontal axis. Real corners avoid the neighborhoods of the edge-on lines. The failure to calibrate this camera is a matter of where the junctions lie, not of noise.


\subsection{Junction sampling against line sampling}

Two RANSAC loops solve the same quadratic \eqref{eq:cfsolver}, one sampling a junction plus one parallel line and the other four lines from the labeled families. On the $59$ images where both return an estimate, their median errors in $f$ are $4.6\%$ and $3.5\%$, and line sampling, as in \citet{wildenauer2012}, wins on $41$ of them (two-sided sign test, $p=0.004$). Classical calibration from three vanishing points, with the principal point taken as their orthocenter, does better still: $2.1\%$ (75th percentile $6.9\%$) on $89$ of $97$ images, and $1.7\%$ on $56$ of the $59$ junction images. Junctions are useful where line families are not available, not where they are.

\subsection{The log-plausibility estimator on York Urban}
The log-plausibility \eqref{eq:pi}, with the spread \eqref{eq:sd} computed from each edge's own length, returns $1126$\,px with the true principal point and $995$\,px with the image center, $67\%$ and $47\%$ too large, with bootstrap intervals that reach the limit of the search range. A robust form of the convex program of \cref{M-sec:program}, in which every constraint gets a slack variable and the total violation is minimized (violations measured as cosines between interpretation planes), returns $677$\,px against a true $675$. That agreement is accidental: bootstrap resamples that include the image containing the junction with the wrong upper bound give a median of $677$\,px, and those that exclude it give $1164$\,px, $72\%$ too large. The single wrong upper bound was supplying the only pressure from above. For the general camera the same program returns $f_x$ and $f_y$ $25\%$ and $33\%$ too large.

\subsection{Mirror resolution on York Urban}
 Of the 387 pairs of junctions that share an image, the most consistent mirror hypothesis matched the ground-truth frame in 99.0\% of cases (Wilson 95\% interval 97.4 to 99.6\%), and the closed form of \cref{M-sec:closed} recovered the frame to a median per-axis error of $3.44^\circ$, with the true $K$.

\subsection{The similarity law}

Measured in units of $\rho$ rather than $f$, the interquartile spread of the lower bound narrows from a factor of $3.3$ to $2.2$; the median $\fmin/\rho$ is $1.12$, and $\fmin$ and $\rho$ correlate at $0.51$. \Cref{M-cor:similar} thus describes real junctions too, if less tightly, since their angles vary as well as their radius.

\subsection{Limitations}\label{sec:proposals}\label{sec:missing}

The regime named in \cref{M-sec:intro}, in which no direction carries enough long parallel lines to locate a vanishing point, is reached by none of the experiments here. The synthetic studies impose exact Manhattan geometry, and York Urban consists of hand-labeled urban scenes in which line families are, by construction, abundant. The experiments above show that the bounds survive real segments and that line families, where present, are the better route; it cannot establish the paper's claim about its own applicability.

Further limitations bear on the results. The principal point is taken as known in the focal-length results. Non-right angles are treated by a sufficient condition and in simulation, and distortion by a guarantee for bounded distortion (Sect.~S6), but the York Urban segments were used as given. The real junctions come from hand-labeled segments, and the camera-free test rejects only $4$ of $39$ false junctions. There is one camera and one data set, and the estimator needs ten or more junctions under favorable orientations.

The missing study is an indoor, cluttered, close-range data set with ground-truth intrinsics, on which to measure the junction yield, the share of images with no usable line family, and the accuracy of the bounds against a learned single-image calibrator. Its pipeline is short: segment detection, endpoint grouping as above, the camera-free test of \cref{M-sec:admissible}, and the consensus of Sect.~S8. A script implementing it is supplied; on rendered scenes of boxes, an idealized test, it found $3.8$ junctions per image, and the certified set contained the true focal length for all of them. We regard this study as the paper's principal outstanding obligation: the positioning of \cref{M-sec:intro} is the hypothesis for which the geometry was developed, not a conclusion that the experiments support. Two further studies require hardware: a corner-only front end for camera rigs, and sign cues from an inertial sensor (Sect.~S14).

\section{Further experiments}\label{sec:Sexp}

The protocol is that of \cref{M-sec:experiments} of the article.

\subsection{The closed form}
Over $1000$ corners per field of view at each noise level, at $\sigma=0$ the recovered edge directions agree with the truth to $3\times10^{-7}$ degrees. With noise the median angular error of the three directions is $0.56^\circ$, $1.1^\circ$ and $2.1^\circ$ at $\sigma=0.5$, $1$ and $2$\,px, with 90th percentiles of $2.4^\circ$, $4.5^\circ$ and $7.5^\circ$ and 95th percentiles of $4.2^\circ$, $7.0^\circ$ and $10.6^\circ$ (\cref{fig:closedsolver}a). The dependence on field of view is negligible: at $\sigma=0.5$\,px the medians are $0.54^\circ$, $0.60^\circ$ and $0.55^\circ$ at $40^\circ$, $60^\circ$ and $90^\circ$. The sign test rejects 2.7\%, 5.2\% and 8.3\% of true corners at the three noise levels, its false-reject rate at this edge length.

\subsection{The corner-plus-line solver}\label{sec:e8}
Over $3000$ corner pairs per noise level, at $\sigma=0$ the selected root reproduces $f$ to $10^{-15}$ relative error. The quadratic \eqref{eq:quadratic} had one positive root in 38\% of noiseless configurations and two in 62\%; among the two-root cases at $\sigma=0.5$\,px the sign test of \cref{M-sec:sign} at the second corner alone removed the spurious root in 34\%, and $E'$ was needed in the remaining 66\%. With noise the quadratic can also lose both roots, which happened in 8.1\%, 12.5\% and 16.0\% of pairs at $\sigma=0.5$, $1$ and $2$\,px; those pairs return no estimate and are excluded from the errors below, so in a RANSAC loop they are simply samples that fail.

\begin{figure*}[t]
\centering
\includegraphics[width=\linewidth]{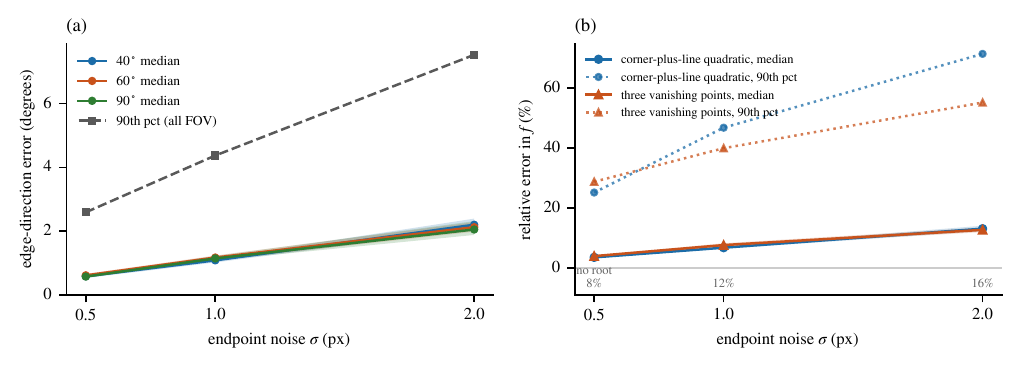}
\caption{(a) Edge-direction error against endpoint noise for one corner with $K$ known: median per field of view with bootstrap 95\% intervals, and the 90th percentile pooled over fields of view (1000 corners per field of view per noise level). (b) Relative focal-length error of the corner-plus-line solver and of three-vanishing-point calibration from two corners, median with bootstrap 95\% intervals (solid) and 90th percentile (dotted); the figures below the axis are the fraction of pairs for which the quadratic has no positive root and the solver returns nothing (3000 pairs per noise level).}
\label{fig:closedsolver}
\end{figure*}

On the pairs that do return an estimate, the median relative error in $f$ is 3.9\%, 6.8\% and 13.5\% at the three noise levels, with 90th percentiles of 26\%, 43\% and 66\%. Three-vanishing-point calibration from two corners of a fully shared frame, on the same data and with the principal point fixed at the image center for both methods, gives medians of 3.9\%, 7.3\% and 11.8\% and 90th percentiles of 27\%, 39\% and 55\%. The solver is therefore level with the baseline in median at low noise and slightly behind it at $\sigma=2$\,px, which is the main result: it reaches the same accuracy from four lines that the baseline needs six to reach. Its tail is the weaker point, heavier than the baseline's at $\sigma=1$ and $2$\,px and driven by pairs whose vertical image lines are nearly parallel. Binning by the angle between the two vertical lines at $\sigma=0.5$\,px, the median error is 1.9\% above $15^\circ$, 3.3\% at $5^\circ$ to $15^\circ$, 6.7\% at $2^\circ$ to $5^\circ$ and 13\% below $2^\circ$ (\cref{fig:closedsolver}b and \cref{sec:conditioning}).

\subsection{Corners against a whole-box fit}\label{sec:e9}
The question this paper opens with is where the corner methods stop being the right tool. A right-angled box with unknown edge-length ratios is imaged with unknown $f$ and known principal point, and its eight vertices are revealed in random order. At each count $V$ the corner estimate, the median over every corner-plus-line configuration available among the visible vertices, is compared against \textsc{box(ml)}, a maximum-likelihood fit of the box by nonlinear least squares over rotation, translation, two length ratios and $f$. \Cref{tab:box} gives the median relative error in $f$ over 400 boxes at $\sigma=0.5$\,px.

\begin{table*}[t]
\centering\small
\caption{Median relative error in $f$ of the corner estimate and of a whole-box fit, by number of visible vertices, over 400 boxes at $\sigma=0.5$\,px}
\label{tab:box}
\begin{tabular}{lcccccc}
\toprule
visible vertices $V$ & 3 & 4 & 5 & 6 & 7 & 8\\
\midrule
corner (\cref{sec:Ssolver})              & 3.0\% & 2.3\% & 1.9\% & 1.8\% & 1.7\% & 1.4\%\\
\textsc{box(ml)}, vertex positions only & ---   & ---   & 4.8\% & 2.6\% & 2.3\% & 1.8\%\\
\textsc{box(ml)}, vertices and edges    & 1.5\% & 1.1\% & 1.0\% & 0.9\% & 0.8\% & 0.8\%\\
\bottomrule
\end{tabular}
\end{table*}

The two \textsc{box(ml)} rows differ only in what the fit is allowed to see, and the difference decides the comparison. Given only the positions of the visible vertices, the box fit is underdetermined below $V=5$ and worse than the corner estimate until $V=8$; it is tempting to read that as the corner methods winning below a crossover. They do not. The corner estimate uses fitted edge \emph{directions}, which are measurements of pixels along the edges that a vertex-only fit never sees, so the two estimators are not being given the same evidence. Once the box fit is also given those edge segments, it is determined from $V=3$ and is better than the corner estimate at every count.

The conclusion is therefore about applicability and not accuracy. If the scene really is one box, the vertices are in correspondence and the model is right, then fitting the box jointly is the better estimator, as it should be: it has strictly more structure to exploit. What the corner methods offer is that they need none of that. A junction needs no box, no vertex labelling, no correspondence across the scene and no assumption that the three edges meeting here belong to the same object as the three meeting elsewhere; one doorframe corner and one other vertical line are enough. That is the regime this paper is for, and it is defined by how little is assumed rather than by a vertex count.

We also caution against reading \textsc{box(ml)} as a bound on what any estimator could do. Maximum likelihood is not optimal at finite sample sizes for a nonlinear model, so these rows record what two particular estimators achieve under their respective observation sets.

\subsection{Re-projection through the junction}
Three independently fitted lines do not meet exactly, whereas the derivation needs every interpretation plane to contain the ray, that is $\vn_i\cdot\vr=0$, which is the statement that every line passes through the corner. Each fitted line is therefore re-projected through the estimated junction, keeping its direction and replacing $e_i$ by $-(a_i\hat u+b_i\hat v)$, before \eqref{M-eq:lin} is applied; the observation is made consistent with the model rather than the model applied to an inconsistent observation. The step is small: the median distance from a fitted line to the estimated junction is $0.23$, $0.48$ and $0.91$\,px at $\sigma=0.5$, $1$ and $2$\,px, and omitting it changes the coverage figures of \cref{M-sec:e2} by at most $0.1$ of a point.

\subsection{A junction whose vertex is not seen}\label{sec:occluded}
Nothing in the derivation requires the corner itself to be seen. What it requires is that the three image lines be concurrent, and they are, at the projection of the vertex, whether or not that point appears in the image. A corner whose vertex is occluded, or lies outside the frame, can therefore be used by intersecting three edge segments seen away from it. The cost is the extrapolation lever arm. With segments of $150$\,px that start $0$, $150$ and $300$\,px from the unseen vertex, at $\sigma=0.5$\,px, the median error in the recovered vertex grows from $0.50$\,px to $1.12$\,px and $1.80$\,px. The admissible set nevertheless contains the true $f$ in $96$ to $98\%$ of cases throughout, against $97.5\%$ with the vertex visible. The first-order model of \cref{prop:noise} survives the extrapolation, though it conditions on the junction estimate. Over $300$ corners with $120$ perturbations each, its predicted miss probability was $3.6\%$, $2.9\%$ and $2.5\%$ at gaps of $0$, $150$ and $300$\,px against $2.9\%$, $2.3\%$ and $1.8\%$ measured, a stable ratio, between the unrounded rates, of $0.80$, $0.77$ and $0.74$. It therefore remains a conservative predictor, which is the safe direction for weighting junctions. This matters most in the regime of \cref{M-sec:intro}, where clutter occludes vertices more often than it hides edges; but an extrapolated junction is also the kind of near-concurrent triple that \cref{M-thm:interval}'s test cannot reject, so it leans harder on the concurrency checks of \cref{sec:e4}.

\subsection{Junction sampling against line sampling}\label{sec:Ssampling}
 The comparison proposed in Sect.~S14 is between two RANSAC loops that solve the \emph{same} quadratic, \eqref{eq:cfsolver} for $\omega^{*}=\operatorname{diag}(f^2,f^2,1)$, and score hypotheses identically. They differ only in whether the minimal sample is a junction plus one parallel line or four lines drawn from the labeled direction families. On the 59 images where both return an estimate, the line-sampled loop of \citet{wildenauer2012} has a median relative error in $f$ of 3.5\% against 4.6\% for the junction-sampled loop, with 75th percentiles of 5.1\% and 12.2\%, and the junction loop is the better of the two on only 18 of the 59 images. Changing the thresholds of the junction detector does not reverse this. We therefore record the hypothesis that junction sampling helps as not supported: the sign test does filter roots, but it does not compensate for the scarcity and noisiness of three-family junctions when whole labeled line families are available instead. Junctions are useful where those families are not available, not where they are.

\subsection{Two-sided junctions and the field of view}
Two findings from simulation bear on where this might work. First, with noise, on scenes with York Urban's edge lengths, the robust program is biased low, by $9\%$ and $19\%$ at $\sigma=0.5$ and $1$\,px. The likelihood errs by about $6\%$ and overestimates the noise, because the first-order model of \cref{prop:noise} understates the spread when edges are as short as $33$\,px. Second, the share of two-sided junctions grows with the field of view, from $12\%$ at $51^\circ$ to $19\%$ at $75^\circ$ and $30\%$ at $100^\circ$, the likelihood's median error at $\sigma=0.5$\,px falling from $3.7\%$ to $2.2\%$ and $2.0\%$. At $51^\circ$ that simulation has ten times as many two-sided junctions as the real images, and the reason is the camera's attitude: it tilted the camera by about $12^\circ$, and these images are nearly level.

\subsection{Robustness of the single-junction bound}\label{sec:Srobust}
The protocol of \cref{M-sec:experiments} of the article idealizes three things: orientations are uniform, edges are $150$\,px long, and the principal point is exact. \Cref{tab:Srobust} relaxes each, at $60^\circ$, $\sigma=0.5$\,px and $2000$ junctions per setting. Tilting the vertical as in York Urban, by angles drawn from that dataset's spread (median $3.2^\circ$) with uniform yaw, changes neither the coverage nor the share of junctions bounded above, so the camera's tilt alone does not explain why upper bounds are rare on the real images. Short edges cost coverage but not upper bounds, and an error in the principal point of up to $20$\,px costs almost nothing.

\begin{table}[h]
\centering\small
\caption{The single-junction bound under the relaxations of the protocol}
\label{tab:Srobust}
\begin{tabular}{lccc}
\toprule
setting & coverage & bounded above & median $f_{\min}/f$\\
\midrule
orientations uniform & 97.3\% & 18.8\% & 0.33\\
vertical tilted as in York Urban & 97.0\% & 20.3\% & 0.38\\
edges of 20\,px & 88.7\% & 20.8\% & 0.33\\
edges of 40\,px & 92.5\% & 18.1\% & 0.33\\
edges of 80\,px & 95.1\% & 21.2\% & 0.36\\
edges of 150\,px & 96.6\% & 18.6\% & 0.33\\
principal point off by 5\,px & 97.0\% & 19.8\% & 0.34\\
principal point off by 10\,px & 96.8\% & 18.3\% & 0.35\\
principal point off by 20\,px & 96.7\% & 18.4\% & 0.33\\
principal point off by 40\,px & 93.7\% & 19.3\% & 0.33\\
\bottomrule
\end{tabular}
\end{table}

\subsection{Junction extraction on York Urban}\label{sec:Sextract}
With the endpoint clustering radius at $10$\,px and a minimum segment length of $15$\,px, the settings of the article, $181$ full junctions are found in $59$ images; the admissible set contains the true focal length for $98.9\%$ of them, and the line-sampled loop has median error $3.5\%$ against $4.6\%$ for the junction-sampled one. At $20$\,px and $10$\,px there are again $181$ junctions, with coverage $95.5\%$ and medians $2.6\%$ against $4.7\%$; at $25$\,px and $10$\,px there are $161$ junctions in $56$ images, with coverage $95.6\%$ and medians $3.1\%$ against $4.4\%$. The comparison does not reverse. The $59$ vertical directions of the York Urban frames lie within $18.6^\circ$ of their mean, half of them within $3.5^\circ$, and the smallest singular value of the matrix they form is $1.1\%$ of the largest, so the conditions of Theorem (several frames) that they impose are nearly dependent.

\part{Outlook}

\section{Directions for further work}\label{sec:directions}
Each direction below follows from \cref{M-sec:bound,sec:two} of the article and \cref{sec:Ssolver}; \cref{tab:proposals} states the lineage of each, because several have a direct ancestor.

\begin{table*}[t]
\centering\small
\caption{Directions and their lineage.}
\label{tab:proposals}
\begin{tabularx}{\linewidth}{l >{\raggedright\arraybackslash}X l >{\raggedright\arraybackslash}X}
\toprule
& Direction & Builds on & Lineage \\
\midrule
\ref{sec:ransac} & Junction-sampled RANSAC, against the line-sampled loop of \citet{wildenauer2012} & \cref{sec:Ssolver} & \citet{wildenauer2012} \\
\ref{sec:signcues} & Single-corner sign resolution from gravity or occlusion & \cref{M-sec:mirror1,sec:opacity} & gravity cue used by \citet{shakunaga1989} \\
\ref{sec:rig} & Corner-only rotation front end for non-overlapping rigs & \cref{sec:multiview} & full-box version by \citet{wilczkowiak2005} \\
\ref{sec:distortion} & Joint distortion or rolling-shutter estimation from edge straightness & \cref{M-sec:closed} & plumb-line calibration; new primitive \\
\ref{sec:detector} & Junction detector that predicts the canonical angles and the sign condition & \cref{M-sec:sign} & none found \\
\bottomrule
\end{tabularx}
\end{table*}

\subsection{The solver in practice}\label{sec:ransac}
\citet{wildenauer2012} already run a four-line minimal solver inside RANSAC on real images, sampling lines clustered by direction, so the open question is not whether a four-line sample is practical but whether sampling \emph{junctions} rather than lines helps. A junction supplies three lines at once with their incidence at a common point already established, and it carries the sign test of \cref{M-sec:sign}, which rejects a sample using only the corner's own geometry and the candidate $K$, before any other line is consulted. Against that, junction detection is harder than line detection, most junctions in a room are not right angles, which edge is vertical comes from a detector, and corner pairs must still be sampled with well-separated vertical lines (\cref{sec:conditioning}). \Cref{sec:e4} runs that experiment on the York Urban Database and settles it against the junction-sampled loop: on the images where both return an estimate it is the less accurate of the two, because full three-family junctions are scarce, about 1.8 per image, where labeled line families are not. The open question that remains is narrower: whether junction sampling helps in the regime it was meant for, namely images in which no direction has enough parallel lines to form a family at all. The York Urban images, which are hand-labeled with such families, do not represent that regime.

\subsection{Resolving the sign from one corner}\label{sec:signcues}
Three cues, in increasing cost. \emph{Gravity}: if one edge is known to be vertical, only one of the two mirror solutions has a vertical edge, since the reflection in the plane perpendicular to the ray tilts a vertical direction unless the ray is horizontal. \citet{shakunaga1989} already chose between candidates by assuming one direction parallel to the ground. \emph{Occlusion}: by \cref{sec:opacity} the barbs of an arrow are occluding edges under the convex reading and convex edges under the concave one, so T-junctions and texture discontinuity on the barbs decide. \emph{Learned}: a two-class convex/concave classifier on a small crop.

\subsection{Corner-only rig calibration}\label{sec:rig}
\citet{wilczkowiak2005} calibrate non-overlapping cameras with a box: a parallelepiped seen by different cameras at different times links their canonic projection matrices, and a factorization gives every camera's intrinsics, rotation and position; see also \citet{esquivel2007}. The corner-only version needs three lines of one junction per view instead of six matched vertices, which matters for rigs where a whole box is rarely in view, and gives rotation only. Translation needs a connected pair and a known edge length, at which point the full parallelepiped method is the better tool.

\subsection{Distortion and rolling shutter}\label{sec:distortion}
The three edges are straight in the scene, so any curvature in their images is attributable to lens distortion or rolling shutter. A single corner provides three curved segments from which a one-parameter radial model can be fitted jointly with the three directions, using sampled edge points rather than fitted lines as observations.

\subsection{The sign condition as a differentiable penalty}\label{sec:difflearn}
\Cref{M-prop:general} is differentiable in both the line parameters and $\omega^{*}$, so it can be attached to a learned junction detector as a penalty on predicted line orientations. The form to use is not the raw product, which scales as the sixth power of a length and is hopeless numerically, but its normalization
\[
\mathcal{L}_{\text{ortho}}=\max\Bigl(0,\ \textstyle\prod_{(i,j)}\widehat{\vn}_i\cdot\widehat{\vn}_j\Bigr),\qquad \widehat{\vn}_i=\vn_i/|\vn_i| ,
\]
a product of three cosines, bounded in $[-1,1]$, scale-free, and zero exactly on the admissible set. Whether this helps a detector is an empirical question we have not tested.

\subsection{Predicting the constraint}\label{sec:detector}
Current junction detectors output line segments, after which a separate step tests orthogonality. A detector could instead output, per junction, the three canonical-view angles and a probability that the junction is a right-angle corner. The sign condition \eqref{M-eq:sign} is a differentiable, label-free regularizer on the predicted angles given $K$, and the detector's confidence can be calibrated against the condition's margin. Synthetic box scenes supply training data.

A last direction, constraining $f$ through regular texture on the faces by way of the face homography that the closed form of \cref{M-sec:closed} determines, has not been formulated here and is left open.


\subsection{Trihedra of known non-right angles}
The restriction to right angles is not a convenience. With $\vd_i\propto k_i\vr+\vt_i$ and a prescribed scene cosine $c_{ij}$ between edges, the pairwise condition is
\[
\bigl(k_ik_j\|\vr\|^2+g_{ij}\bigr)^2=c_{ij}^2\bigl(k_i^2\|\vr\|^2+\|\vt_i\|^2\bigr)\bigl(k_j^2\|\vr\|^2+\|\vt_j\|^2\bigr).
\]
At $c_{ij}=0$ the right side vanishes and the condition is a perfect square, $(k_ik_j\|\vr\|^2+g_{ij})^2=0$, so that $k_ik_j=-g_{ij}/\|\vr\|^2$ depends on $g_{ij}$ alone; that decoupling is what produces the closed form and, through \eqref{M-eq:gfactor}, the linearity in $f^2$. At any other angle the normalizations $\|\vt_i\|^2$ survive, and since they depend on $f$ nonlinearly the structure is lost. A trihedral of known non-right angles still constrains $f$, but not in closed form, and we have not characterized its admissible set.

\bibliographystyle{plainnat}
\bibliography{refs}